%% file: main.tex
\documentclass[11pt]{article}

\usepackage[utf8]{inputenc}
\usepackage{lineno}
\usepackage{amsmath}
\usepackage{graphicx}
\usepackage[colorlinks=true, allcolors=blue]{hyperref}
\usepackage{xspace}
\usepackage{easyReview}
\usepackage{setspace}
\usepackage{rotating,tabularx}
\usepackage{scrextend}
\usepackage{amssymb}
\usepackage{geometry}
\usepackage{subcaption}

\usepackage[style=nejm,%
            minnames=3, maxnames=6,%
            terseinits=true,%
            firstinits=true,
            backend=biber]{biblatex}
\title{\textbf{\framework:\\A Unifying Causal Framework for Scaling Real-World Evidence Generation with Biomedical Language Models}}

\author{Javier González$^{1*}$, Risa Ueno$^{1*}$, Cliff Wong$^1$, Zelalem Gero$^1$, Jass Bagga$^1$, \\
Isabel Chien$^{2+}$, Eduard Orakvin$^{3+}$, Emre Kiciman$^1$, Aditya Nori$^1$, Roshanthi \\
Weerasinghe$^4$ , Rom S. Leidner$^4$, Brian Piening$^4$ , Tristan Naumann$^1$,\\
Carlo Bifulco$^4$, Hoifung Poon$^{1}$\footnote{Javier González and Risa Ueno contributed equally to this manuscript. Correspondence: \url{Gonzalez.Javier@microsoft.com}, \url{hoifung@microsoft.com}\\
\indent $^+$Work done at Microsoft \\
\indent $^1$Microsoft\\
\indent $^2$University of Cambridge\\
\indent $^3$University of Oxford\\
\indent $^4$Providence Genomics}
}

\newcommand{\eat}[1]{\ignorespaces}

\input{tex/acronyms}

\begin{document}
\maketitle



\setstretch{1.2}

\input{sections/0_abstract}

\input{sections/1_introduction}
\input{sections/2_methods}
\input{sections/3_results}
\input{sections/4_discussion}

\printbibliography



\input{sections/5.2_supplementary}

\end{document}

%% file: tex/acronyms.tex
\newcommand{\acr}[1]{\textsc{#1}\xspace}

\newcommand{\framework}{TRIALSCOPE\xspace}

\newcommand{\tfidf}{\acr{\tfidf}}

%% file: sections/0_abstract.tex
\begin{abstract}

The rapid digitization of real-world data presents an unprecedented opportunity to optimize healthcare delivery and accelerate biomedical discovery. However, these data are often found in unstructured forms such as clinical notes in electronic medical records (EMRs), and is typically plagued by confounders, making it challenging to generate robust real-world evidence (RWE). Therefore, we present TRIALSCOPE, a framework designed to distil RWE from population level observational data at scale. TRIALSCOPE leverages biomedical language models to structure clinical text at scale, employs advanced probabilistic modeling for denoising and imputation, and incorporates state-of-the-art causal inference techniques to address common confounders in treatment effect estimation. Extensive experiments were conducted on a large-scale dataset of over one million cancer patients from a single large healthcare network in the United States. TRIALSCOPE was shown to automatically curate high-quality structured patient data, expanding the dataset and incorporating key patient attributes only available in unstructured form. The framework reduces confounding in treatment effect estimation, generating comparable results to randomized controlled lung cancer trials. Additionally, we demonstrate simulations of unconducted clinical trials — including a pancreatic cancer trial with varying eligibility criteria — using a suite of validation tests to ensure robustness. Thorough ablation studies were conducted to better understand key components of TRIALSCOPE and establish best practices for RWE generation from EMRs. TRIALSCOPE was able to extract data cancer treatment data from EMRs, overcoming limitations of manual curation. We were also able to show that TRIALSCOPE could reproduce results of lung and pancreatic cancer clinical trials from the extracted real world data.

\end{abstract}

%% file: sections/1_introduction.tex
\begin{figure*}[t!]
\caption{\emph{
A medical journey of a (de-identified) lung cancer patient, showing the various sources of relevant clinical information. Real-world data is rich, complex and mostly unstructured, so manual curation is difficult to scale. \textbf{A:} Each bar represents a clinical note with information about the patient at each corresponding time. \textbf{B, C, D}: Example synthetic notes with key information such as diagnosis and outcome. \textbf{E}: Distribution of the most abundant note types in our real-world dataset.
}\label{fig:data_complexity}}
\vspace{0.5cm}
\centering
\includegraphics[width=17cm]{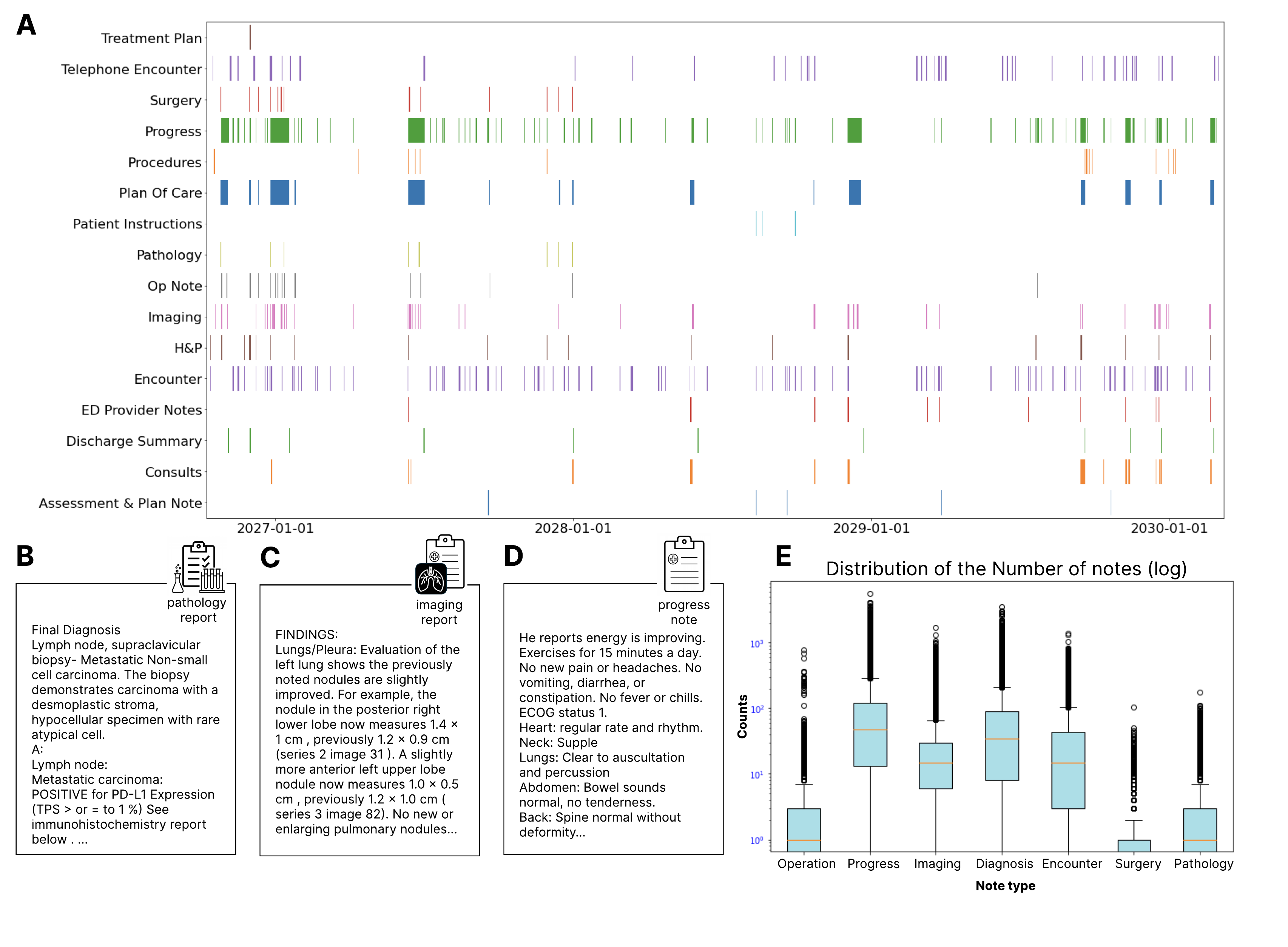}
\end{figure*}

\section{Introduction}\label{sec:introduction}

Real-world evidence (RWE) refers to the use of real-world data (RWD) routinely collected from clinical sources, such as electronic medical records (EMRs), to generate clinical evidence on the potential benefits and risks of medical treatments. RWE is increasingly recognized as a valuable resource for clinical decision-making and improving patient care. Notably, RWE represents a promising direction to augment randomized controlled trials (RCTs)~\parencite{concato2022real,hernanUsingBigData2016,chen2021feasibility}, as recognized by various regulatory bodies (\textit{e.g.}, the U.S. Food and Drug Administration's (FDA) Real World Evidence Program~\parencite{FDA2024} and the National Institute for Health and Care Excellence (NICE)~\parencite{Nice2022}). While RCTs remain the gold standard for establishing treatment efficacy~\parencite{Young2011,Lawlor2004}, they are time-consuming, extremely costly, and often difficult to conduct successfully~\parencite{liu2022real}. Furthermore, outcomes from RCTs may lack applicability to broader populations~\parencite{kennedy2015literature}. Conversely, RWE captures real-life medical scenarios and can encompass larger and more diverse patient populations, including historically underrepresented sub-populations and patients with rare conditions or those undergoing concurrent treatments~\parencite{Sharma2021}, enabling clinical insights in situations where RCTs are impractical, insufficient, or unethical.

In recent years, EMRs have facilitated data-driven approaches to inform treatment planning, outcome prediction, and diagnosis, among others, with emerging use cases such as post-market monitoring, early approval, and establishing historical controls~\parencite{kennedy2015literature}. A recent example of success is the use of RWE in the label expansion of Ibrance (i.e., PAL, a cyclin-dependent kinase 4/6 inhibitor) to treat male breast cancer~\parencite{Bartlett2019}, for which RCTs are infeasible given the rarity of this disease. Currently, more than 90\% of rare diseases lack approved treatment~\parencite{kaufmannScientificDiscoveryTreatments2018}. RWE can also help identify predictive biomarkers~\parencite{chenhuBiomarker2019} and allows analysis of long-term outcomes beyond the typical clinical trial duration~\parencite{FRITZSCHING2022100275}. With growing digitization in healthcare, RWE presents an opportunity to accelerate clinical discovery and reduce the burden on patients, providers, and caregivers.

However, RWE also pose challenges due to the often noisy and unstructured nature of RWD. For example, manual curation of clinical notes is expensive and difficult to scale (see \autoref{fig:data_complexity} for an example of medical data sources associated with one patient). Issues such as data incompleteness, inconsistency, and heterogeneity also affect its regulatory validity and hinder overall adoption~\parencite{nazha2021benefits}. Obtaining causal insights into treatment efficacy can be challenging due to inherent biases in the data, such as confounding. For widespread adoption, it is crucial to establish methods that ensure data quality and enable robust analysis.

To this end, we developed \framework, a framework for generating population-level RWE from EMRs at scale. \framework combines automated EMR curation from unstructured and noisy data sources with a clinical trial simulator for causal treatment effect estimation. We integrate state-of-the-art methods to enhance data quality, with a quality evaluation process designed for robustness and transparency. Prior works~\parencite{liu2021eligibility} exploring clinical trial simulation rely on available structured databases such as Flatiron~\parencite{Flatiron2023}, which require expensive and laborious manual curation. \framework instead leverages the rapid progress in generative AI, specifically biomedical language models, to scale real-world data structuring, thus unlocking the use of much more abundant unstructured real-world data.~\parencite{head2015phacking,schuemie2020legend,schuemie2020principles} We apply \framework to a large-scale retrospective study of real-world lung cancer patients from a large integrated delivery network (IDN) comprising healthcare systems in five western states in the United States.

%% file: sections/2_methods.tex
\section{Methods}\label{sec:methods}

In our experiments, we assessed our automation pipeline for structuring EMR data and validated the quality of our simulations. Next, we demonstrate the use of \framework to conduct experiments for treatment effect estimation, while maintaining transparency in its assumptions and methods. We showcase its ability to simulate trials that do not have published results from a RCT and develop a validation test suite to assess its reliability.

\subsection{\framework components}\label{method:trialscope}

\begin{figure}[t!]
\caption{\emph{
\framework is an end-to-end causal framework for scaling clinical trial simulation and real-world evidence generation from electronic medical records (EMRs). \textbf{A}:
1. Biomedical language models automates the structuring of EMR texts. 2. Probabilistic latent-variable models provide a principled way for data denoising and imputation to improve data quality. 3. Assembling of a virtual cohort for the given clinical research hypothesis, akin to clinical trial matching. 4. Applying a causal Cox survival model with inverse propensity weighting to alleviate confounding bias. 5. Validation test suite to assess simulation quality. \textbf{B}: Comparison between TRIALSCOPE and traditional manual curation. 
}}
\centering
\includegraphics[width=17cm]{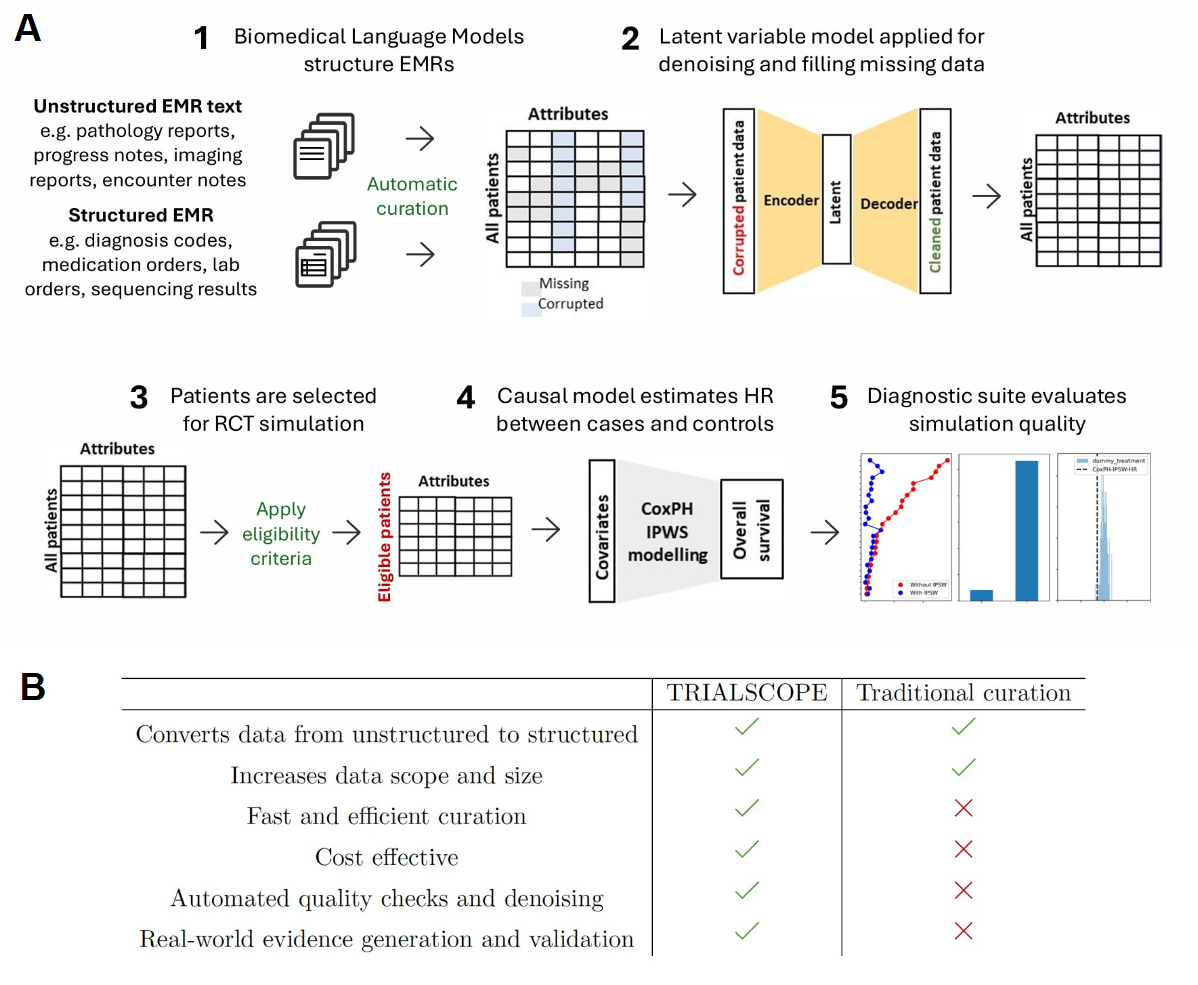}
\label{fig:trialscope}
\end{figure}

\framework consists of five main components (\autoref{fig:trialscope}A): (i) a data structuring pipeline based on large language models that converts scanned clinical reports to text to generate structured representations of patient journeys; (ii) a probabilistic latent-variable model for data denoising and imputation; (iii) a patient triaging pipeline to identify real-world patients according to a target trial specification; (iv) a causal model for simulating the target trial; and (v) a set of validation tests to assess the trustworthiness of the simulation. The comparison summary between \framework and traditional curation is shown in \autoref{fig:trialscope}B.

\subsection{Performance assessment}\label{method:performance}


Following prior work such as \textcite{Liu2021}, we focused on advanced non-small cell lung cancer (aNSCLC) given its prevalence and high mortality rate (see Supplementary Appendix \ref{sup:oncotree} for the oncotree identifiers). For real-world data, we used the cancer patient cohort from a large U.S. Health Network~\parencite{preston2023toward}. See Supplementary Appendix \ref{sec:methods_supp} for more information on these methods.

\subsubsection*{EMR curation}

For attributes such as tumor site, histology, and staging, extraction requires synthesizing multiple notes in longitudinal patient records complicated by linguistic variations and ambiguities. Consequently, we build on \textcite{preston2023toward} and leveraged advanced biomedical language models such as PubMedBERT~\parencite{Gu2021} and OncoBERT~\parencite{preston2023toward} for automatic structuring from EMRs. Following \textcite{preston2023toward}, we used available cancer registry test data for validation. For relatively simple attributes such as Eastern Cooperative Oncology Group (ECOG) Performance Status (which measures how a patient’s disease affects their daily life and physical capacity), medications, and PD-L1 immunohistochemistry (IHC) biomarkers (relevant for immunotherapy trials), we used a conventional information extraction pipeline (see Supplementary Appendix Sections \ref{sec:methods_pipeline} and \ref{sec:curation_evaluation} for methodological details on the curation pipeline and validation process). 

\subsubsection*{RCT emulation}

We simulated a total of 11 previously published aNSCLC clinical trials to evaluate whether the evidence produced is consistent with validated biomedical evidence. We used the hazard ratio (HR) for overall survival as the key metric to evaluate the success of each simulation \parencite{George2020WhatsTR}. For trials with published HR, we compared our simulated HR against the reference HR with a standard two-sample statistical equivalence test. We conducted an search in \textit{ClinicalTrials.gov} for eligible standard two-arm trials with sufficient support in our dataset (see Supplementary Appendix \ref{sup:trial_selection}). In addition to seven trials considered in \textcite{Liu2021}, we explored four new trials that have not been explored in prior work: FLAURA, CHECKMATE057, CHECKMATE078, KEYNOTE010, OAK, KEYNOTE024, KEYNOTE033, LUXLUNG6, CHECKMATE057017, EMPHASIS and NCT02604342\footnote{Code as available in ClinicalTrials.gov}. We found a total of 18,038 aNSCLC patients that received the treatment in at least one of the arms in any of the selected trials. Pre-processing was applied to the structured data to guarantee that the RCT and the simulations were as comparable as possible, including data imputation and other data cleaning steps (see Supplementary Appendix \ref{sec:study_cohort} and \ref{sec:missing_data}).


To address confounders that impact the quantification of treatment effects~\parencite{Schuemie2020b}, we incorporated state-of-the-art causal survival analysis methods based on the potential outcome framework ~\parencite{Imbens2015,Dahabreh2022}. We used the Cox proportional hazards (Cox-PH) model \parencite{cox1972regression} to account for biases due the lack of randomization in the data, where confounders are incorporated via inverse propensity score weighting (IPSW) with weights computed using logistic regression \parencite{rosenbaum1983central}. See Supplementary Appendix Section \ref{sec:causal_modeling} for further technical details and references. 

\subsection{Robustness testing for trial simulations}\label{method:validation}

\framework can be used to simulate any trial specified by the user. Here, we outline a set of methods to assess the credibility of a simulation for cases where a reference HR is not available. While we use causal assumptions (see Supplementary Appendix Section \ref{sec:causal_modeling}) that are not readily verifiable from data, diagnostic tests can capture model behaviors that indicate potential violations of these assumptions. Such considerations are essential to ensure reliable real-world evidence generation. To assess the reliability of our trial simulation where a reference hazard ratio (HR) is unavailable, we applied the following tests:


\begin{enumerate}
    \item We checked the balancing between treatment and control groups before and after using the IPSW correction: Compute the standardized mean distance (SMD) for all confounders in both cases. The differences should be close to zero when the IPSW correction is applied and remain large when it is not. This would indicate that the differences between the groups due to factors beyond the treatment are properly corrected.
    \item We checked the percentage of the data with non-null probability that could have been selected for either control or treatment groups: The larger the set, the more suitable the dataset is for causal analysis \parencite{oberst2020characterization}. 
    \item We tested the strength of the signal: For a patient cohort, randomly reassign each patient to either the treatment or control group, and compute the HR for this randomized permutation. This process is repeated 100 times to generate a distribution of HRs. With random assignment, any strong initial signal should vanish.
    \item We assessed model stability against a random noise variable: Add a variable drawn from a zero-mean Gaussian distribution (i.e., pure noise) to the model. Run several scenarios in which the standard deviation of this noise variable is systematically increased (e.g., between 0.1 and 5). We generate 100 replicates for each scenario and compute the 95\% confidence interval (CI) for the replicates using the 5\% and 95\% percentiles. If the signal-to-noise ratio is robust, adding this extra variable should not affect the HR estimation and the variation should be minimal.
    \item We computed HR with random down-sampling of data: We perform simulations while retaining only 95\%, 90\%, 75\%, 50\%, and 25\% of the patients (100 repetitions in each scenario). A well-behaved simulator should provide the same average HR in all cases, with an increase in variance when the sample size decreases.
\end{enumerate}


%% file: sections/3_results.tex
\section{Results}\label{sec:results}


\subsection{\framework performance}\label{sec:performance}

\begin{figure*}[t!]
\centering
\includegraphics[width=12cm]{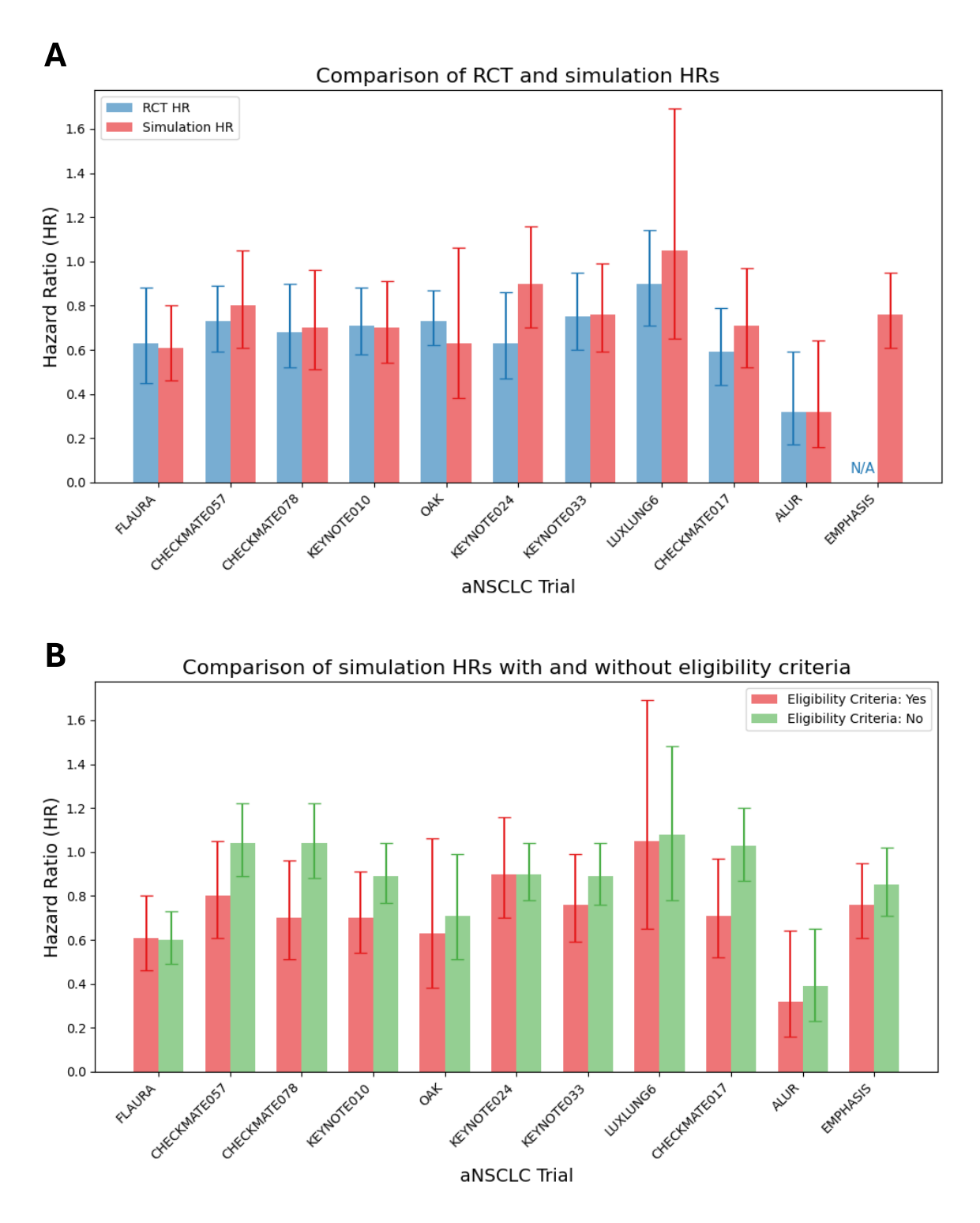}
\caption{A: Comparison between RCT results and our simulation of the 11 selected single-drug trials using a Cox-PH model with inverse propensity re-weighting. The hazard ratio (HR) and its 95\% confidence interval are shown for the original RCTs and for the simulations. EMPHASIS indicates a simulated trial with no reported (ground-truth) HR. B: Comparison between simulated HRs with and without applying the eligibility criteria. “Yes” means only patients that meet the original eligibility criteria were included in the simulation. “No” means we use datasets of patients that were filtered only by line of therapy and the treatment and control drugs of the trial.}
\label{fig:performance_plots}
\end{figure*}



\begin{table}[t!]
\centering
\small
\vspace{0.5cm}
\begin{tabular}{lc|cc|cccc} 
\hline \hline
\multicolumn{2}{c|}{\textbf{Trial and E. Criteria}} & \multicolumn{2}{c|}{\textbf{RCT}} & \multicolumn{4}{c}{\textbf{Simulation HR}}  \\
&  &  \textbf{HR}& \textbf{95\%CI} &  \textbf{HR}& \textbf{95\%CI} &  \textbf{C} & \textbf{T}  \\
\hline \hline
FLAURA & \textbf{Yes} & 0.63 & (0.45, 0.88) & 0.61 & (0.46, 0.80) & 268 & 200\\
 &  \textbf{No} & & & 0.60 & (0.49, 0.73) & 576 & 312 \\
 \hline
CHECKMATE057 & \textbf{Yes} & 0.73 & (0.59, 0.89) & 0.80 & (0.61, 1.05) & 263 & 152\\
 &  \textbf{No} & & & 1.04 & (0.89, 1.22) & 491.0 & 435 \\
 \hline
CHECKMATE078 & \textbf{Yes} & 0.68 & (0.52, 0.9) & 0.70 & (0.51, 0.96) & 194 & 111\\
 &  \textbf{No} & & & 1.04 & (0.88, 1.22) & 492 & 439 \\
 \hline
KEYNOTE010 & \textbf{Yes} & 0.71 & (0.58, 0.88) & 0.70 & (0.54, 0.91) & 261 & 362 \\
 &  \textbf{No} & & & 0.89 & (0.77, 1.04) & 494 & 867 \\
 \hline
OAK & \textbf{Yes} & 0.73 & (0.62, 0.87) & 0.63 & (0.38, 1.06) & 276 & 41\\
 &  \textbf{No} & & & 0.71 & (0.51, 0.99) & 572 & 83 \\
 \hline
KEYNOTE024 & \textbf{Yes} & 0.63 & (0.47, 0.86) & 0.90 & (0.70, 1.16) & 180 & 410 \\
                             &  \textbf{No} & & & 0.90 & (0.78, 1.04) & 606 & 1147 \\
 \hline
KEYNOTE033 & \textbf{Yes} & 0.75 & (0.60, 0.95) & 0.76 & (0.59, 0.99) & 230 & 360\\
 &  \textbf{No} & & & 0.89 & (0.76, 1.04) & 492 & 862 \\
 \hline
LUXLUNG6 & \textbf{Yes} & 0.90 & (0.71, 1.14) & 1.05 & (0.65, 1.69) & 2621 & 37\\
 &  \textbf{No} & & & 1.08 & (0.78, 1.48) & 5187 & 82 \\
 \hline
CHECKMATE017 & \textbf{Yes} & 0.59 & (0.44, 0.79) & 0.71 & (0.52, 0.97) & 197 & 111\\
 &  \textbf{No} & & & 1.03 & (0.87, 1.20) & 492 & 437 \\
 \hline
EMPHASIS & \textbf{Yes} & ? & ? & 0.76 & (0.61, 0.95) & 370 & 439\\
 &  \textbf{No} & & & 0.85 & (0.71, 1.02) & 534 & 615\\
 \hline
NCT02604342 & \textbf{Yes} & ? & ? & 0.32 & (0.16, 0.64) & 1774 & 26\\
 &  \textbf{No} & & & 0.39 & (0.23, 0.65) & 2625 & 42 \\

\hline\hline
\end{tabular}\caption{\emph{Simulation of the 11 selected single-drug trials using a CoxPH model with inverse propensity re-weighting. Each pair of rows in the table contains the results for one trial. Rows labelled with \emph{No} are results with datasets of patients that were filtered only by line of therapy and the treatment and control drugs of the trial. Rows in the table labelled with \emph{Yes} also exclude patients filtered using all the eligibility criteria. The hazard ratio (HR) and its 95\% confidence interval are shown for the original RCTs and for the simulations that use the Providence dataset. The sample size of the treatment (T) and control (C) groups are also included. Note that the HR from the RCTs are only comparable to the simulation results in which patients are filtered by eligibility criteria (EC). }}\label{table:simulation_results}
\end{table}

\vspace{0.5cm}
\noindent
\textbf{Automatic curation of electronic medical records}: We obtained high test accuracy for patient attributes curated from both biomedical language models (AUROC $>$96.6\%) and information extraction pipelines (precision and recall consistently over 87\%). For a full table of results, see Supplementary Appendix Section \ref{sup:test_results_curation}. 
This demonstrates \framework's ability to automate the supply of high quality structured EMRs from a variety of unstructured data sources, and could effectively augment manual data curation efforts. 



\vspace{0.5cm}
\noindent
\textbf{Success of simulations of aNSCLC trials}: \framework simulations aligned closely with the published trials, demonstrating statistical equivalence in the majority of cases and accurately capturing the hazard ratio (HR) for overall survival between treatment and control groups. The results of the 11 simulated trials are summarized in Table \ref{table:simulation_results} and \autoref{fig:performance_plots}A. Of the 11 trials, 2 do not have published HR due to the lack of statistically significant results in the original trials. In the remaining 9 trials, the simulations and the published trial are statistically equivalent, although the evidence for KEYNOTE024 is weaker. The alignment is consistent regardless of the value of the HR in the original publication. They are consistent also in cases where differences between groups are significant and also when they are not. Trials such as FLAURA (HR-RCT: 0.63\footnote{See Table \ref{table:simulation_results} for 95\% CI}, $n=556$) and CHECKMATE057 (HR-RCT: 0.73, $n=582$) with $HR<1$ are correctly simulated. On the other hand, the LUXLUNG6 trial (HR-RCT: 0.9, $n=364$) had a non-statistically equal-to-one HR. The simulations capture this effect with a value of $HR=1.05$ ($n=2658$). This is an important result that highlights the correct alignment of information provided by the structuring of EMRs using language models, with existing comparative evidence of lung cancer treatments. 


\vspace{0.5cm}
\noindent
\textbf{Effects of the eligibility criteria}: It is possible that for many trials the eligibility criteria can be relaxed while maintaining a similar level of efficacy and safety. We demonstrate this by comparing simulation results with and without applying the original eligibility criteria (\autoref{table:simulation_results} and \autoref{fig:performance_plots}B, for rows with eligibility criteria = Yes and No, respectively). This conclusion is consistent with \textcite{Liu2021}. The differences in the estimated HR vary between trials: in KEYNOTE024, the variation is minimal (HR-No-Criteria=0.90, n=2753; HR-with-Criteria=0.90, n=580), whereas in the CHECKMATE057 the differences are larger (HR-no-Criteria=1.04, n= 926; HR-with-Criteria: 0.80, n=415). \framework shows close alignment with published trials and established literature, demonstrating its robustness for these tasks.


\subsection{Simulating unconducted trials}\label{sec:sim_validation}

EMPHASIS is a trial that was discontinued due to lack of enrolled patients. This is a common problem: many clinical trials fail solely due to the inability to recruit eligible patients \parencite{briel2021exploring}. TRIALSCOPE enables the simulation of unconducted trials to predict the HR. As an example, we simulated the efficacy of the drugs used in the EMPHASIS trial as first-line treatment. (see \autoref{table:simulation_results}). This results in $HR=0.76$ (with 95\% confidence interval of $[0.61, 0.95]$, n=1009), which suggests evidence for treatment efficacy. This exemplifies how this framework could be used, for example, to collect early evidence of the result of a trial. 

This trial simulation passes all five diagnostic tests outlined in Methods, demonstrating its robustness: for example, 93\% of the RWD for this trial satisfies the property for test 2 (see Supplementary Appendix Section \ref{sup:emphasis_diagnostics_results} for detailed results). 

Another important aspect to consider is the characteristics of patients involved in the trials. The trial and simulation populations are expected to be similar if both studies are comparable. In the case of our simulation, we observed strong alignment in the two populations in the age variable but not necessarily in other variables (see Supplementary Appendix Section \ref{sup:EMPHASIS_summary_stats}). It is important to note that even though the models and assumptions in the simulations are accurate, as per the aforementioned test, conclusions about the differential effects of the trial and control drugs are applicable only to the population of patients represented in the simulation.  


In summary, this trial is an exemplar of expected simulation behavior. For all trial simulations, we recommend that validation tests like these are conducted to ensure its reliability.

\subsection{Generalizability beyond lung cancer}

To explore the potential for \framework to extend its application beyond aNSCLC, we emulate the Metastatic Pancreatic Adenocarcinoma Clinical Trial (MPACT; NCT00844649 \parencite{VonHoff2013}) that compares nab-paclitaxel plus gemcitabine versus gemcitabine monotherapy for metastatic pancreatic cancer patients (see Supplementary Appendix \ref{sup:pancreatic} for details). With the eligibility criteria applied, we obtain HR = 0.77 (95\% CI = [0.48, 1.24], n = 131), which is statistically equivalent to the reported HR = 0.72 (95\% CI = [0.62, 0.83]). Removing the eligibility criteria to include all available patients in our database, and including locally advanced pancreatic cancer patients, our simulation yields HR = 0.69 (95\% CI = [0.485, 0.985], n = 214). The framework shows potential for adaptation to different disease types and treatment regimens, and its flexibility to modify the simulation eligibility criteria can help us understand how interventions may perform in diverse real-world populations. These analyses can become increasingly robust as data availability and quality continue to improve.

\subsection{Comparison with traditional curation}

\begin{figure*}[t!]
\centering
\includegraphics[width=17cm]{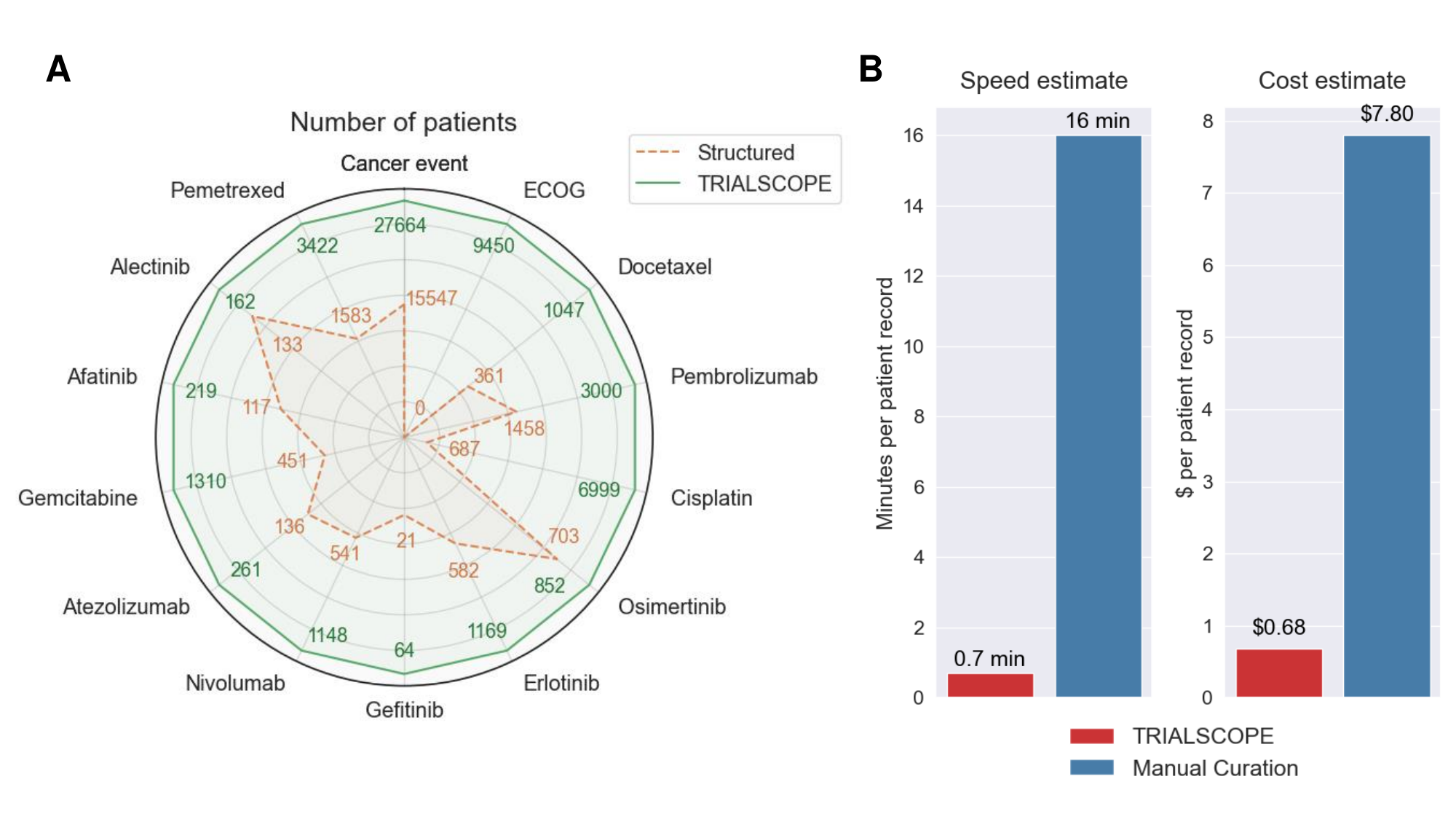}
\caption{\emph{Quantified improvements made by \framework: \textbf{A}: Number of patient records extracted for cancer event, ECOG, and 12 lung cancer medications (accessible sample sizes in the Providence dataset) when using only structured data (orange) and the combination of structured and free-text data (green). For ECOG, all records originate from unstructured free-text. \textbf{B} A bar plot comparing estimated speed and cost of \framework compared to manual curation, by minutes taken per patient curated and USD per patient curated, respectively.}}
\label{fig:advantages_summary}
\end{figure*}

\textbf{Increase in scope, size and quality of patient records:} By combining structured and unstructured free-text EMR data, we increased the number of patient records and patient attributes available in our studies. \autoref{fig:advantages_summary}A compares the number of unique patients available in our dataset originating from structured data only and from the combination of structured data and unstructured free-text data (i.e., the resulting dataset from automated curation), for the following variables: lung cancer event, ECOG score, and all 12 lung cancer medications used in our RCT emulation study. We obtain more than 12,000 additional lung cancer patient records using our pipeline, and we see a substantial increase in the number of records with individual medications. Key attributes like ECOG and certain biomarkers are often not readily available in structured form. For such categories, all records originate from information extracted from free-text. With our automated curation, we integrated probabilistic latent-variable models to provide a principled way for data denoising and imputation to improve data quality. This reduces the effect of noise and missing data in downstream analysis and real-world evidence generation (see Supplementary Appendix Section \ref{sec:missing_data}).

\vspace{0.5cm}
\noindent
\textbf{Improvements in speed and cost efficiency}: By automatically extracting and structuring data from raw EMRs, we increase the size of high quality patient records substantially, which may otherwise require hours of manual work at a significant cost. Manual curation by trained abstractors could take hours for a single cancer patient\parencite{levie2023quality}. Based on our curation speed experiment and running cost calculations (estimated with compute resources used and internally conducted manual curation - see Supplementary Appendix Section \ref{sec:speed_cost_method}), we estimate more than a 20-fold improvement for speed and a 10-fold improvement for cost per patient curated (\autoref{fig:advantages_summary}B). 
\vspace{0.5cm}

While subject to variability in practice, these estimates suggest that automation can improve compute efficiency and cost-effectiveness by orders of magnitude. A key advantage is the ability to automatically update as new patient data becomes available, significantly reducing overhead. This is especially valuable given the continued growth in medical data digitization, where maintaining up-to-date, high-quality datasets is critical for downstream real-world evidence generation.




%% file: sections/4_discussion.tex
\section{Discussion}\label{sec:discussion}

We have presented TRIALSCOPE, a framework for scaling clinical trial simulation and RWE generation directly from EMRs, where the majority of information resides in unstructured data such as clinical notes. Our experimental results demonstrate that combining structuring capabilities powered by biomedical language models and causal inference capabilities can transform EMRs into scalable real-world evidence generators. We have also shown that evidence about the efficacy of drugs does not need to be limited to cases where a randomized trial exists. 

With increasing digitization of medical records, this framework has the potential to extract valuable insights from EMRs with potential to accelerate research and improve clinical workflows in many critical downstream applications, such as assisting clinicians in identifying optimal treatments based on patient characteristics, informing future clinical trial design, and facilitating efficient patient-to-trial matching \parencite{Hutson2024, Jin2024}.

Crucially, TRIALSCOPE is intended as a tool to augment and accelerate the work of clinical and statistical experts, whose critique and input are essential for robust evidence generation. Processes incorporated in TRIALSCOPE can complement existing frameworks, including established commercial RWE platforms. While existing solutions typically rely on already harmonized datasets, TRIALSCOPE addresses upstream data preparation challenges and integrates causal-inference pipelines for novel trial simulations in a transparent end-to-end workflow.



\framework also has limitations and opportunities for future development. We hope these serve as a guide for responsible use and as an inspiration for further research to expand its capabilities:

\vspace{0.5cm}
\noindent
\textbf{State-of-the-art language models:} With the recent rapid development of artificially intelligence tools language models like GPT-4 in particular \parencite{openai2023gpt4}, we envision that some components of \framework will rapidly evolve. Enhancing the automated extraction and structuring of clinical data in both accuracy and scope — and continually ensuring the highest data quality standards — will be essential for producing reliable RWE.

\vspace{0.5cm}
\noindent
\textbf{Causal modeling:} The causal modelling aspects of this work have been kept intentionally simple to increase transparency. Methods other than the Cox-PH worth exploring are time-dependent techniques \parencite{agarwal2018immortal} or those that model confounding bias in censoring mechanisms \parencite{chapfuwa2021enabling}. Additionally, although we use the HR to compare our simulations to RCT results, it is worth noting that it does not have a causal interpretation, even under randomization \parencite{hernan2010hazards}. Randomization may not hold as time progresses and the two sub-populations may become less balanced due to the loss of patients. An alternative here is to use the causal HR \parencite{Axelrod}. Interpreting results and selecting appropriate models requires deep statistical expertise.

\vspace{0.5cm}
\noindent
\textbf{Subgroup fairness}: The HR can be significant for the patients involved in a trial, but does not guarantee the uniformity of the response across sub-populations. Fairness considerations must be taken into account to ensure that no patient is left behind. This is an important problem that, although not addressed in mainstream clinical trials literature, has received recent attention \parencite{Chien}. 

\vspace{0.5cm}
\noindent
\textbf{Simulation of combination trials}: Combo trials are trials in which the treatment medication consist of a combination of drugs that are tested when they are administrated simultaneously. The simulation of combo trials requires further investigation. We find that combo trials are more challenging to simulate than single drug trials, which is consistent with \textcite{Liu2021}. 

\vspace{0.5cm}
\noindent
\textbf{Robustness and missing confounders:} All the confounders used in this work are based on previous analysis and `reasonable' factors that can simultaneously affect the assignment to the treatment and the response. However, unmeasured confounding (e.g., confounding by indication) can be a limitation in trial emulation. Direct collaboration with subject matter experts is crucial for the valid application of this tool. Research efforts to develop systematic and statistically grounded ways of identifying sources of confounding will be key for the general adoption of RWE tools. Furthermore, future work should aim to address remaining challenges inherent to RWD, such as handling time-varying confounders, assessing for potential collider bias, and exploring beyond IPW for confounder control.

\vspace{0.5cm}
\noindent
\textbf{Application in other medical domains:} Although our study focuses on oncology, the potential for \framework to be used in wider domains (e.g., autoimmune diseases, rare diseases, and non-cancer conditions) is significant. These conditions often suffer from similar challenges and could significantly benefit from robust RWE.

\vspace{0.5cm}
\noindent
\textbf{Ethical and regulatory considerations:} It is essential that RWE generation tools align with ethical and regulatory guidance for real-world clinical impact (e.g., see the FDA’s Advancing RWE Program \parencite{FDA2024}). Addressing ethical concerns is critical — this includes safeguarding data privacy and security, mitigating biases to ensure fairness, and promoting healthcare equity. 

\vspace{0.5cm}
Despite remaining challenges, we envision this work serving as a reference for framing problems and answers in RWE in a scalable and transparent manner. Ultimately, we hope that \framework can help increase transparency in the use of RWE, speeding up its adoption and improving the lives of patients and practitioners across the clinical domain.

%% file: sections/5.2_supplementary.tex
\appendix

\newpage





\clearpage

{
\centering
\Large\bfseries
Supplementary Materials
\par
}
\bigskip 

\section{Methods}\label{sec:methods_supp}

\subsection{Human subjects/IRB, data security, and patient privacy}\label{sec:dataset}
This work was performed under the guidance of an Institutional Review Board (IRB)-approved research protocol (Providence protocol ID 2019000204) and was conducted in compliance with Human Subjects research and clinical data management procedures—as well as cloud information security policies and controls—administered within Providence Health. All study data were integrated, managed and analyzed exclusively and solely on Providence-managed cloud infrastructure. All study personnel completed and were credentialed in training modules covering Human Subjects research, use of clinical data in research, and appropriate use of IT resources and IRB-approved data assets.

The patient data that we use comes from Providence Health \& Services. Providence is a health care system operating multiple hospitals across seven states in the US. They employ around 120,000 caregivers serving in 52 hospitals and 1,085 clinics. Providence is the 10th largest hospital system in the US by number of hospitals (52) \textcite{top10hospitals2022}. The Providence patient dataset used in this work consists of electronic medical records (EMR) from about 3.3 million patients. About 1 million of those patients are cancer patients. From this dataset we collected patients with advanced non-small-cell lung cancer. For each patient we extracted clinical notes, history and physical notes, treatment plans, discharge summaries, etc. For each patient we also pull semi-structured information from the inpatient billing system as well as available information at the start of the treatment time. This includes cancer lab test results, staging, tumor description, date of diagnosis, date of death, date of the last follow-up for the selected patients. Demographics and other patient characteristics like age, gender, ethnicity are extracted from the structured patient data.

\subsection{\framework data curation pipeline}\label{sec:methods_pipeline}

\begin{figure}[t!]
    \centering
        \caption{\emph{
        \textbf{B}: Structured and normalized patient data is stored in a SQL database that is regularly updated. 
        \textbf{C}: Trial specification serves as a unifying representation of research hypotheses. Trial eligibility criteria structured similarly and represented by logic statements. 
        \textbf{D}: Virtual cohort assembled akin to clinical trial matching. 
        \textbf{E}: Resulting structured real-world dataset ready for trial simulation and real-world evidence generation.
        }}
\label{fig:matching_diagram}
    \includegraphics[width=1\textwidth]{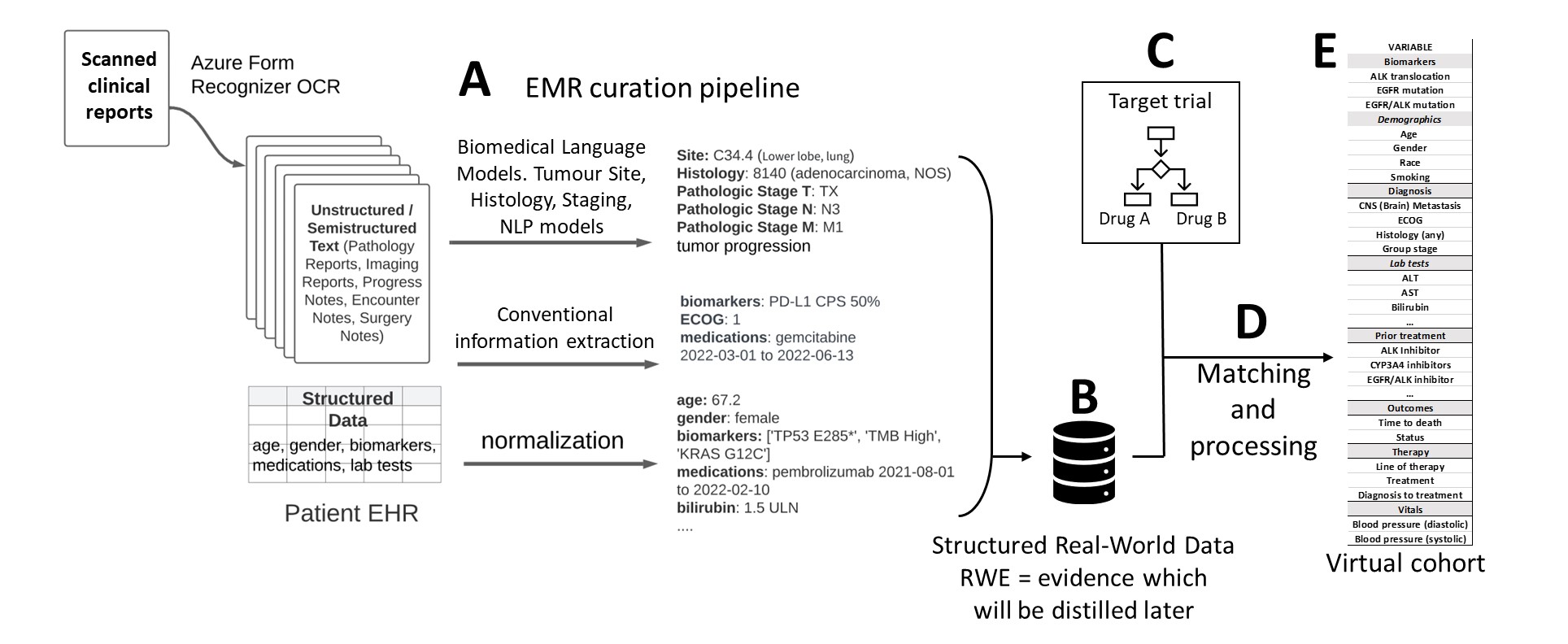}
\end{figure}

Our automatic curation pipeline for structuring EMR for real-world evidence generation is shown in a diagram in~\autoref{fig:matching_diagram}. 

Scanned reports are converted to text by state-of-the-art OCR models and clinical text is structured using a combination of biomedical language models and conventional information extraction systems. For our conventional information extraction pipeline, we use customized spaCy sentence segmentation and NLTK for tokenization. For attribute extraction, we use domain-specific rules to identify relevant entities, potential relations and determine if they are positive assertions. See Section \ref{sec:data_structuring} for details. For ECOG evaluation, we construct a gold test set by randomly selecting 565 progress notes for as many cancer patients and asked a domain expert to manually extract all performance status mentions, resulting in 77 unique patient performance status gold labels (not all notes contain performance status). To ensure generalizability, the test set is chosen to ensure no cross contamination from notes used to develop the domain-specific extraction rules. For PD-L1 biomarkers, we randomly sample 298 cancer patients and exhaustively identify PD-L1 mentions by a domain expert, including PD-L1 expression levels and the score type (combined positive score, tumor proportion score), resulting in 173 unique patient-expression level-score type relations. The test set is similarly chosen to avoid contamination from extraction-rule development. For structuring clinical notes with advanced biomedical models, we use PubMedBERT~\parencite{Gu2021} and OncoBERT~\parencite{preston2023toward}, and will explore the most recent advances such as GPT-4~\parencite{wong2023} in future work. To combat data quality challenges, we incorporate state-of-the-art probabilistic methods, e.g. \parencite{pmlr-v97-mattei19a}, for data denoising and imputation (see Section \ref{sec:missing_data}). Details of how we perform attribute structuring is outlined in Section~\ref{sec:data_structuring}. Structured and normalized patient data is stored in a SQL database that is regularly updated. 

Trial specification serves as a unifying representation of research hypotheses. For details on how we identify trials to simulate in this work, see Sections~\ref{sec:trial_selection}. Trial eligibility criteria are structured similarly and represented by logic statements. A virtual cohort is assembled akin to clinical trial matching (see Section~\ref{sec:study_cohort}). This resulting structured real-world dataset is then ready for trial simulation.

\subsection{Model evaluation}\label{sec:curation_evaluation}

To evaluate our language model pipeline, we have sampled 565 cancer patients who had active cancer after that date and selected one progress note per patient within nine months after that date. An expert manually annotated these progress notes and labeled 79 notes with an explicit performance status score. The average word length per progress note in this test set is 1015.2 word tokens. Section \ref{sup:curation_performance_data} summarizes the approach taken to extract each group of attributes (attribute-level details given when needed) as well as the ground truth data used to validate the extraction. See Section \ref{sup:data_schema} for further information on the data schema.


\subsection{Trials selection}\label{sec:trial_selection}

\begin{figure}[h!]
\caption{\emph{
Selection process for the clinical trials used in our simulation. We included all seven trials explored by \textcite{Liu2021}. Additionally, we conduct an exhaustive search of \texttt{ClinicalTrials.gov} with eligible trials for advanced non-small cell lung cancer, with ``other filters'' to only include standard two-arm trials with active control and sufficient real-world instances in our Providence dataset.
}\label{fig:trial_selection}}
\vspace{0.5cm}
\centering
\includegraphics[width=13.5cm]{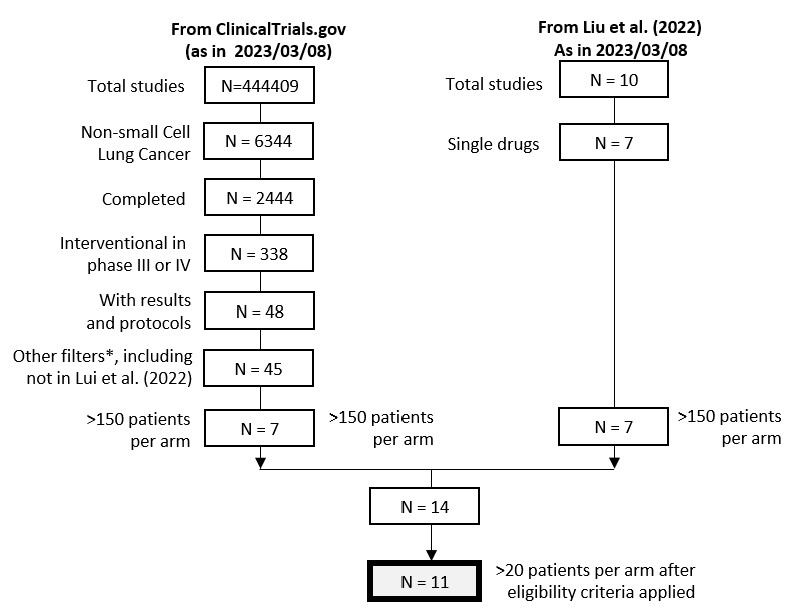}
\end{figure}

For our experiments, we follow a systematic approach to select the trials to simulate, using \parencite{Liu2021} as a reference (see \autoref{fig:trial_selection} in Section~\ref{sup:trial_selection}). We consider for simulation all single-drug trials with at least 150 patients in both arms before any data filtering is applied published in this work (7 trials). We augment this set by conducting a search on \texttt{ClinicalTrials.gov}. Since this is an evolving dataset, we fix our search to the 2023/03/08 and we use the following filters: non-small cell lung cancer trials (6344 studies), completed studies (2444 studies), interventional in phase III or IV (338 studies), with study protocols (48 studies) and have two arms with different treatments with available reports and not already in \textcite{Liu2021} (45 studies). For consistency, we also select trials with at least 150 patients in both arms before data processing. This leads to a total of another 7 extra eligible trials. 3 of the total of 14 trials are not considered in the analysis due to having less that 20 patients per arm after applying the eligibility criteria.


\subsection{Study cohort selection}\label{sec:study_cohort}

We build a cohort of patients per trial. We select all the patients that resemble as much as possible a population that could have been eligible for each trial while maximizing the number of patients per trial. The database contains a total of one million patients, 18,038 of which received the control or target treatment in at least one of the analyzed trials. Patients are also selected according to the line of therapy of the trials. If the line of treatment is missing, this criteria is ignored and the patient is added to the trial as in \textcite{Liu2021}.

Patients are included in the analysis if they were diagnosed with lung cancer (as per the international classification of diseases and had pathology consistent with NSCLC) with stage IIIA, IIIB, IIIC, IVA or IVB. 



Patients with inconsistent start diagnosis and death dates are removed as well as those individuals with more that two years between diagnosis and treatment. We remove all treatment patients who took the control drug during ``trial'' and the control patients who took the treatment drug. We also remove any duplicated patients with other data inconsistencies. Patients for which the event (death or last visit) has been recorded in a time beyond the duration of the trial are corrected and considered censored with a time event equal to the duration of the trial. Data processing is done in \texttt{python} using \texttt{pandas}.



\subsection{EMR structuring and representation}\label{sec:data_structuring}

The choice of the method to structure each attribute in the dataset depends on the availability and quality of structured data, the difficulty of the task, and the importance of the data for our analysis. When the same information is available from several sources, we select the value extracted using the method with higher confidence, which is typically the structured source.


\vspace{0.5cm}
\noindent
\textbf{Patient Demographics:}
Patient demographics refer to the date of birth, gender, race, ethnicity, vital status, last contact date, and death date. This information comes structured from the hospital's internal records. However, for the death date, we extract this information from both the hospital records and social security death records. If a patient's death information is not found, we extract the last contact date by using the latest date from all records for that patient.

\vspace{0.5cm}
\noindent
\textbf{Oncology:}
For oncology attributes (site, histology, staging, diagnosis date), we combine structured data from the cancer registry and predictions from self-supervised large language models \parencite{Preston2022}. The LLM model inputs include pathology reports, progress notes, imaging reports, encounter notes, diagnosis code description, Op Notes, and Surgery notes. The model predicts ICD-O-3 site and histology codes, which we then map to OncoTree IDs for clinical trial matching. To predict diagnosis date, we use the case-finding model only on pathology reports.

\vspace{0.5cm}
\noindent
\textbf{Biomarkers:}
Biomarkers are obtained in structured form but requires normalizing to three fields: gene/protein, variant, and variant type. For variant, we try to normalize to HGVS nomenclature when possible. However, some biomarkers are only available to us in unstructured form, such as pathology reports that contain third-party laboratory test results. For PD-L1 IHC expression results, we utilize an information extraction pipeline that contain 3 steps: entity extraction, relation extraction, and intent detection. The entity extraction step will extract PD-L1 test names such as combined positive score (CPS), Tumor Proportion Score (TPS). We also test result values such as negative, positive, high, low, a specific percentage value (e.g., 10\%), or a range of percentage values (e.g., $>50\%$). We then apply relation extraction to determine if there is a relation between that PD-L1 test and the test result value. If so, we then classify the intent of that relation as to whether or not this patient actually has that biomarker relation. Oftentimes, patient notes may contain biomarker entities but the patient may not actually have that biomarker. The text may contain a description of the biomarker test, writing about a hypothetical situation, or negation. We do not extract the date of the PD-L1 measurement mentioned in the note but use the note date for the PD-L1 measurement. In future work, we will also extract the date associated with each PD-L1 measurement.

\vspace{0.5cm}
\noindent
\textbf{Treatment:}
Information about patients' medication is extracted from both structured and unstructured medical records. We use two sources of structured medication information: ``Ordered-Meds'' and ``Administered-Meds''. Ordered-Meds contains all medications ordered for a patient which may or may not have been administered to the patient. Administered-Meds are all medications that are given to the patient at the hospital and are of high confidence. Using medication descriptions from the structured data, we standardize each medication to the NCI (National Cancer Institute) Thesaurus concept ID. This standardization allows for a seamless use of drug synonyms, abbreviations, and  collective names interchangeably. We rely on our medication extraction module to complement the structured medication information with data extracted from free clinical notes since not all patients have structured medication information and even those with structured medication information may not have a complete structured medication history. Clinical notes contain more comprehensive and detailed patient drug information but such information is buried in large unstructured formats and not easily accessible. Our treatment extraction pipeline to extract medication information from free clinical texts has the following modules:
\begin{itemize}
    \item Extract medications: The first step in the pipeline, extracts medication mentions. These can be in short-forms, medication codes or any of the drug synonyms.
    \item Extract attributes: For each of the extracted medications, we find attributes from the span. The most important ones are dosage (amount of a medication used in each administration), frequency (how often each dose should be taken), mode (route for the medication), date (date of medication administration) and discontinuation/Substitution (whether the medication is still active or being discontinued/substituted)
    \item Link Entities: Here, each of the extracted attributes are linked to their corresponding medication
    \item Determine Administration: Using all the information above, this module determines whether the medication mentioned in the clinical note is administered to the patient or not. This is a challenge since medications can be mentioned in a clinical note as a suggestion, in reference to past history, or in a hypothetical framing.
\end{itemize}

Using the treatment extraction pipeline and the structured medication information, we obtain a complete treatment information about a patient. Each patient record is enriched with details such as the list of all medications taken with dose, frequency and mode, date of each administration, and when substitutions or discontinuations occur in the patient timeline.

\vspace{0.5cm}
\noindent
\textbf{Line of Therapy (LoT)} Determining the LoT is critically important to assess a patient's eligibility for a given clinical trial. However, there are not any universally accepted set of criteria to enumerate LoT. To alleviate this, we have adopted the guidelines suggested in \textcite{saini2021determining} and \textcite{meng2021automated}. The following is the final guideline we have followed to determine LoT in our cohort of patients:
\begin{enumerate}
    \item The first LoT is defined as the first SACT(systemic anti-cancer therapy) drug recorded after date of diagnosis.
    \item If clinical progression of disease is documented, assign a new LoT to the next SACT.
    \item If a SACT drug is discontinued and substituted by another drug of the same class, retain the same LoT.
    \item If one or more new anti-cancer agent is added to an ongoing SACT, assign a new LoT.
    \item If one or more anti-cancer agent is discontinued from an ongoing SACT, retain the same LoT for the remaining anti-cancer agents.
\end{enumerate}

\vspace{0.5cm}
\noindent
\textbf{ECOG}
Performance status is mentioned only in the unstructured free text, particularly progress notes. We extract various metrics mentioned such as Eastern Cooperative Oncology Group (ECOG), Karnofsky Performance Status Scale (KPS), Lansky Performance Status Scale, and Palliative Performance Scale (PPS). We map these different metrics to ECOG for ease of comparison. This is extracted using our information extraction module similar to the PD-L1 extraction pipeline described above. We convert all the extracted metrics to ECOG.

\vspace{0.5cm}
\noindent
\textbf{Smoking history}: Smoking history or status for patients was extracted from available data about patient diagnosis, represented by ICD-10 codes. As shown previously by \parencite{wiley2013icd9}, the specificity of using ICD-9 codes is high, ``indicating the exceptional utility of these codes for identifying true smokers'' and supporting ``the use of these codes for the identification of smokers for clinical studies". However, using NLP on clinical notes combined with the ICD-9 codes results in higher sensitivity, as discussed by \parencite{wang2016comparison}, which we will explore in future work.
These ICD-9 codes (305.1 Tobacco use disorder, and V15.82 Personal history of tobacco use) were converted to ICD-10, by including more specific codes that represent ``tobacco", ``smoking", ``nicotine", ``cigarettes" use/abuse while filtering out instances of explicitly non-smoking use like ``chewing tobacco”. Other codes that indicate use or abuse of tobacco currently or in the past, in addition to chronic exposure of environmental smoke were also included.

\vspace{0.5cm}
\noindent
\textbf{CNS metastasis}: Patients with Central nervous system (CNS) metastasis, also referred to as brain metastasis in our dataset, are collected by filtering for ICD-10 code C79.31 (Secondary malignant neoplasm of brain) from patient history. \textcite{eichler2009utility}
reports that the code for secondary brain or spinal cord neoplasms, 198.3 (converts approximately to ICD-10-CM C79.31), had good recall (sensitivity), precision (positive predictive value), and specificity for identifying patients with brain metastases. The precision and specificity increases when the code recurred on different days.

\vspace{0.5cm}
\noindent
\textbf{Laboratory tests}: 
Laboratory test results are typically found in structured form (between 69.3 \% to 76.9 \%) and thus we do not extract this from the unstructured text, even though it is commonly mentioned. We filter commonly encountered lab test names for each lab test type from our dataset, and categorize each test into one of three standard units: upper limits of normal (ULN), grams per deciliter (g/dL), and count per microliter (/muL). We then convert the values to required units. To calculate ULN we take the value of the lab and divided it by the higher limit of normal reference range. Each lab test has between ~2.7 million to ~4.99 million entries, in our final dataset. We also convert other fluid measurement units to g/dL or count/microliter.



\subsection{Missing values}\label{sec:missing_data}

We use a hybrid approach to deal with missing data. This is due to the structure of the missingness and their numbers across the dataset. In the structured dataset there are missing values in the line of therapy, start, diagnosis and end date and in all of the lab tests. The missing values in the laboratory test variables (ALT, AST, hemoglobin, etc.), smoking and CNS metastasis are imputed using an in-house implementation of the approach by \textcite{pmlr-v97-mattei19a}. The line of therapy is missing in 90\% of the patients. Following \textcite{Liu2021}, we only filter it in patients in which it is available. Patients with missing or incoherent dates are removed from the dataset. 

We deal with the missingness of categorical variables via one-hot encoding. All the data included in the trials in this study are imputed simultaneously. The dimension of the latent variable is fixed to six. The width of the multi-layer perceptron in the variational auto-encoder is fixed to 32 and the decoder and we use a depth of 3 layers with ReLU activation functions. The training is carried out for 500 epochs and we use 100 samples for estimating the distribution of the unobserved data given the observed data. The imputation of the missing data is done as described in \textcite{pmlr-v97-mattei19a}.

\subsection{Causal survival analysis}\label{sec:causal_modeling}

\textbf{Survival outcomes and treatments}: Survival outcomes are defined as $(Y_i, D_i)$. $D_i \in \{0, 1\}$ represents the death event based on the follow up of the patient. $Y_i \in \mathcal{Z}^+$ represents the observed survival time, in days, since the start of the therapy. $Y_i$ is computed as $Y_i = min (T_i, C_i)$ where $T_i$ is the time of death and $C_i$ is the censoring time. To make the simulation of the trials as accurate as possible we use the duration of the trial as censoring time for those patients with a larger duration. When the event \emph{death} is observed, $Y_i$ accounts for the number of days between the start of therapy and the event. When death is not observed, the last contact date with the patient is used and considered censored. The treatment that the patient receives is denoted by $W_i\in \{0,1\}$. This is a binary variable because we only consider two-arm trials in this work but this can be generalized further.

\vspace{0.5cm}
\noindent
\textbf{Covariates selection}: The covariates of the model are described by $X_i$ and include age, gender smoking, histology, CNS metastasis, ECOG score, race, hemoglobin, lymphocites, ALT, AST, ALP and the days between diagnostics and treatment. In cases in which a covariate is part of the eligibility criteria of the trial, these are not included in the analysis. 

\vspace{0.5cm}
\noindent
\textbf{Assumptions}: Simulating a trial requires some assumptions that we make explicit here. We follow the standard potential outcomes point of view and we refer to \textcite{Imbens2015} for further details. We also recommend \textcite{Dahabreh2022} for a review on assumptions combining experimental and observational studies.

\begin{enumerate}
    \item \emph{Conditional exchangeability}, $Y({W=w}) \perp W| \textbf{X} $, for $a=0,1$ where $Y({W=w})$ is the potential outcome of an individual assigned to group $W=w$. This assumption specifies that the right confounders have been observed, which implies that conditioning on $X$ the outcome of the treatment of an individual once it has received a treatment is independent of the selection mechanism (a randomized experiment can be simulated).

    \item \emph{Positive support}, $P(X) > 0$. All eligible individuals can be selected for the study.

    \item \emph{Overlap}, $1> P(W =1 |X) >0)$. All the eligible individuals have a non-null probability of being selected as cases and controls.

    \item \emph{Stable Unit Treatment Value Assumption (SUTVA)}. There is no interference between patients. The SUTVA assumption consists of two assumptions: first, the potential outcomes for any patient do not vary with the treatments assigned to other units (patients are independent in their responses), and second, for each unit, there are no different forms or versions of each treatment level that lead to different potential outcomes (the treatment is always the same). Note that SUTVA does not say anything about how the treatment assigned to one unit affects the treatment assigned to another unit.

    \item \emph{Trial population match}, $P_{RCT}(X) = P_{RWD}(X)$. When the goal is to simulate the result of a previous trial, the distribution of the confounders in the observational study and in the trial should be the same.
\end{enumerate}

Assumptions 2-5 can be tested from observation data. Assumption 1 cannot, but we can find indications in the data that the confounders in the data are removing the correct biases. See below for a list of data-driven tests that are used the validate these hypotheses, to the extent of what is possible with observational data.

\vspace{0.5cm}
\noindent
\textbf{Confounding bias corrections}:
To evaluate the balancing between treatments and control we compute the propensity score (PS). The PS of a patient $i$ is defined as
$$e(X_i) = \mathbb{P}(W_i | X_i)$$
It captures the probability taking the treatment $W_i$ in the presence of the confounders $X_i$. In balanced randomized experiments in holds that $e(X_i) = 0.5$ for all patients \parencite{rosenbaum_rubin_2006}. The propensity score is a balancing score. This means that by conditioning on the propensity score it is expected for the distribution of observed covariates to be the same in the treatment and the control groups. Propensity scores are commonly used to reduce the bias due to confounding in observational studies by re-weighting the observations in the model of the outcomes. The weights of individuals with a propensity score close to 1 or 0 is reduced. These are individuals whose assignment is identified to be heavily affected by the presence of the values of the confounding variables. On the other hand, individuals with propensity score close to 0.5 are over-weighted because their assignment is independent of the covariates, as it is the case in RCTs where individuals are assigned randomly to cases and controls. In the \framework codebase, the propensity scores are computed with a logistic regression using \texttt{scikit-learn}\footnote{https://scikit-learn.org/}. 

\vspace{0.5cm}
\noindent
\textbf{Survival models}: The hazard rate of an individual, denoted usually by $h(t| X, W)$ is the rate in which that a patient will die at time $t$ in the presence of the covariates $X$ and treatment $W$. In survival analysis, the Hazard Ratio for the treatment is defined as
\begin{equation}
HR (t) = \frac{h(t| X, W=1)}{h(t | X, W=0)},
\end{equation}
which accounts for the relative risk of death at time $t$ between the treatment and control groups \textcite{hernan2010hazards}.

We use the Cox proportional hazards model (Cox-PH) \textcite{cox1972regression} to perform the survival analysis. This model assumes that
\begin{equation}
h(t| X)={h}_{0}(t)\exp \left({b}_{w}W+\mathop{\sum }\limits_{j=1}^{p}{b}_{j}{X}_{j}\right),
\end{equation}
where ${h}_{0}(t)$ is known as the baseline-hazard, ${b}_{w}$ is the parameter accounting for the effects of the treatment and ${b}_{j}$ is accounting for the effect of the $jth$  covariate. The particular factorization of Cox-PH allows to associate each parameter of the model with the HR of each covariate. In particular, the HR for the treatment is, 
$$HR = \exp{(b_w)},$$
which is independent of the time due to the structure of the Cox-PH model. In our experiments we use the the class \texttt{CoxPHFitter} from the library \texttt{lifelines}\footnote{https://lifelines.readthedocs.io/en/latest/}. We use two different approaches to compute the Hazard Ratio:
\begin{itemize}
    \item Unadjusted Cox proportional hazard (\textbf{CoxPH-U}): This model is used as baseline. No confounding correction is used. We use the target treatment $W$ as covariates. The reported HR is the exponential of the target treatment parameter.
    \item Cox proportional hazard with inverse propensity re-weighting (\textbf{CoxPH-IPSW}): We compute the hazard ratio using a Cox-PH model trained on $W$ using a inverse propensity treatment weighting \parencite{COLE200445}. The weights for each patient are obtained as
$$ w_i = \frac{W_i}{e(X_i)} + \frac{(1-W_i)}{1-e(X_i)}  $$ 
\end{itemize}

\vspace{0.5cm}
\noindent
\textbf{Simulation validation tests}: For each trial we perform a series of tests to guarantee the stability and coherence of the results. Causal assumptions like the conditional exchangeability cannot be tested with observational data only. However, a sensitivity analysis of the HR with respect to several varying aspects in the data can help to identify potential issues. In our study we perform the following analysis:

\begin{itemize}
\item \textbf{Random Confounder}: We add a simulated confounder to the problem by randomly generating data from a Gaussian variable with mean zero and variance $\theta$. We recompute the $HR$ for a grid of 10 values on $\theta \in [0.1, 5]$. Each experiment is repeated 300 times. \emph{Expected positive test result}: No statistical variation of the hazard Ratio across different values of $\theta$. This test aims to test the stability of the chosen confounders.
\item \textbf{Placebo treatment}: We randomly permute the assignment to the treatment. We repeat the experiment 300 times. \emph{Expected positive test result}: The newly computed Hazard ratios are not statistically significant ($HR=1$). This test aims to validate that the quality of the signal.
\item \textbf{Downsampling}: We select samples of decreasing sizes with replacement and we recompute the HRs. We take subsamples of size $90$, $75$, $50$ and $25\%$ of the original dataset. We repeat each experiment 30 times. \emph{Expected positive test result}: Same average HR in each scenario with an increase in variance when the sample size decreases. This test aims to capture the stability of the results and its consistence under random variations in the sample.
\end{itemize}

Beyond indications for potentially missing confounders, it is important to evaluate that given the chosen confounders the balancing between the groups has been carried out correctly and that the structure of the data allows for causal analysis. To evaluate this issues, we perform the following tests:

\begin{itemize}
\item \textbf{Groups rebalancing}: To test for the quality of the signal, and how imbalances between groups are corrected when using the IPSW approach, we compute the standard mean difference for all the confounders with and without re-weighting for the treatment and control groups. \emph{Expected positive test result}: Re-weighting makes the differences between the treatment and control groups close to zero in all confounding variables. 

\item \textbf{Overlapping sets}: We test that all patients have a non-negative probability of being assigned to either of the two groups. We use the method described in \textcite{oberst2020characterization} and the available in \texttt{dowhy}\footnote{https://github.com/py-why/dowhy} \parencite{sharma2020dowhy}. \emph{Expected positive test result}: A large percentage of the same ($>95\%$) satisfies the condition. This test aims to capture the overlap of cases and controls (the more overlap we have the better the data are for causal analysis).
\end{itemize}

The HR is a quantity that describe a population of individuals characterized by $P(X)$. Therefore, for the HRs in the RCT and the observational to be comparable it is needed that $P_{RCT}(X) = P_{RWE}(X)$. Filtering patients in the observational dataset according to the eligibility criteria only guarantees that the support of these two densities are the same so further validation is needed. Unfortunately, data from the RCT are rarely available. To test this assumptions we use the summary statistics that we published together with the RCTs and we compare those with the equivalent values in the structured EMRs.

\subsection{Speed and cost estimation}\label{sec:speed_cost_method}

To estimate the speed for \framework, we run the automatic extraction pipeline on 48 patients to extract 10 attributesto obtain the average total time taken per patient. The cost is based on the running cost of our compute resource, 100x VM D12 v2 - 4 vCPU and 28 GB RAM. To estimate values for traditional curation, we internally conduct manual curation for 44 patient records for 10 attributes, and use an example of \$29 the hourly pay for a certified cancer registrar in the US.

\section{Data and code availability}

The EMR data for this study are not made publicly available due to privacy and compliance considerations established by the research protocol. The EMR data cannot be redistributed beyond those who have obtained a Material Transfer Agreement. Instead, we have provided a detailed description of the data selection pipeline and processing in Section \ref{sec:data_structuring}. Queries about these data should be directed to the corresponding authors indicated above.
The biomedical language models and its pretraining algorithm are detailed in \textcite{Gu2021} and are publicly available\footnote{\url{https://aka.ms/pubmedbert}}.
 

\clearpage
\section{Patient attributes for curation performance validation}\label{sup:curation_performance_data}

\begin{table}[h!]
    \centering
    \small
    \begin{tabular}{ll|ccc|c}
    \hline
        & &  \textbf{Structured} & \textbf{Conventional} & \textbf{Biomedical}  &  \\
        & &  \textbf{data} & \textbf{information} & \textbf{language}  & \textbf{Ground truth} \\
        & &  \textbf{pipeline} & \textbf{extraction} & \textbf{models}  &  \\
       \hline\hline
       \textbf{Biomarkers}        &    &   & &  & Expert labelled data \\
           & \textit{All (see Sup. Mat. \ref{sup:data_schema})}   &  \checkmark & \checkmark &   & -\\ 
        \hline
    
       \textbf{Demographics}  &  &  & &  &\\
                  & \textit{All (see Sup. Mat. \ref{sup:data_schema})}    &  \checkmark &   &   & - \\ 
        \hline
       \textbf{Diagnosis} &  &  & &  &\\
       & \textit{CNS Metastasis}  & \checkmark  & &  & Diagnosis code\\
       & \textit{ECOG} &  & \checkmark  &  & Expert labelled \\ 
       & \textit{Histology} & \checkmark & & \checkmark & Cancer Registry\\
       & \textit{Staging (Clinical)} & \checkmark & & \checkmark  & Cancer Registry\\
       & \textit{Staging (Pathologic)} & \checkmark & & \checkmark  & Cancer Registry\\
       & \textit{Tumor site} & \checkmark & & \checkmark  & Cancer Registry\\
       & \textit{Diagnosis date} & \checkmark & & \checkmark  & Cancer Registry\\
    \hline
       \textbf{Lab tests} &  &  & &  &\\ 
                  & \textit{All (see Sup. Mat. \ref{sup:data_schema})}    &  \checkmark  &   &    & Hospital lab results\\ 
        \hline
       \textbf{Pre-treatment} &  &  & &  &\\ 
                  & \textit{All (see Sup. Mat. \ref{sup:data_schema})}   &  \checkmark  &  \checkmark  &   & Expert labelled data\\ 
        \hline
       \textbf{Outcomes} &  &  & &  &\\ 
             & \textit{All (see Sup. Mat. \ref{sup:data_schema})}   &  \checkmark  &   &   & - \\ 
        \hline
        \textbf{Treatment} &  &  & &  &\\ 
            & \textit{Line of Therapy}  &  & \checkmark   &    & Guidelines\\ 
            &\textit{Medication} &  \checkmark  &  \checkmark &   &  Expert labelled data\\ 
             & \textit{Start date}   &  \checkmark  & \checkmark   &    &  Expert labelled data\\ 
             & \textit{End  date}   &  \checkmark  & \checkmark   &    &  Expert labelled data\\ 
        \hline
        \hline
    \end{tabular}
    \caption{\emph{
    Patient attributes used in this work, with structuring provenances, and reference ground truth data for development validation. The tick mark in the ``Structured data pipeline'' column indicates that the corresponding structured attribute exists in EMRs, although in practice its value often may not be populated.
    }}
    \label{table:data_sources}
\end{table}

\clearpage
\section{Trial selection process}\label{sup:trial_selection}


\clearpage
\section{Database schema}\label{sup:data_schema}

The file \texttt{DataBaseSchema.xls} contains the details of all the variables used in this work, their type, description and, when relevant, the units in which they have been recorded.






\clearpage
\section{Oncology tree for advanced non-small-cell lung Cancer types }\label{sup:oncotree}

\begin{figure*}[!h]
    \centering
        \caption{\emph{Oncology tree of the types of Cancer that we have considered in this work. \emph{Source: } \texttt{http://oncotree.mskcc.org/}}}
\vspace{0.5cm}
    \label{fig:oncotree}
    \includegraphics[width=22.5cm, angle=90]{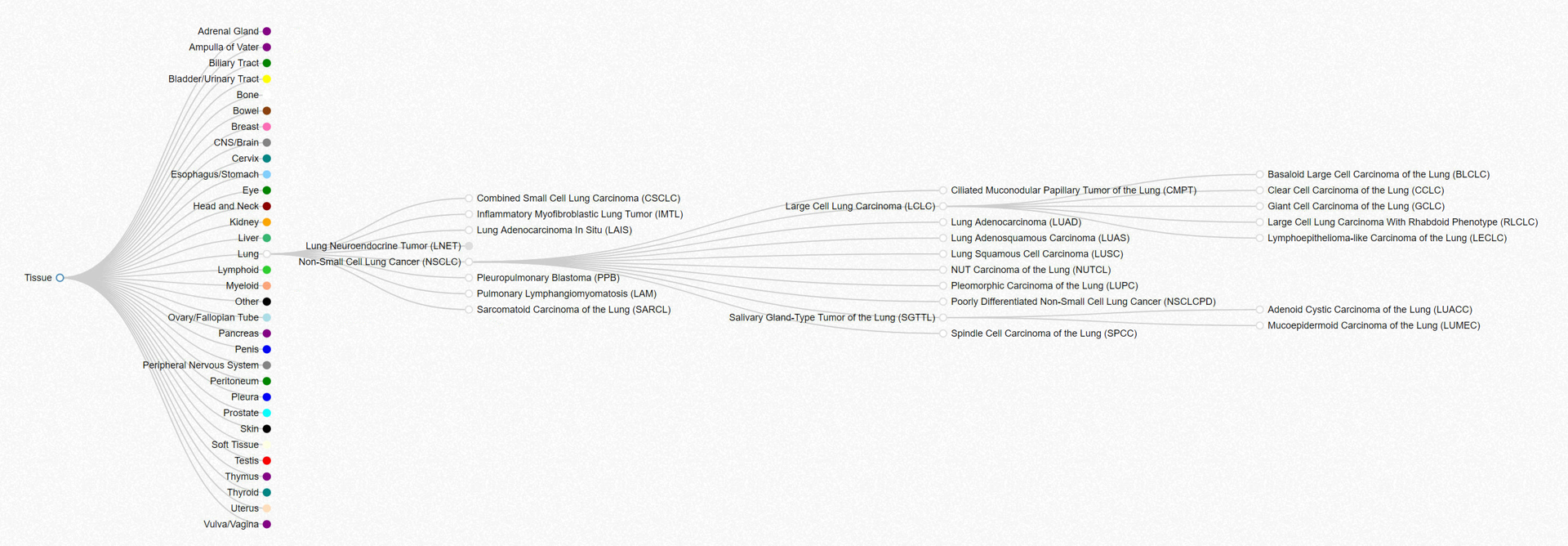}
\end{figure*}

\clearpage
\section{Missing data imputation }\label{sup:missing}


\begin{figure}[h!]
\caption{\emph{Imputation results for the KEYNOTE10 trial. \textbf{A}: Heatmap of the values extracted from the laboratory text. In this dataset there is ~35\% of missing data in these variables. \textbf{B}: Imputed dataset using a latent variable model. See Section \ref{sec:methods} for details. \textbf{C}: Validation of the imputation approach. Once the matrix of laboratory test has been imputed we artificially inject missing values in the same proportion to those originally in the dataset and we repeat the reconstruction procedure. Horizontal axis contains the true laboratory tests values of the missing (log scale). Vertical axis contains imputed values (log scale).}}
\centering\includegraphics[width=17cm]{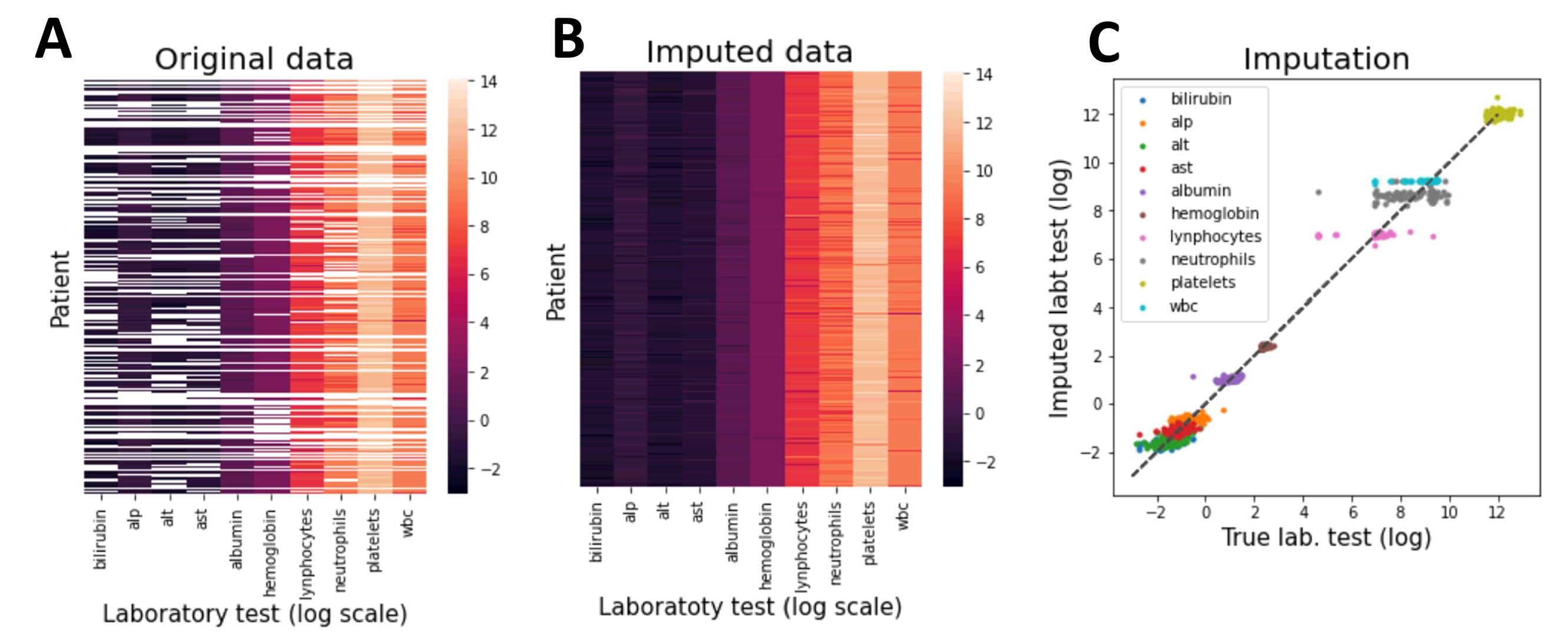}
\end{figure}\label{fig:inputation_keynote10}

\clearpage
\section{Patients filters}\label{sup:patients_filters}

\begin{figure}[h!]
\caption{\emph{Percentage of patients filtered by each eligibility  criteria in all the selected trials. This figure contains the filters over the originally 18 trials selection following the selection process in Figure \ref{fig:trial_selection}.}}\label{figure:patients_selection}
 \vspace{0.5cm}
\centering
\includegraphics[width=16cm]{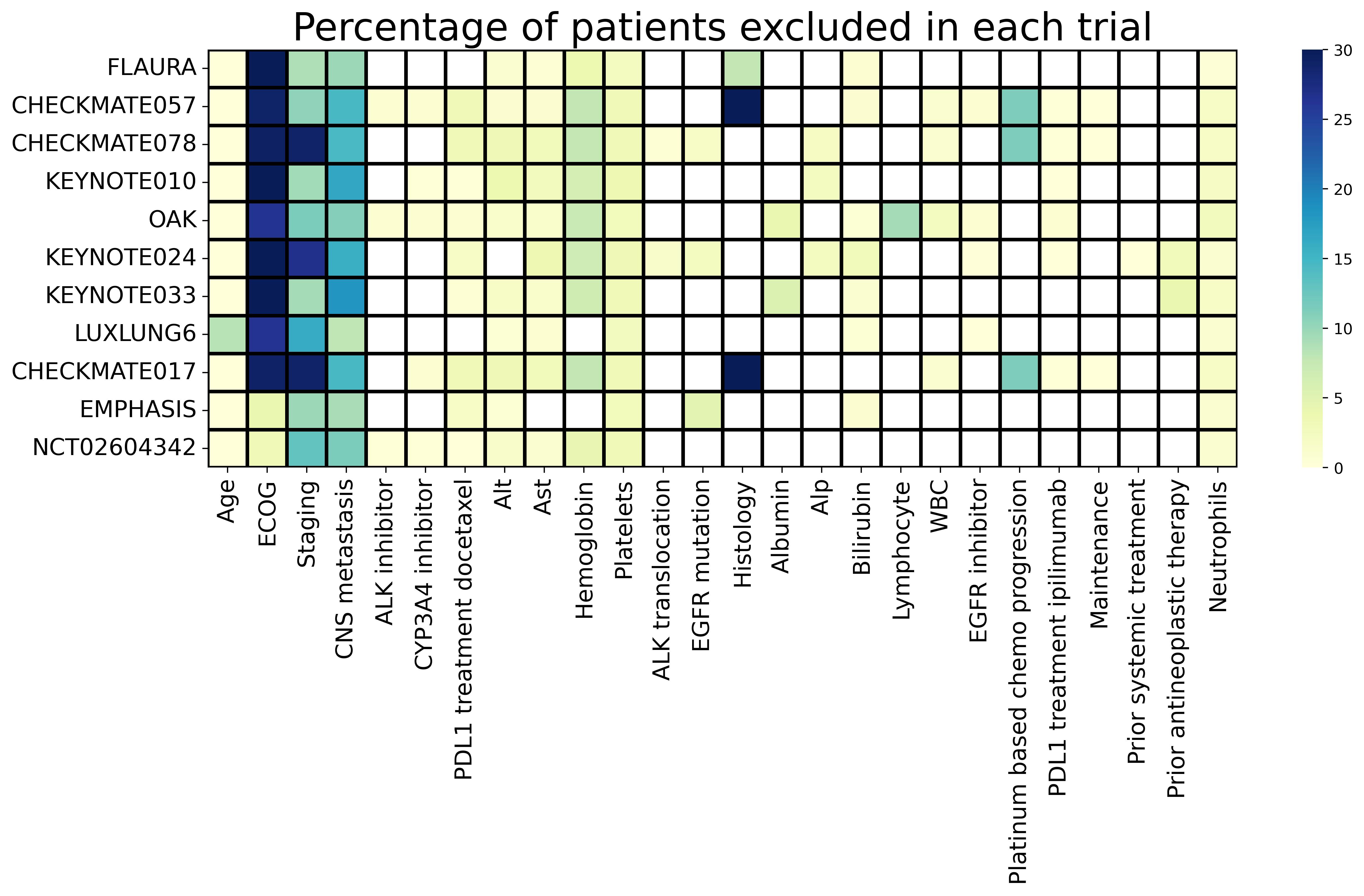}
\end{figure}

\clearpage
\section{Test results for automatic curation}\label{sup:test_results_curation}

\begin{table}[h!]
\centering
\small
\begin{tabular}{c|c |c| c| c} 

 \hline
 \textbf{Biomedical language model pipeline} & \textbf{AUPRC} & \textbf{AUROC} & \textbf{Accuracy} & \textbf{Problem Formulation} \\ 
 \hline
\textbf{Group stage (clinical)}  &91.7 & 96.6 &85.5  &  Multi-class classification\\ 
\textbf{Group stage (pathologic)} &93.7 & 96.8 & 87.8 &   Multi-class classification\\ 
\textbf{Tumor Site} & 76.7 & 99.3 & 69.1 &  Multi-class classification \\ 
\textbf{Histology}  &87.2 & 99.4 & 81.2 &   Multi-class classification\\ 
  \textbf{Diagnosis date (case finding)} & - & - & 97.3 & Binary classification \\
 \hline \hline
 \textbf{Information extraction pipeline} & \textbf{Precision} & \textbf{Recall} & \textbf{F1} & \textbf{Problem Description} \\ 
 \hline
 \textbf{Medication and Prior T.} & 92.3 & 87.1 & 89.6 & Binary classification \\
 \textbf{ECOG} & 100.0 & 96.2 & 98.1 & Relation Extraction \\
 \textbf{PD-L1 Biomarker} & 97.5 & 89.6 & 93.4 & Relation Extraction \\
 \hline

\end{tabular}
\caption{\emph{
Cancer patient attributes used in this work, with test results for automatic curation in \framework.
\textbf{Top}: Test results for biomedical language models from \textcite{preston2023toward}.
\textbf{Bottom}: Test results for conventional information extraction. 
}}
\label{table:processing_results}
\end{table}

\clearpage 
\section{Effects of confounder correction}\label{sup:experiment_confounder_correction}

\begin{figure}[h!]
\centering
\caption{\emph{Effect of correcting the simulators by confounders in all the trials. We show the absolute difference between the HR in the original trial and the estimation with and without IPSW. The estimation of the HR benefits or remain the same in all the trials when confounders are added to the models.}}\label{fig:data_sources}
\centering\includegraphics[width=8cm]{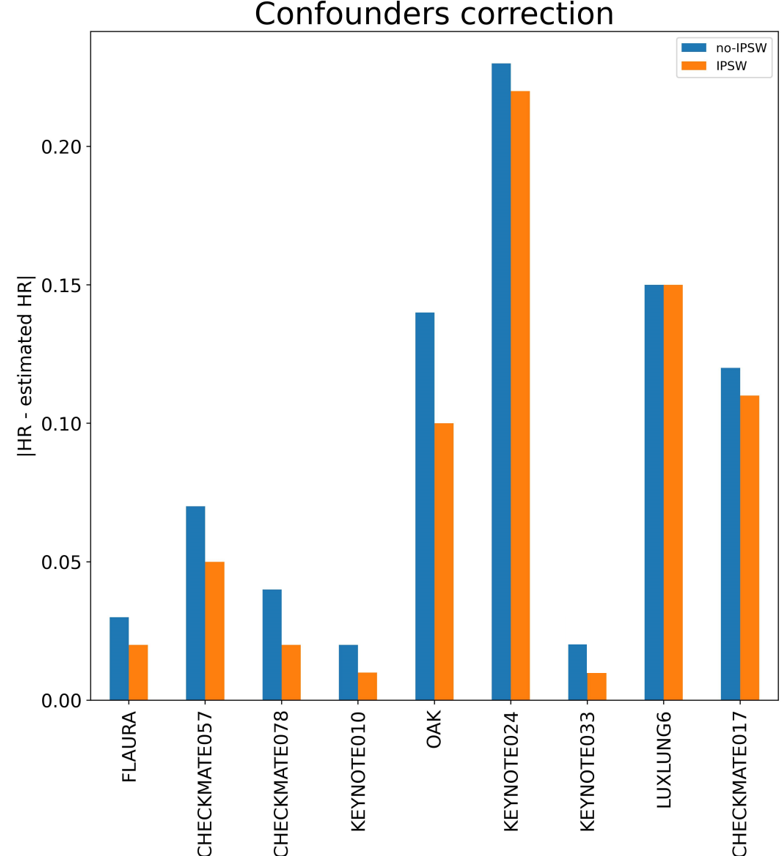}
\vspace{0.5cm}
\end{figure}

\autoref{fig:data_sources} shows the differences for the absolute differences between the RCT HRs and the estimated HRs with and without the confounders correction in the Cox-PH model, when all the eligibility criteria are applied (see Section \ref{sup:results} for simulation results without applying the correction). Correcting for confounders moves the estimation of the HR towards the ground truth values in 8 of 9 cases with strong corrections in CHECKMATE057 and OAK. This demonstrates that this procedure has a considerable effect in offsetting the confounding bias present in EMR. This is consistent with those of \textcite{Zeng2022}, which shows that confounding correction reduces bias in survival models built with EMR.

\clearpage
\section{Trial arm cohort size}\label{sup:drugs}

\begin{figure}[h!]
\caption{\emph{Summary of the number of patients taking each drug (treatment arm) in the database}}\label{figure:drugs}
 \vspace{0.5cm}
\centering
\includegraphics[width=10cm]{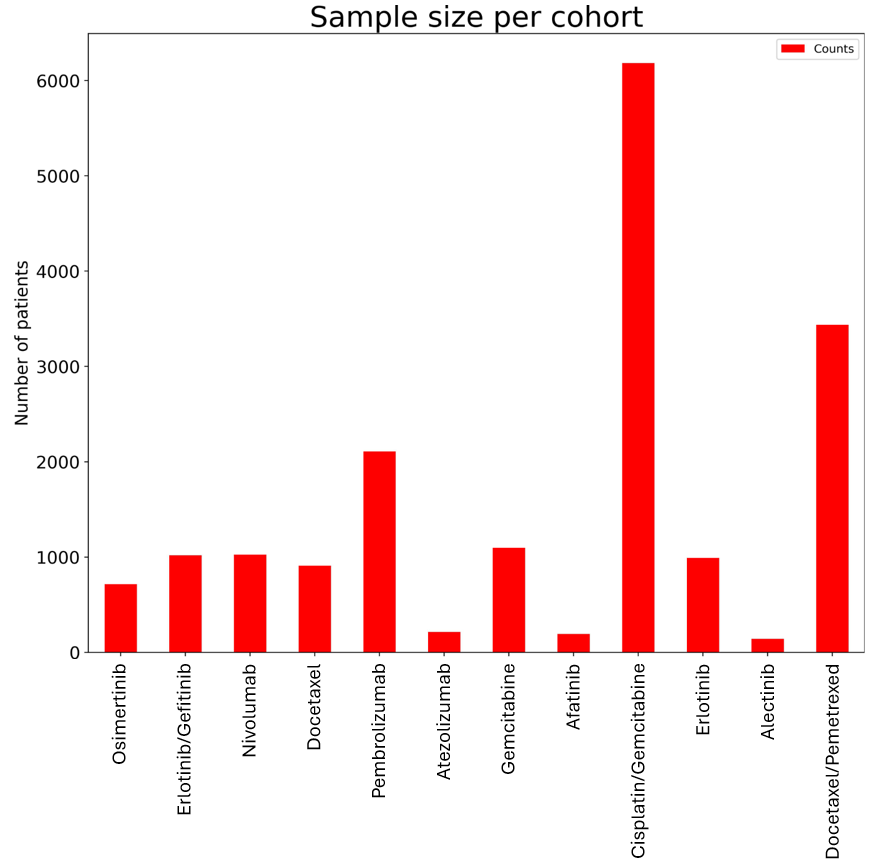}
\end{figure}

\clearpage
\section{Simulation results with no confounding correction}\label{sup:results}

\begin{table}[h!]
\centering
\vspace{0.5cm}
\begin{tabular}{ll|cc|cccc} 
\hline
\multicolumn{2}{c|}{\textbf{Trial name and dataset}} & \multicolumn{2}{c|}{\textbf{RCT}} & \multicolumn{4}{c}{\textbf{Simulation}} \\
& &  \textbf{HR}& \textbf{95\%CI} &  \textbf{HR}& \textbf{95\%CI} &  \textbf{C} & \textbf{T}\\
\hline \hline
FLAURA & \textbf{Yes} & 0.63 & (0.45, 0.88) & 0.59 & (0.45, 0.76) & 268 & 200\\
 &  \textbf{No} & & & 0.58 & (0.48, 0.70) & 576 & 312 \\
 \hline
CHECKMATE057 & \textbf{Yes} & 0.73 & (0.59, 0.89) & 0.73 & (0.57, 0.93) & 263 & 152\\
 &  \textbf{No} & & & 1.05 & (0.90, 1.22) & 491 & 435 \\
 \hline
CHECKMATE078 & \textbf{Yes} & 0.68 & (0.52, 0.90) & 0.72 & (0.54, 0.97) & 194 & 111\\
 &  \textbf{No} & & & 1.05 & (0.90, 1.23) & 492 & 439 \\
 \hline
KEYNOTE010 & \textbf{Yes} & 0.71 & (0.58, 0.88) & 0.73 & (0.60, 0.90) & 261 & 362\\
 &  \textbf{No} & & & 0.85 & (0.74, 0.97) & 494 & 867 \\
 \hline
OAK & \textbf{Yes} & 0.73 & (0.62, 0.87) & 0.59 & (0.36, 0.96) & 276 & 41\\
 &  \textbf{No} & & & 0.66 & (0.48, 0.92) & 572 & 83 \\
 \hline
KEYNOTE024 & \textbf{Yes} & 0.63 & (0.47, 0.86) & 0.91 & (0.71, 1.17) & 188 & 474\\
 &  \textbf{No} & & & 0.94 & (0.82, 1.08) & 664 & 1394 \\
 \hline
KEYNOTE033 & \textbf{Yes} & 0.75 & (0.60, 0.95) & 0.77 & (0.63, 0.95) & 255 & 437\\
 &  \textbf{No} & & & 0.83 & (0.74, 0.95) & 556 & 1070 \\
 \hline
LUXLUNG6 & \textbf{Yes} & 0.9 & (0.71, 1.14) & 1.05 & (0.71, 1.55) & 2621 & 37 \\
 &  \textbf{No} & & & 1.18 & (0.91, 1.53) & 5187 & 82 \\
 \hline
CHECKMATE017 & \textbf{Yes} & 0.59 & (0.44, 0.79) & 0.74 & (0.55, 0.99) & 197 & 111\\
 &  \textbf{No} & & & 1.04 & (0.89, 1.21) & 492 & 437\\
 \hline
EMPHASIS & \textbf{Yes} & ? & ? & 0.69 & (0.58, 0.82) & 37 & 439\\
 &  \textbf{No} & & & 0.71 & (0.61, 0.82) & 534 & 615 \\
 \hline
NCT02604342 & \textbf{Yes} & ? & ? & 0.38 & (0.20, 0.72) & 1774 & 26\\
 &  \textbf{No} & & & 0.44 & (0.27, 0.71) & 2625 & 42 \\
\hline \hline
\end{tabular}\caption{\emph{Simulation of the 11 selected single-drug trials using a Cox-PH model without adjustment.}}\label{table:simulation_results_no_adjustment}
\end{table}


\clearpage
\section{EMPHASIS trial diagnostics test results}\label{sup:emphasis_diagnostics_results}


\begin{figure*}[h!]
\caption{\emph{Results of the diagnostics test for the EMPHASIS trial. \textbf{A}: Group Re-balancing: the standardized mean distance (SMD) for all covariates is close to zero after rebalancing the groups. \textbf{B}: Overlapping set: 99\% of the RwD set had non zero probability of being selected for the treatment and control group (\autoref{sec:causal_modeling}). \textbf{C}: Placebo treatment: the signal in the simulation vanishes when the assignment to the treatment is randomly permuted (100 replicates per scenario). \textbf{D}: Random  confounder: the estimation of the HR is robust against wrongly selected confounders that are unrelated with the treatment and the response. The figure shows the CI for the HR for a randomly generated Gaussian with mean zero and varying standard deviation (100 replicates per scenario). \textbf{E}: Down-sampling: the estimation of the HR is robust against sub-sampling. In average the same HR is observed even when only 25\% of the sample is randomly selected and used to to simulate the trial (100 replicates per scenario). Diagnostics results for other trials simulated in this study can be found in Section \ref{sup:validation}.}}
\label{fig:diagnosis}
\centering
\includegraphics[width=16.5cm]{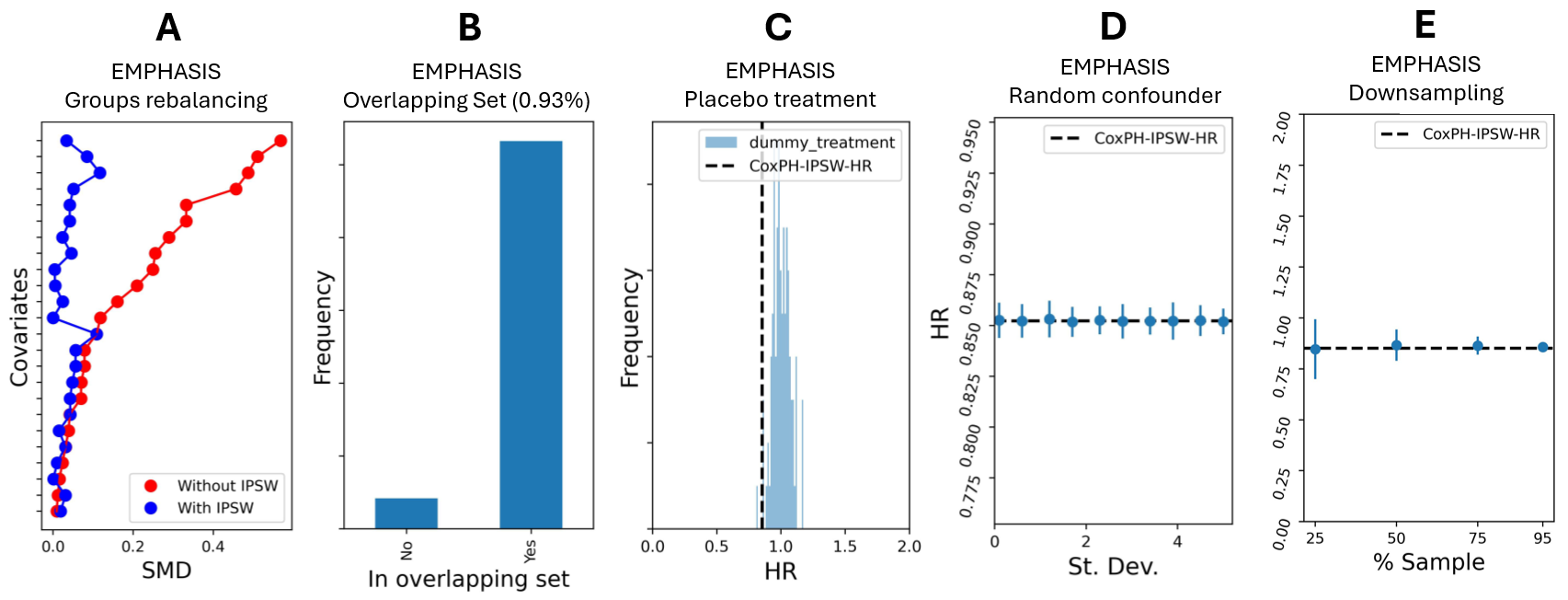}
\end{figure*}

\vspace{0.5cm}
\clearpage
\section{EMPHASIS trial summary statistics}\label{sup:EMPHASIS_summary_stats}

\begin{table}[h!]
\centering
\vspace{0.5cm}
\small
\begin{tabular}{ll|cc|cc}
\hline \hline
& & \textbf{Simulation-T} & \textbf{Simulation-C} & \textbf{RCT-T} & \textbf{RCT-C}\\ 
\hline \hline
\textbf{Counts}	&&	439	&	370	&	38	&	42 \\
\hline
\textbf{Age}	    &	\textbf{Mean}	 &	66.09	&	69.22	&	66.7	&	70.1 \\
	&	\textbf{Median}	&	67	&	69	&	NA	&	NA \\
        &	\textbf{Min}	&	35	&	32	&	44.4	&	53.3 \\
	&	\textbf{Max}	&	92 	&	95 	&	82	&	84 \\
 \hline
\textbf{Gender}  &	\textbf{Female}	&	256 (58.31  \%)	&	163 (44.05  \%)	&	7 (18.4\%)	&	7 (16.7\%) \\
	&	\textbf{Male}	&	183 (41.69  \%)	&	207 (55.95  \%)	&	31 (81.6\%)	&	35 (83.3\%) \\
 \hline
\textbf{ECOG}    &	\textbf{0}	&	29 (6.61  \%)	&	59 (15.95  \%)	&	12 (31.6\%) 	&	15 (35.7\%) \\
	&	\textbf{1}	&	219 (49.89  \%)	&	236 (63.78  \%)	&	24 (63.2\%)	&	22 (52.4\%) \\
	&	\textbf{2}	&	191 (43.5\%)	&	75 (20.27\%)	    &	2 (5.3\%)	&	5 (11.9\%) \\
 \hline
\textbf{Race}	&	Asian&		71 (16.17  \%)	&28 (7.57  \%)	&NA	&NA \\
	&	\textbf{Black/African American}&		12 (2.73  \%)	&	8 (2.16  \%)	&	NA	&	NA \\
	&	\textbf{White/Caucasian}&		297 (67.65  \%)	&	292 (78.92  \%)	&	NA	&	NA \\
	&	\textbf{Other}&		59 (13.46\%)	&	42 (11.35\%)	&	NA	&	NA \\
 \hline
\textbf{Smoking}	&	\textbf{No} &		342 (77.9  \%)&		206 (55.68  \%)	&	16 (4.21\%)&		12 (28.6\%) \\
	&	\textbf{Yes}&	97 (22.1  \%)	&	164 (44.32  \%)	&	22(57.9\%)	&	27 (64.3\%) \\
\hline \hline
\end{tabular}\caption{\emph{Summary statistics for the data in the original RCT publication and the simulation (Providence data) for the reported covariates in EMPHASIS (T=treatment, C=Control). NA stands for not available values. The two populations differ, which implies that the conclusions of the simulation cannot be extrapolated to the trial.}}\label{table:summary_statistics}
\end{table}

\vspace{0.5cm}
\clearpage
\section{Diagnostics and summary statistics}\label{sup:validation}


\begin{figure}[h!]
\caption{\emph{Results of the diagnostics test for FLAURA, CHECKMATE057 and KEYNOTE078}}\label{fig:diagnosis_1}
\vspace{0.5cm}
\centering\includegraphics[width=3.2cm]{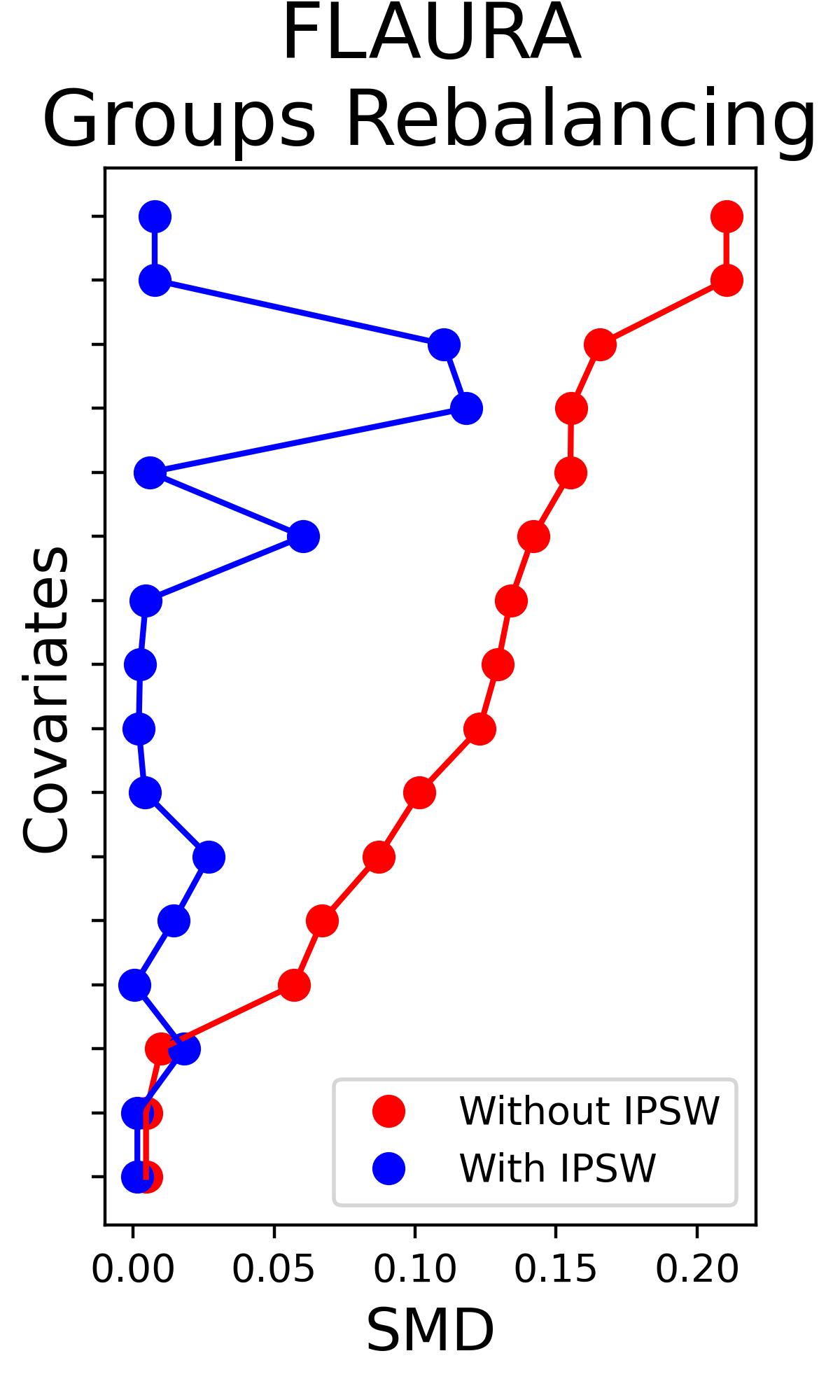}
\centering\includegraphics[width=3.2cm]{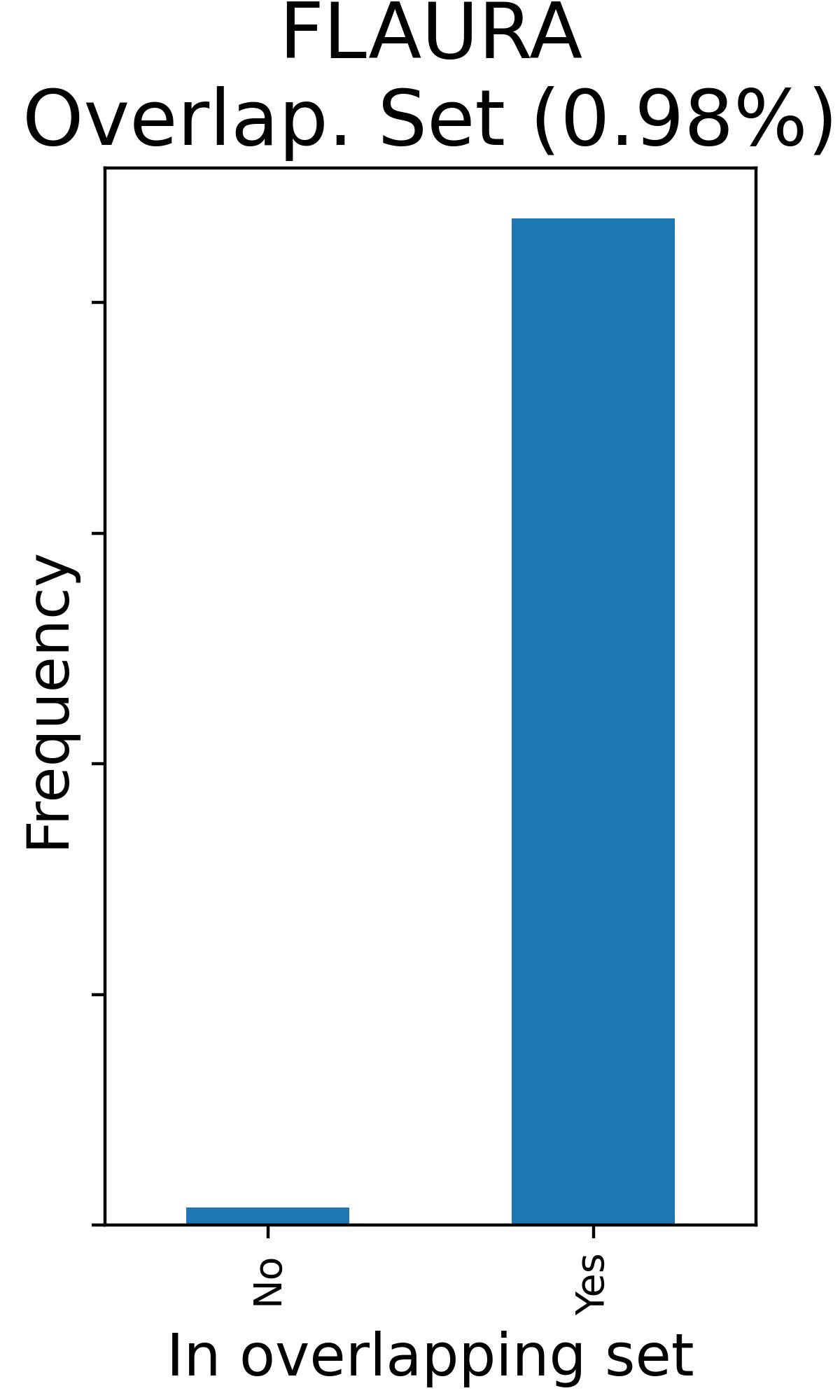}
\centering\includegraphics[width=3.3cm]{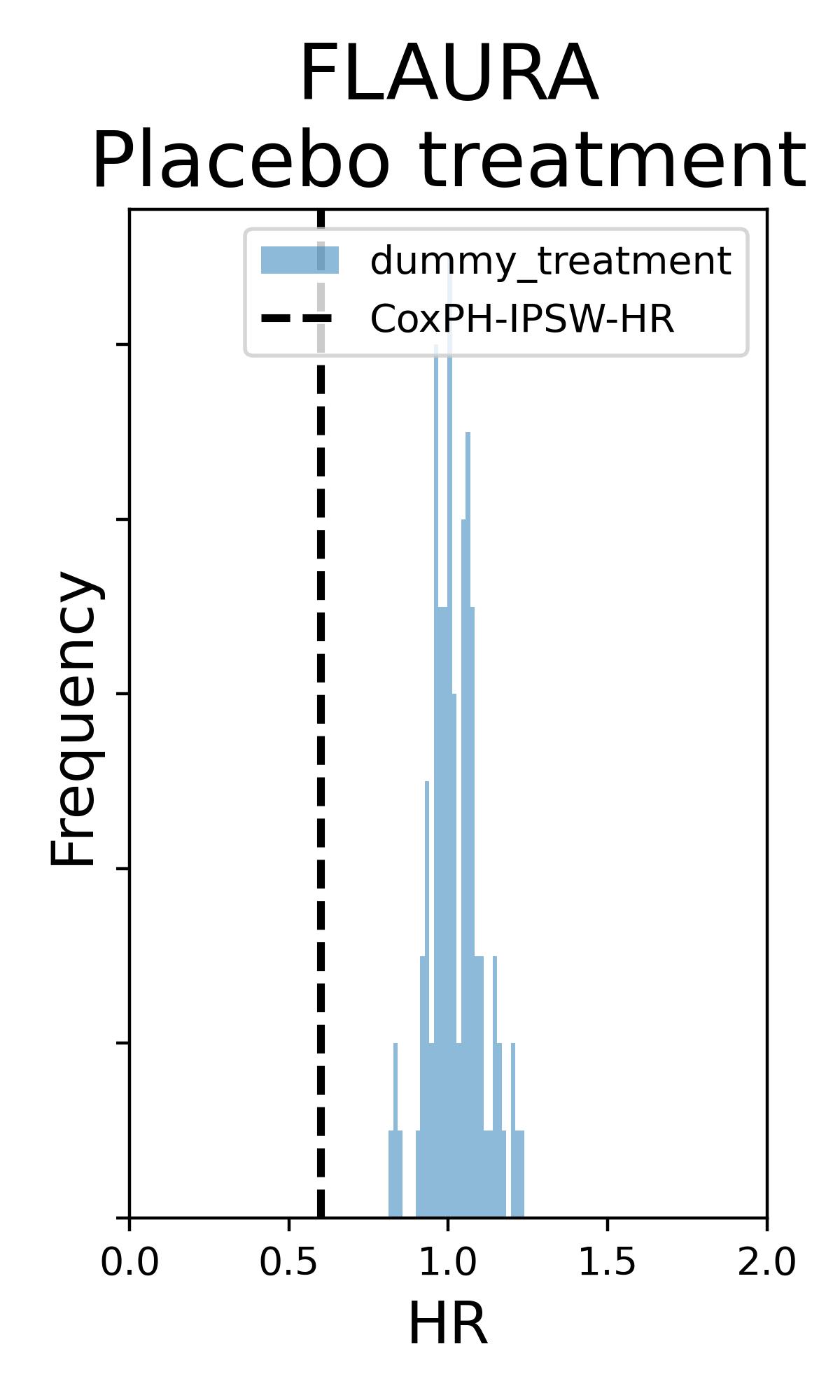}
\centering\includegraphics[width=3.3cm]{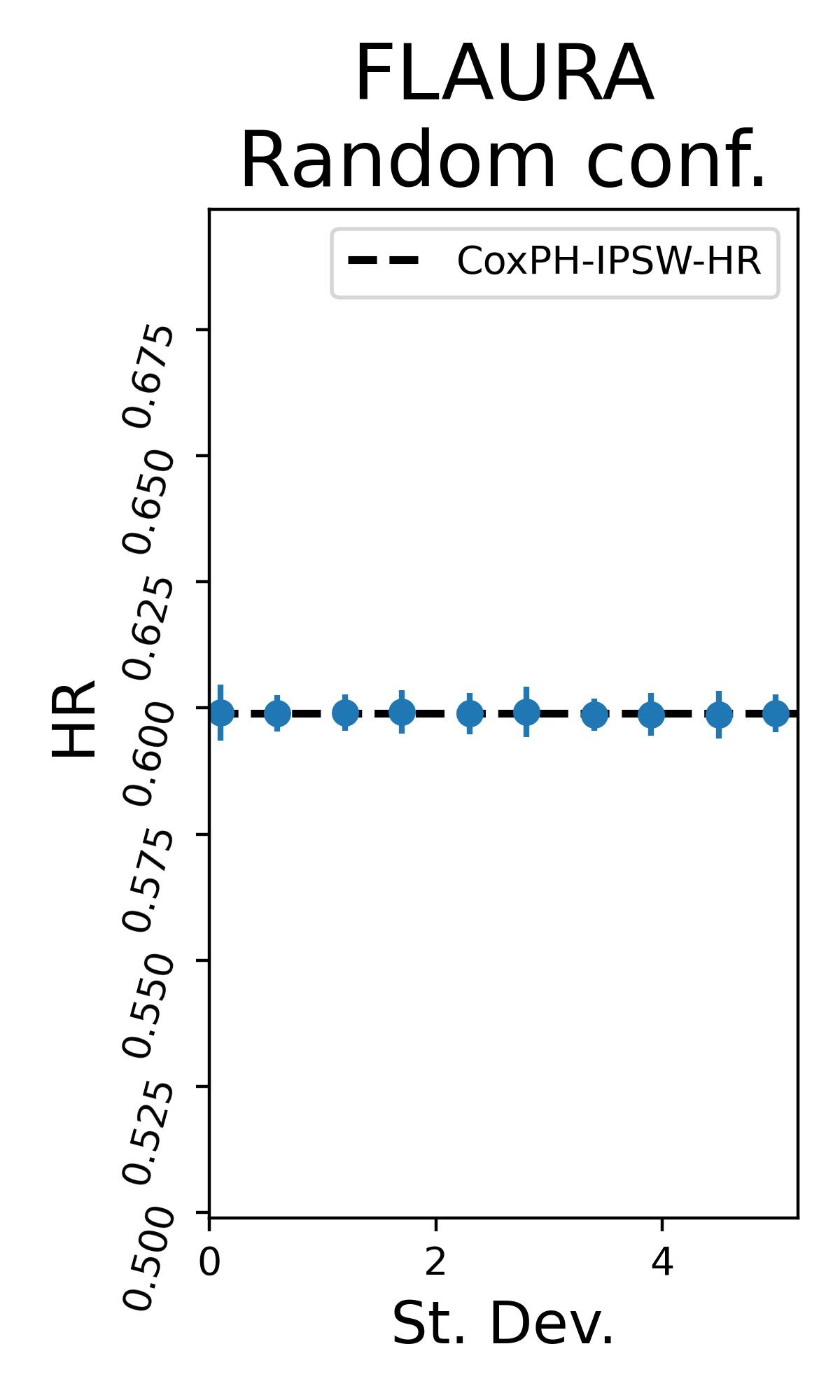}
\centering\includegraphics[width=3.2cm]{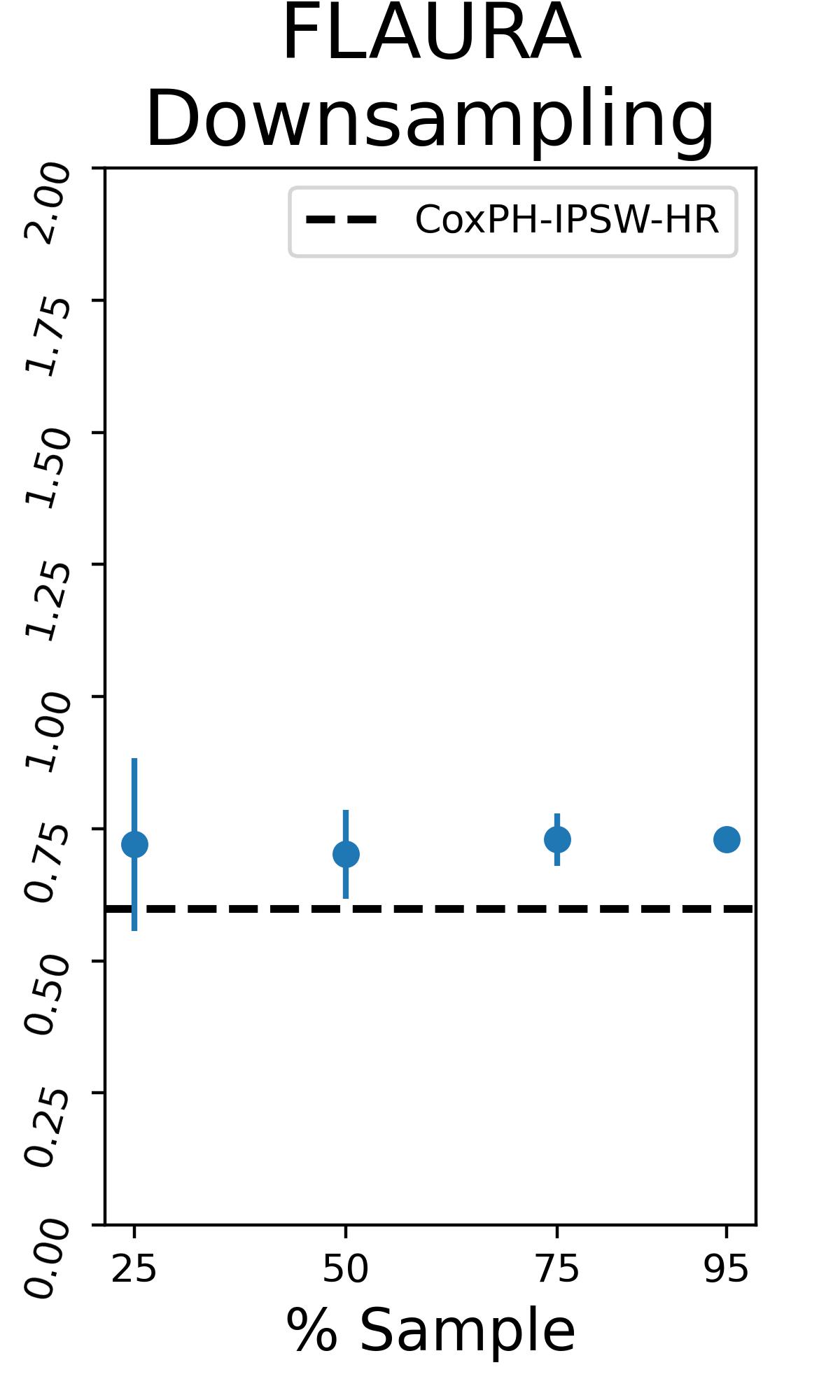} \\
\vspace{0.5cm}
\centering\includegraphics[width=3.2cm]{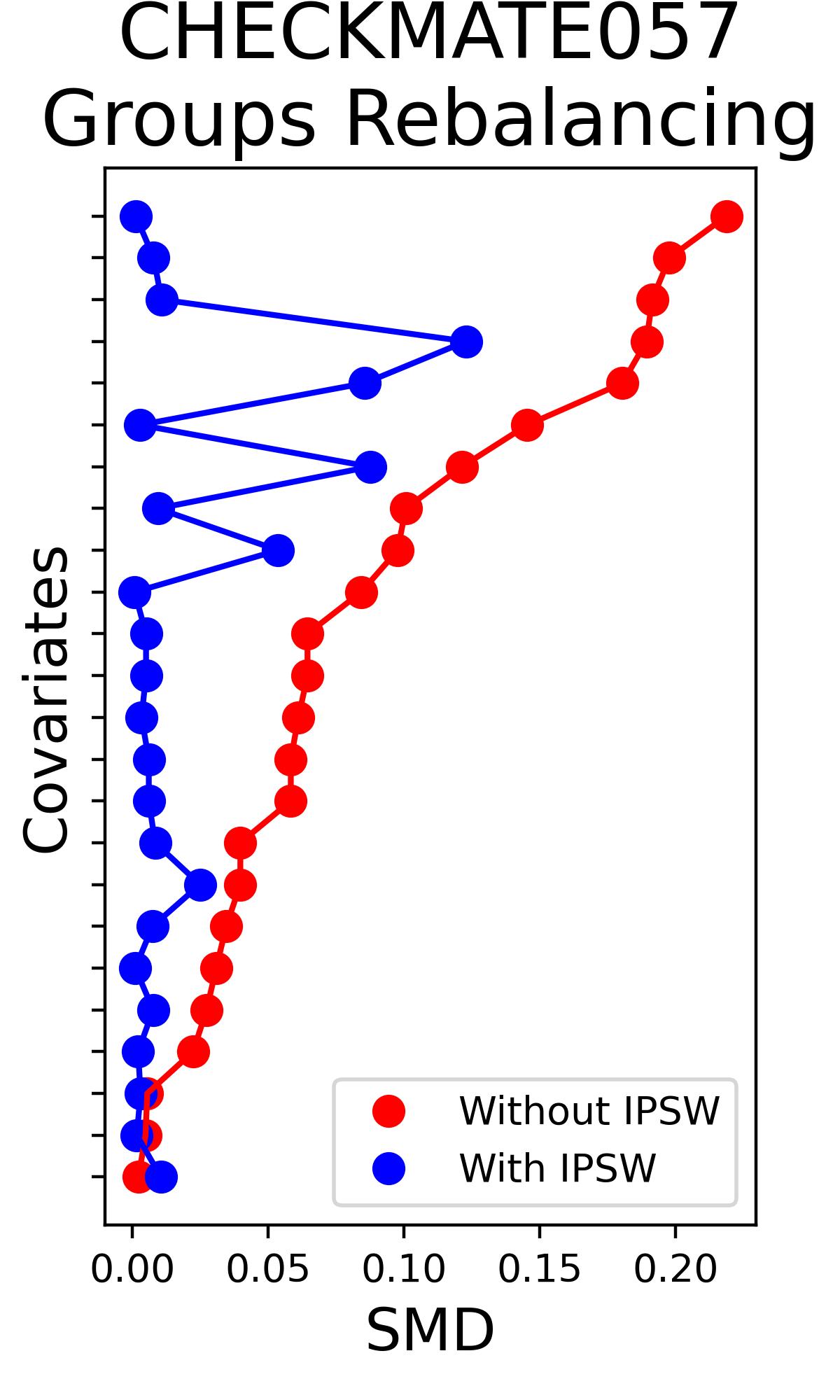}
\centering\includegraphics[width=3.2cm]{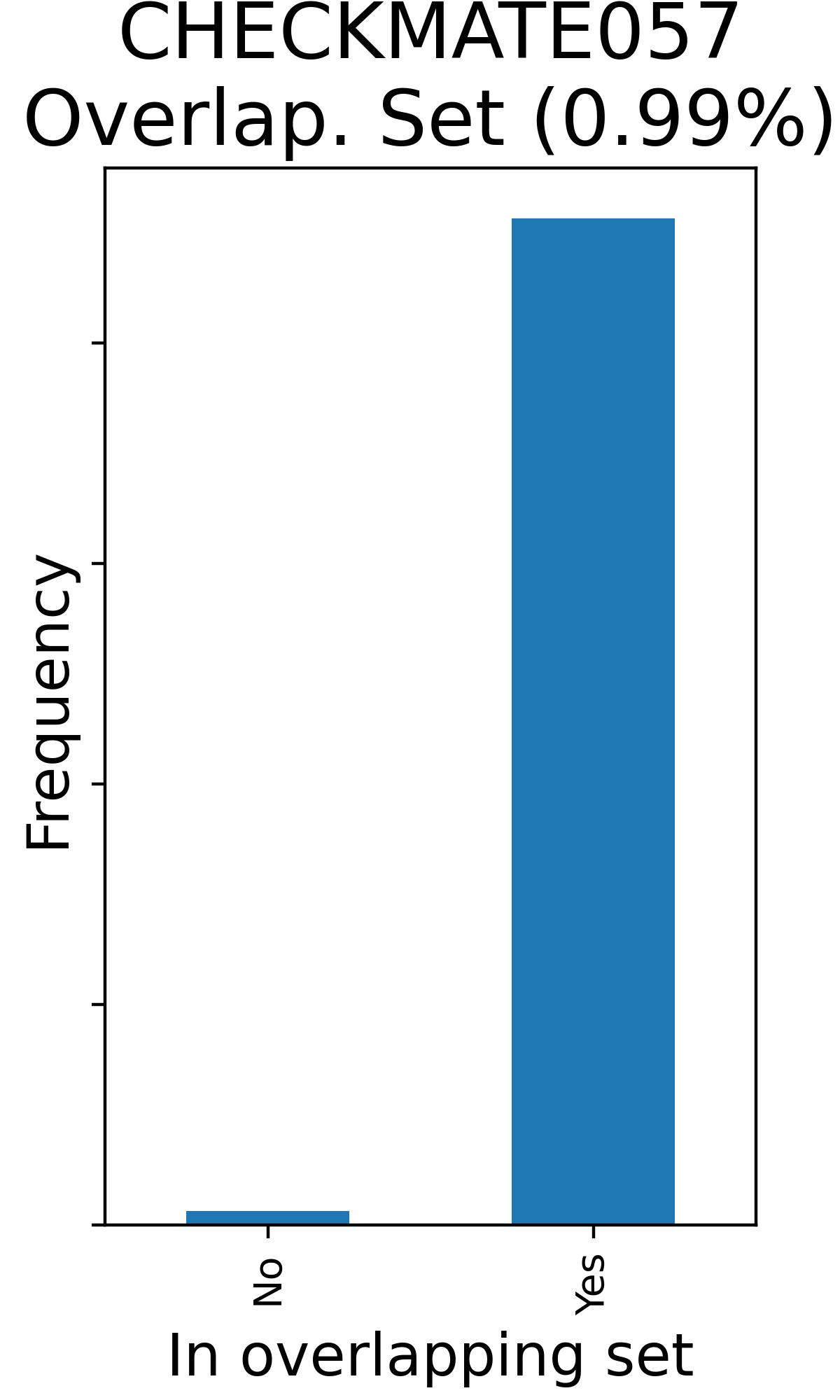}
\centering\includegraphics[width=3.3cm]{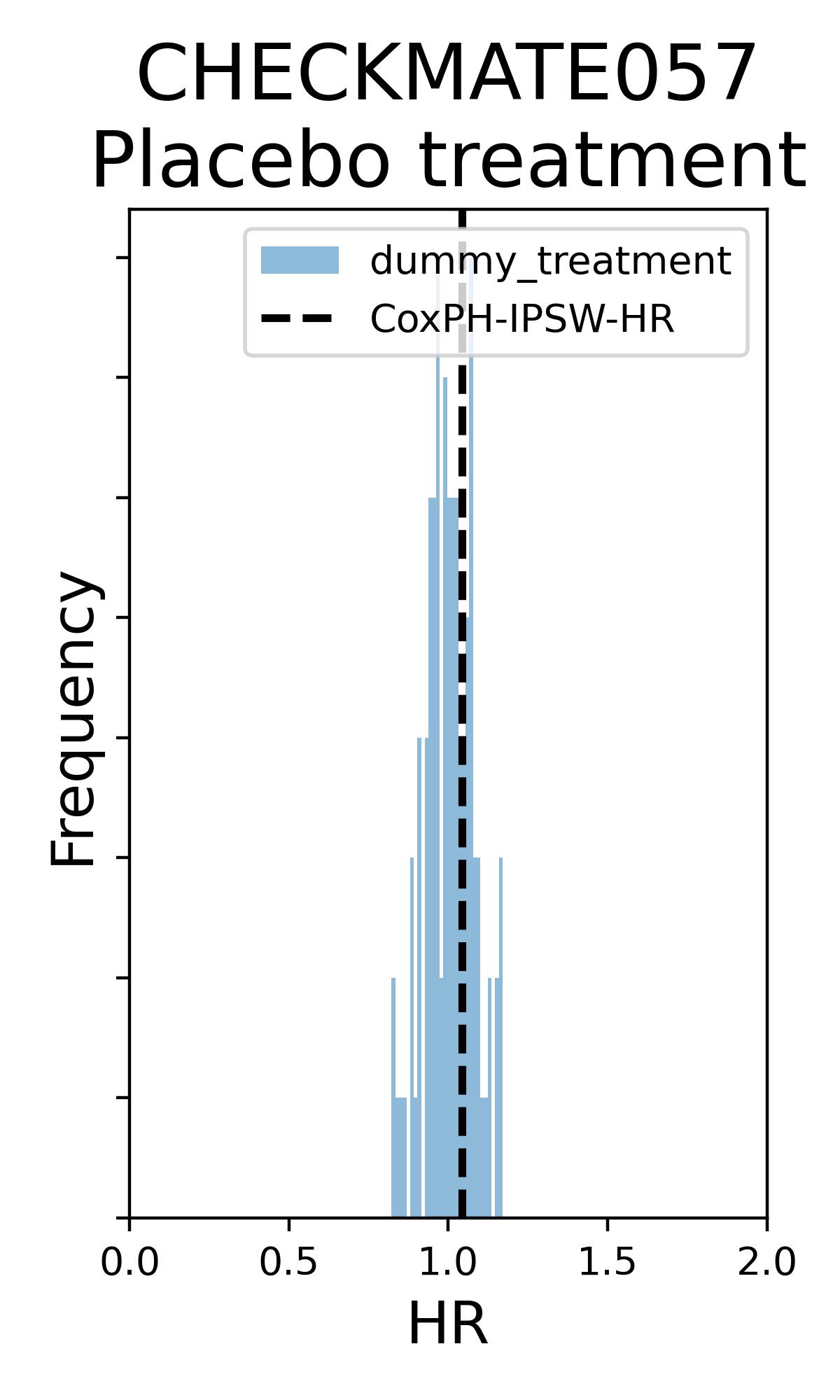}
\centering\includegraphics[width=3.3cm]{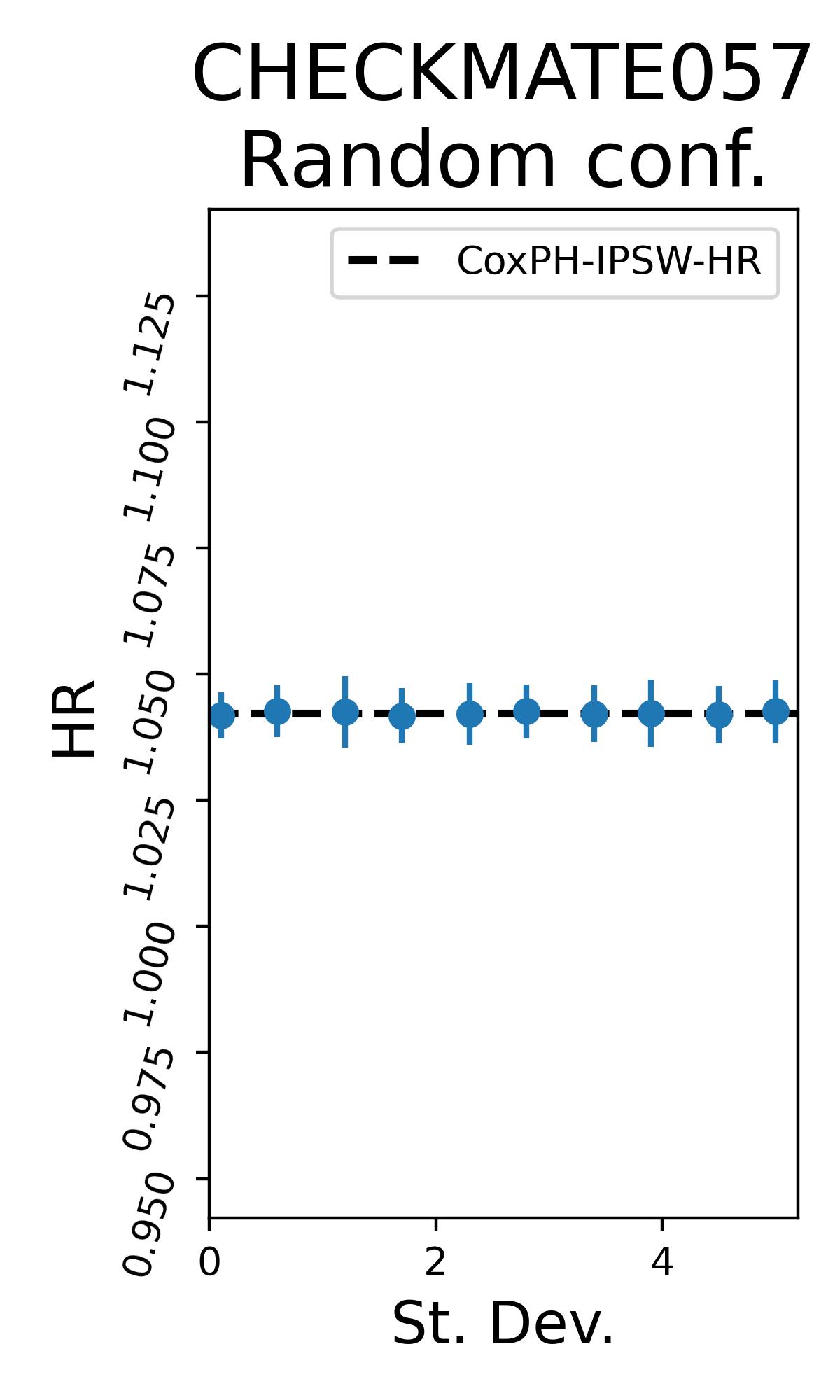}
\centering\includegraphics[width=3.2cm]{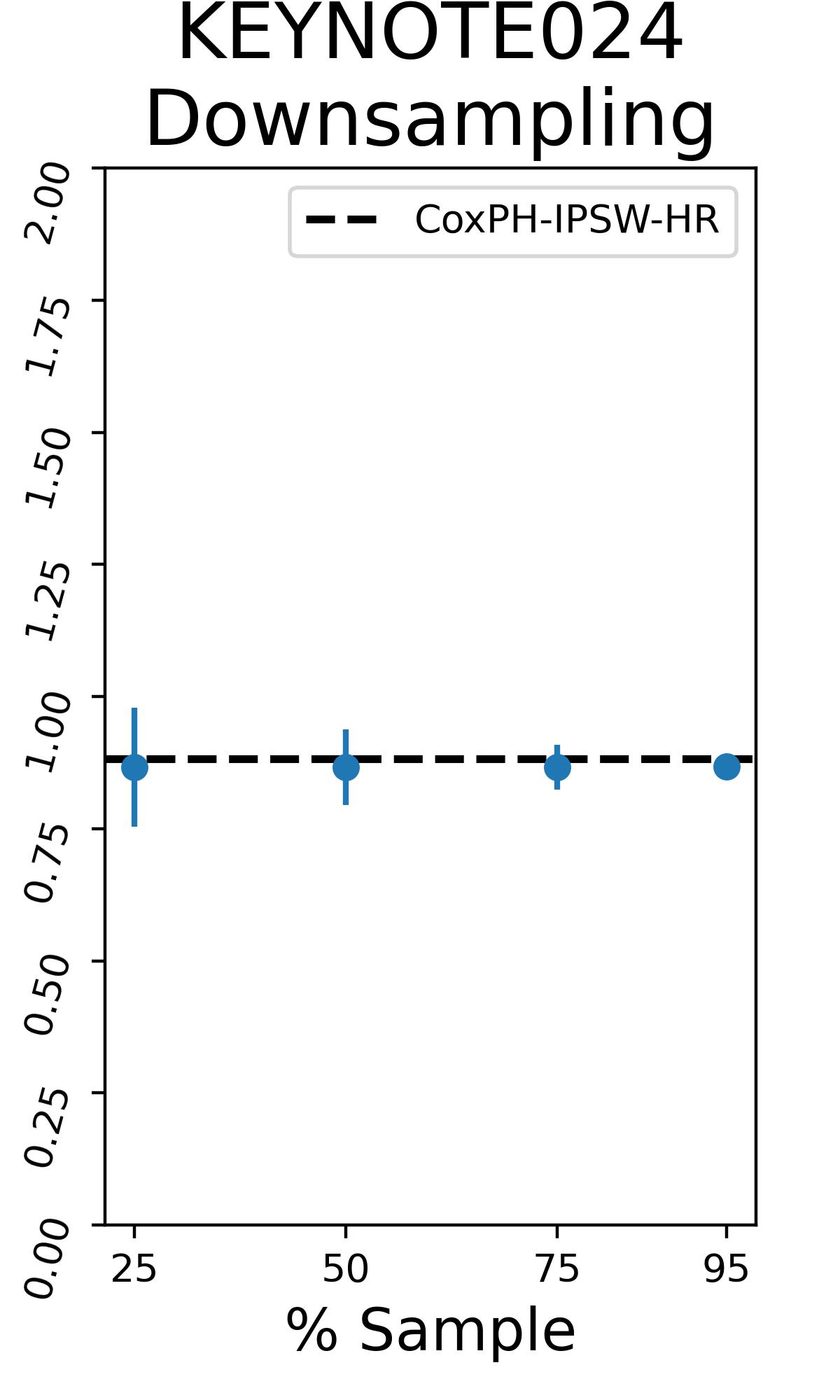} \\
\vspace{0.5cm}
\centering\includegraphics[width=3.2cm]{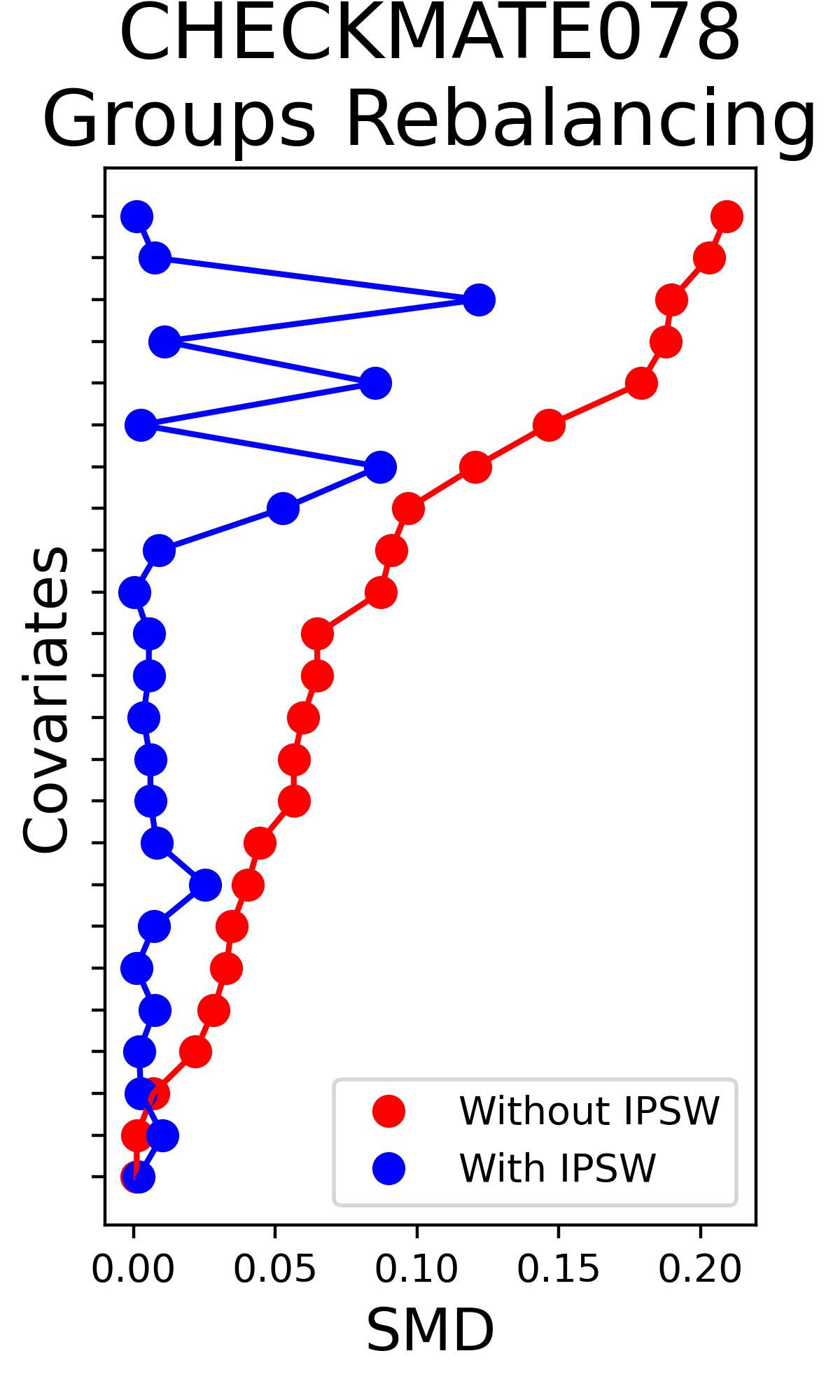}
\centering\includegraphics[width=3.2cm]{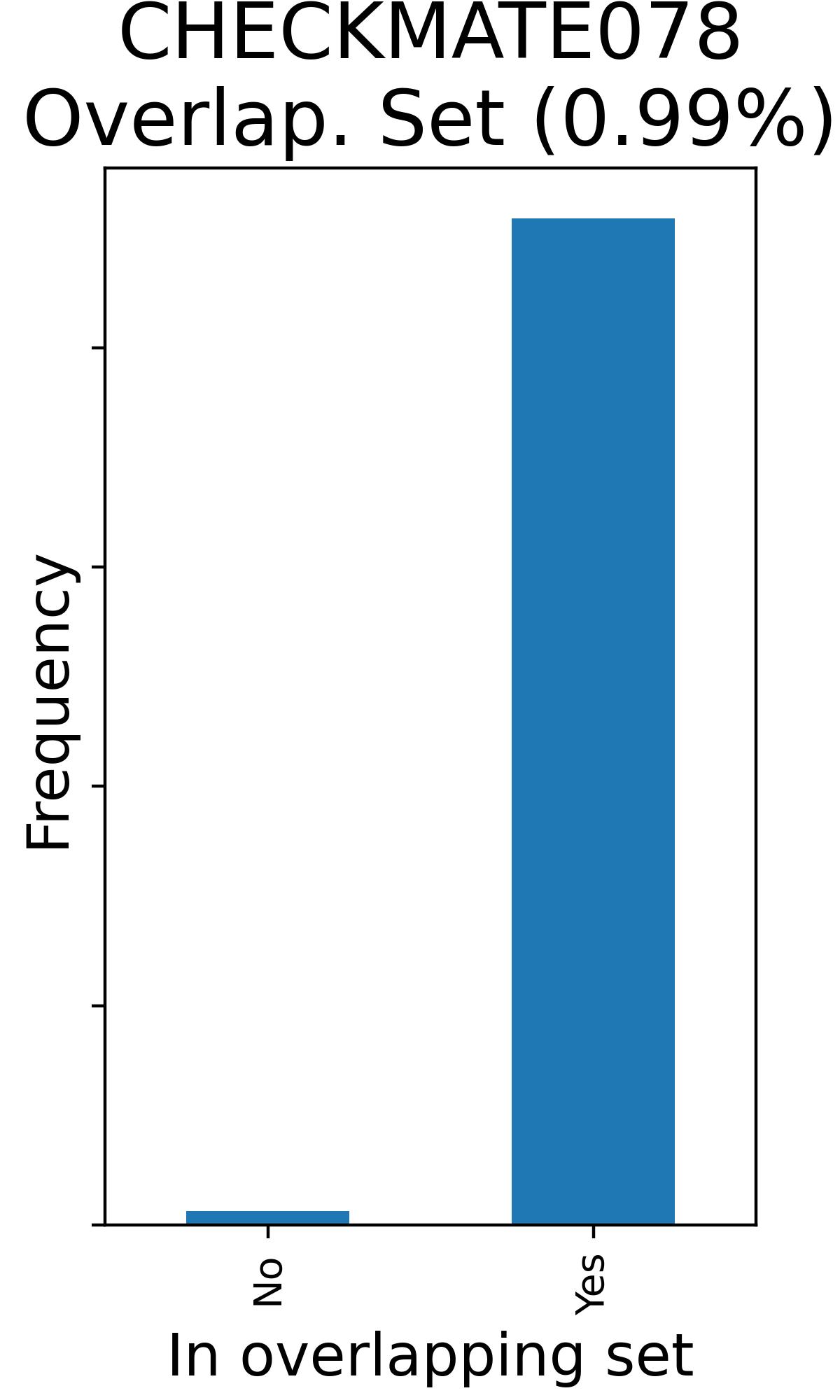}
\centering\includegraphics[width=3.3cm]{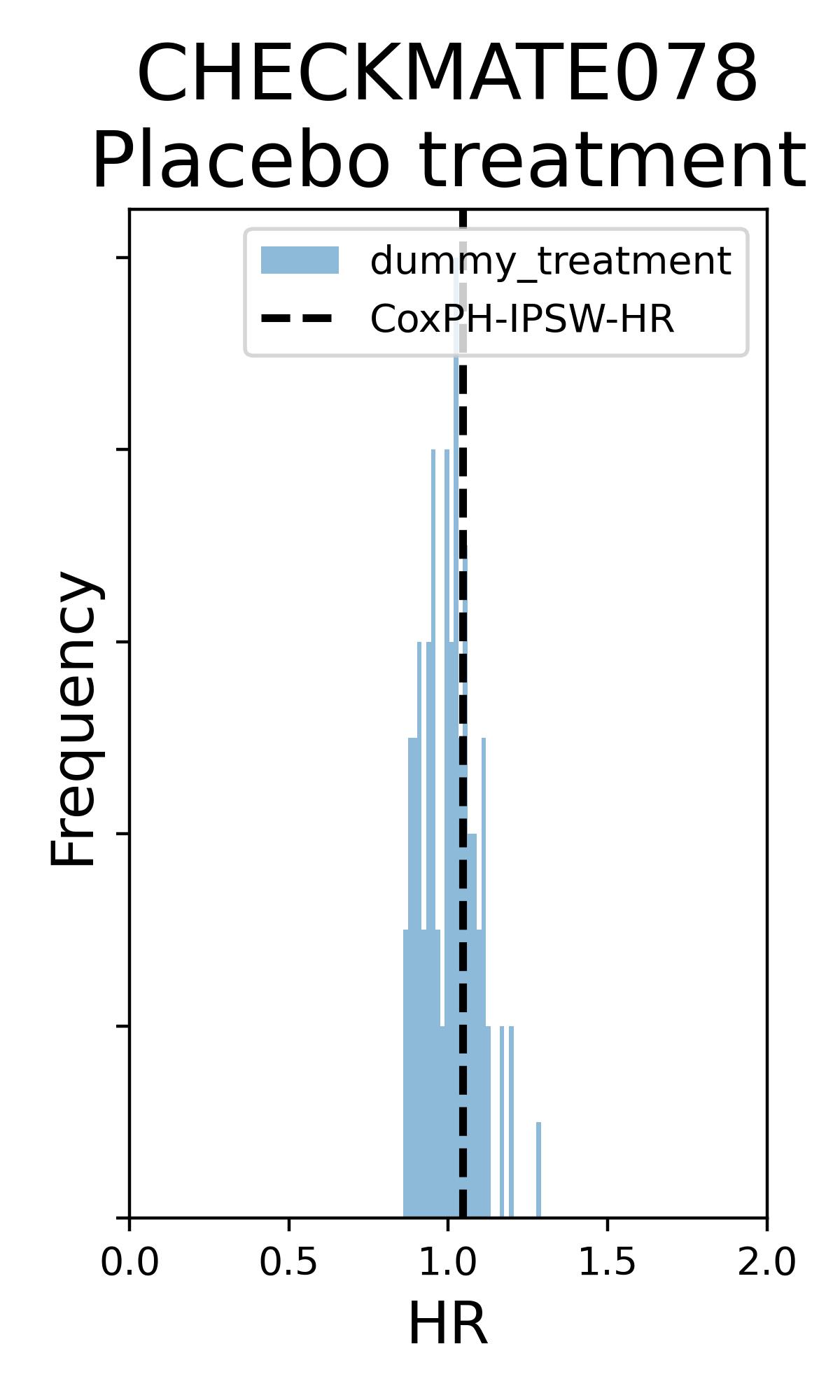}
\centering\includegraphics[width=3.3cm]{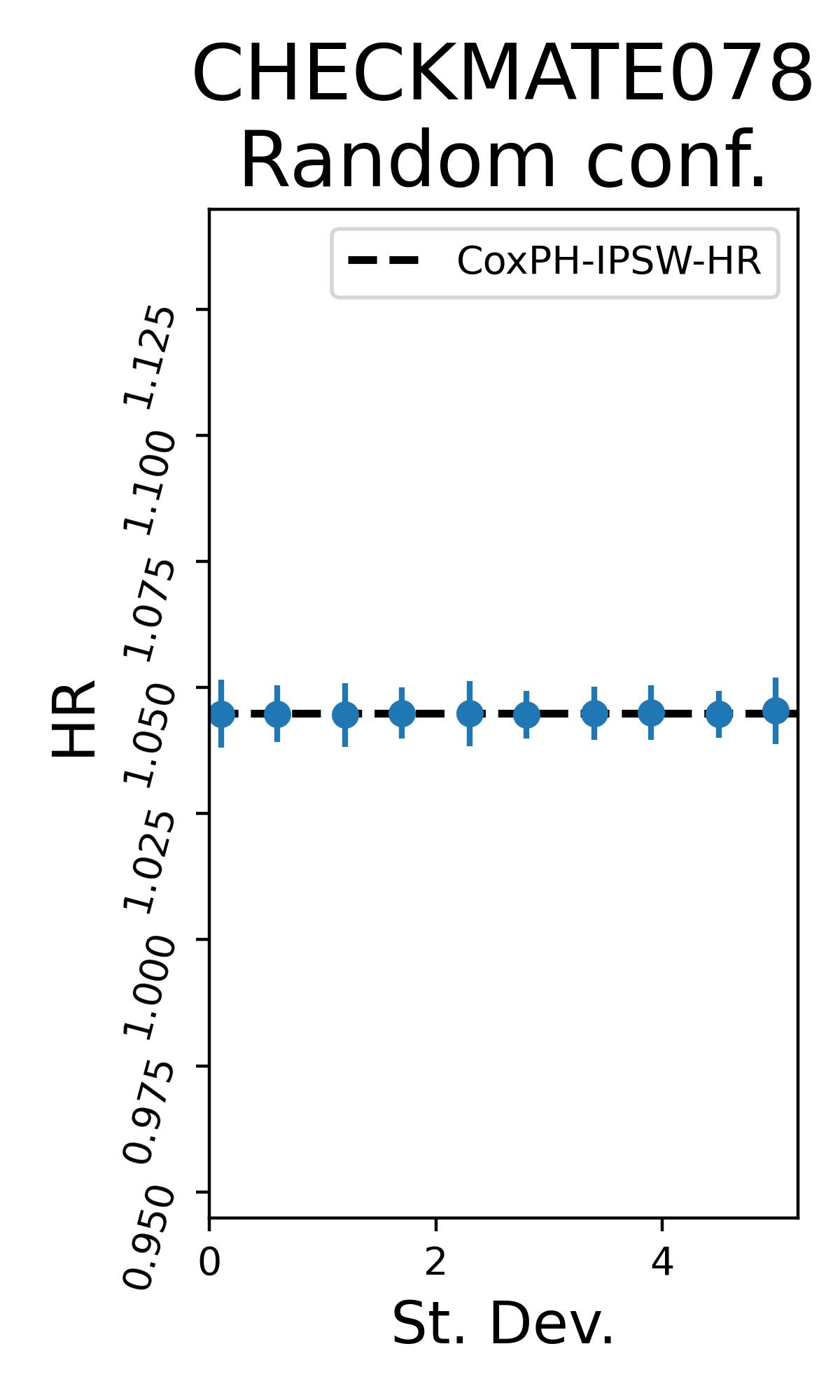}
\centering\includegraphics[width=3.2cm]{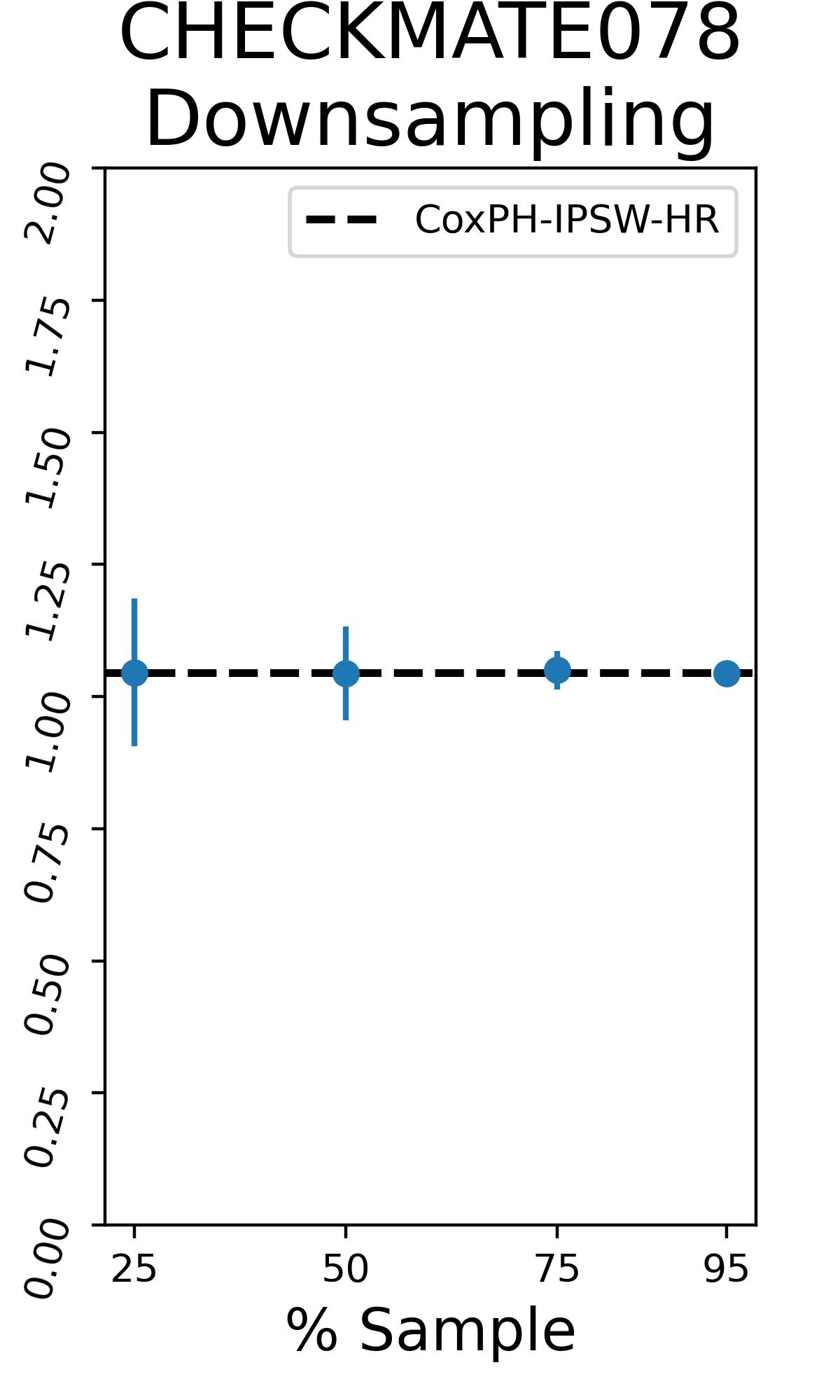} \\

\end{figure}

\begin{figure}[!ht]
\caption{\emph{Results of the diagnostics test for KEYNOTE010, OAK, KEYNOTE24 and KEYNOTE033}}\label{fig:diagnosis_2}
\vspace{0.5cm}
\centering\includegraphics[width=3.2cm]{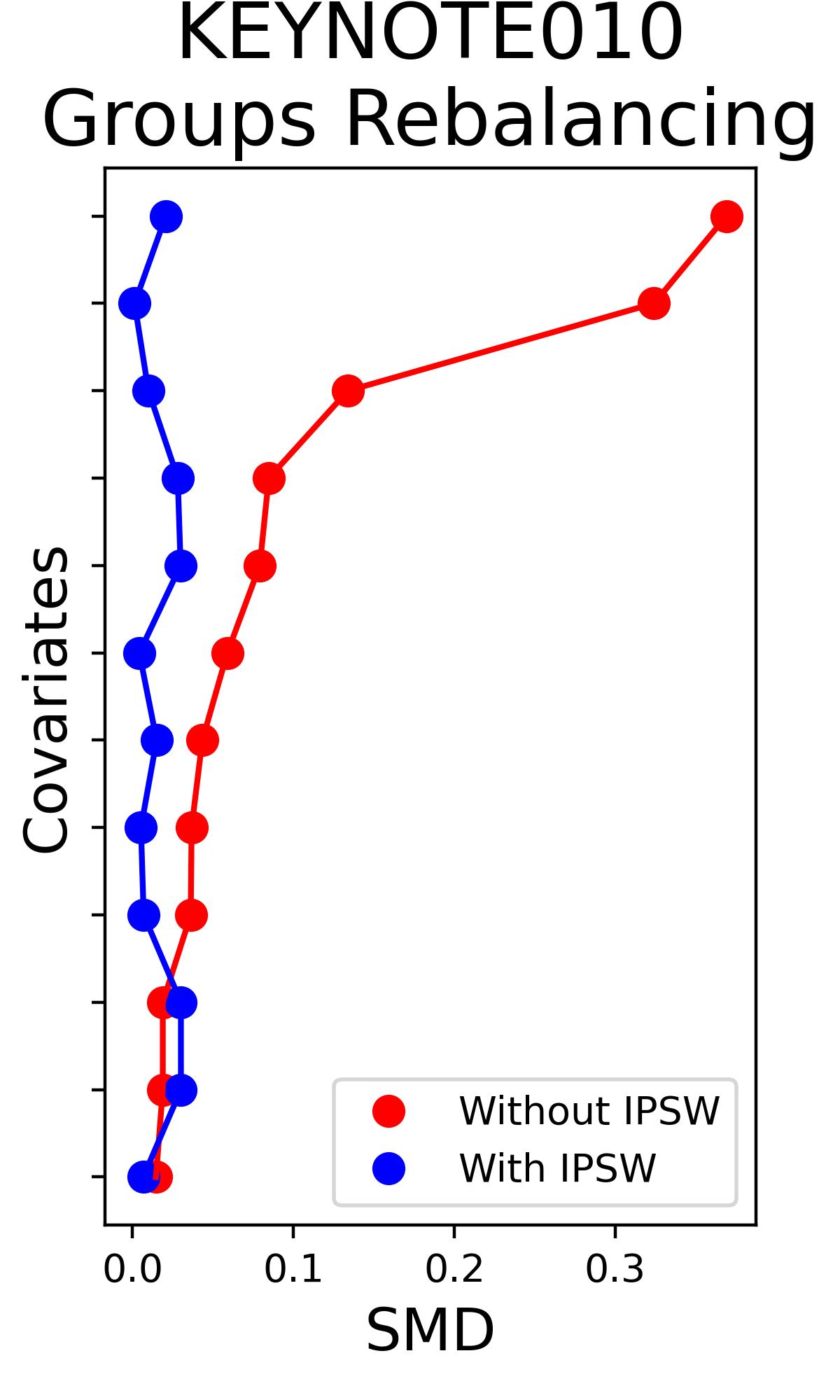}
\centering\includegraphics[width=3.2cm]{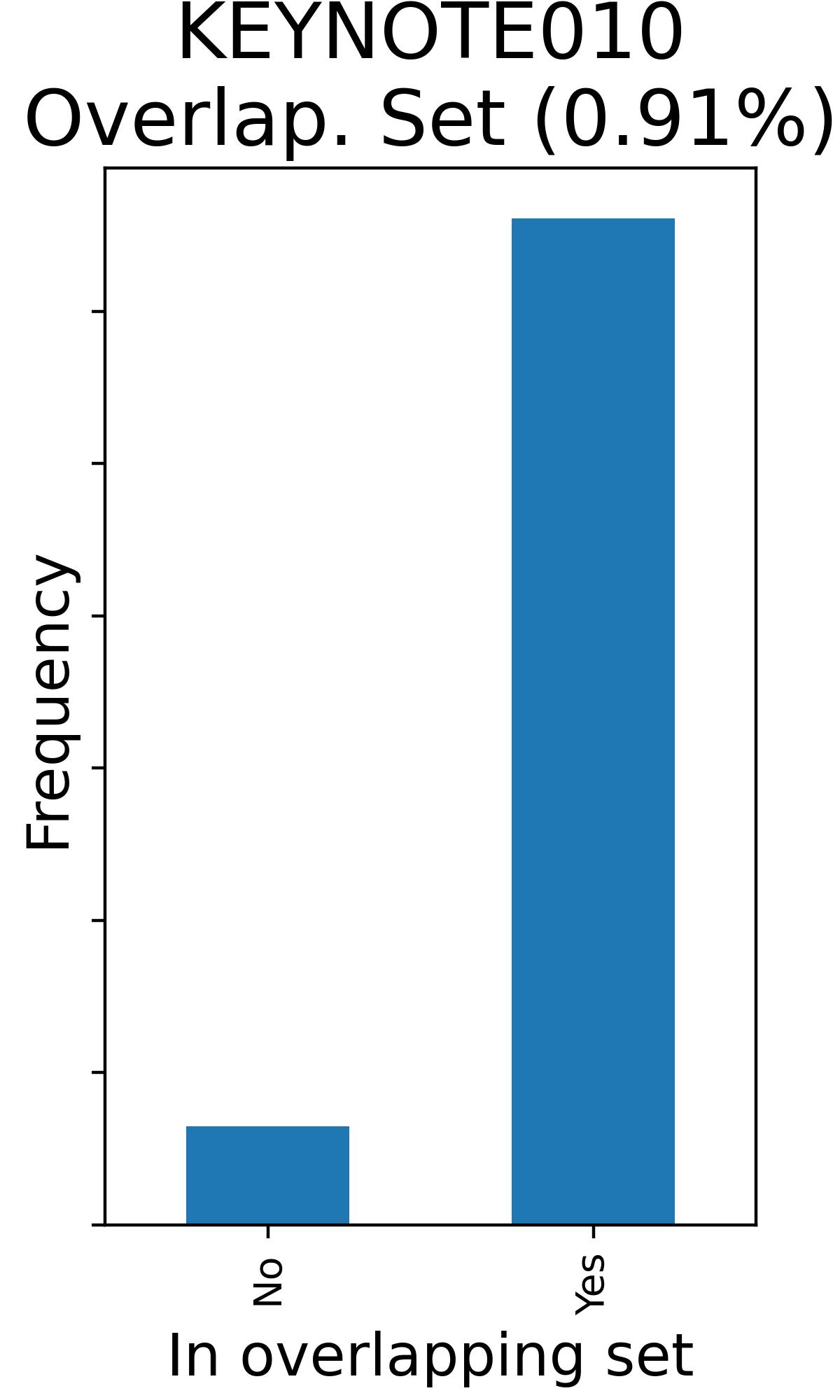}
\centering\includegraphics[width=3.3cm]{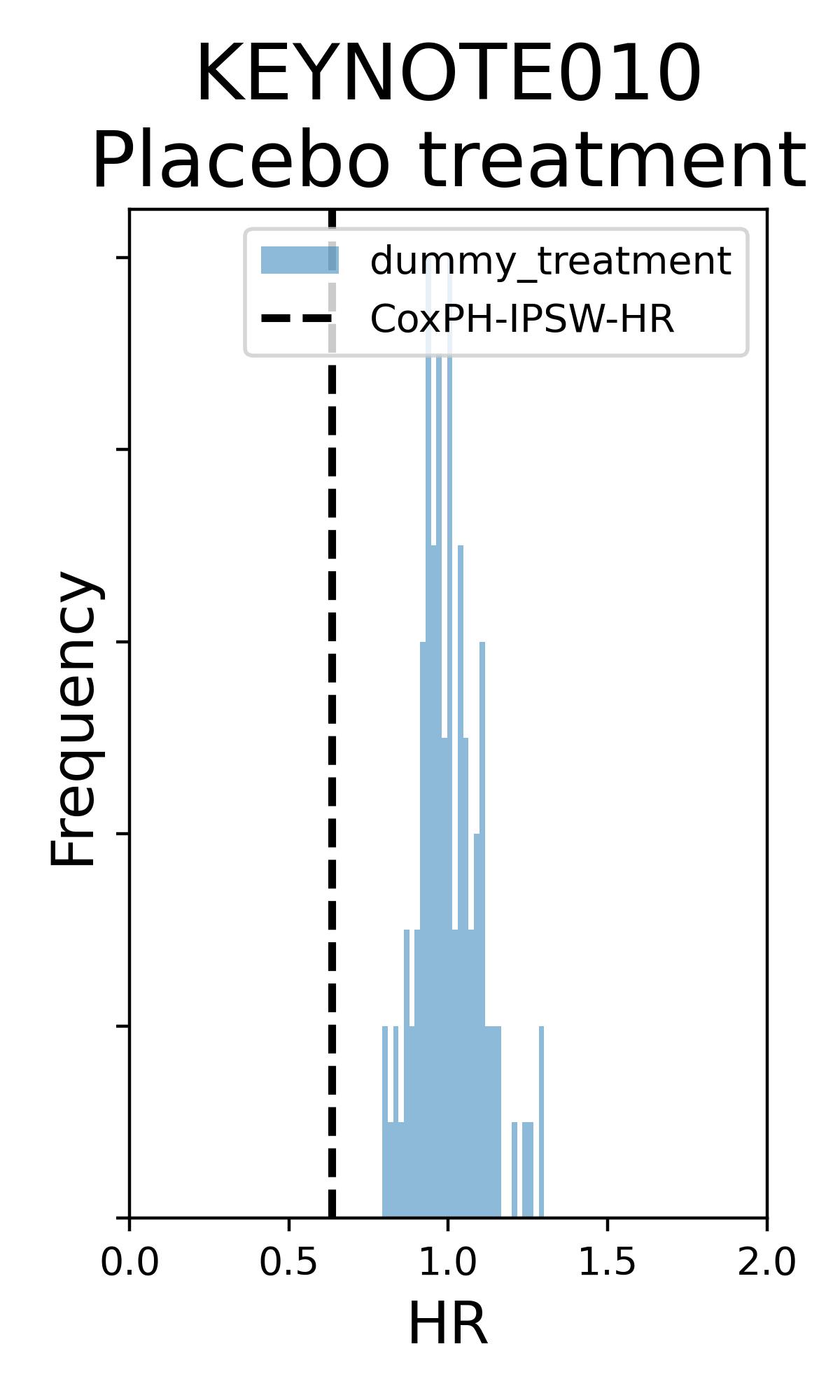}
\centering\includegraphics[width=3.3cm]{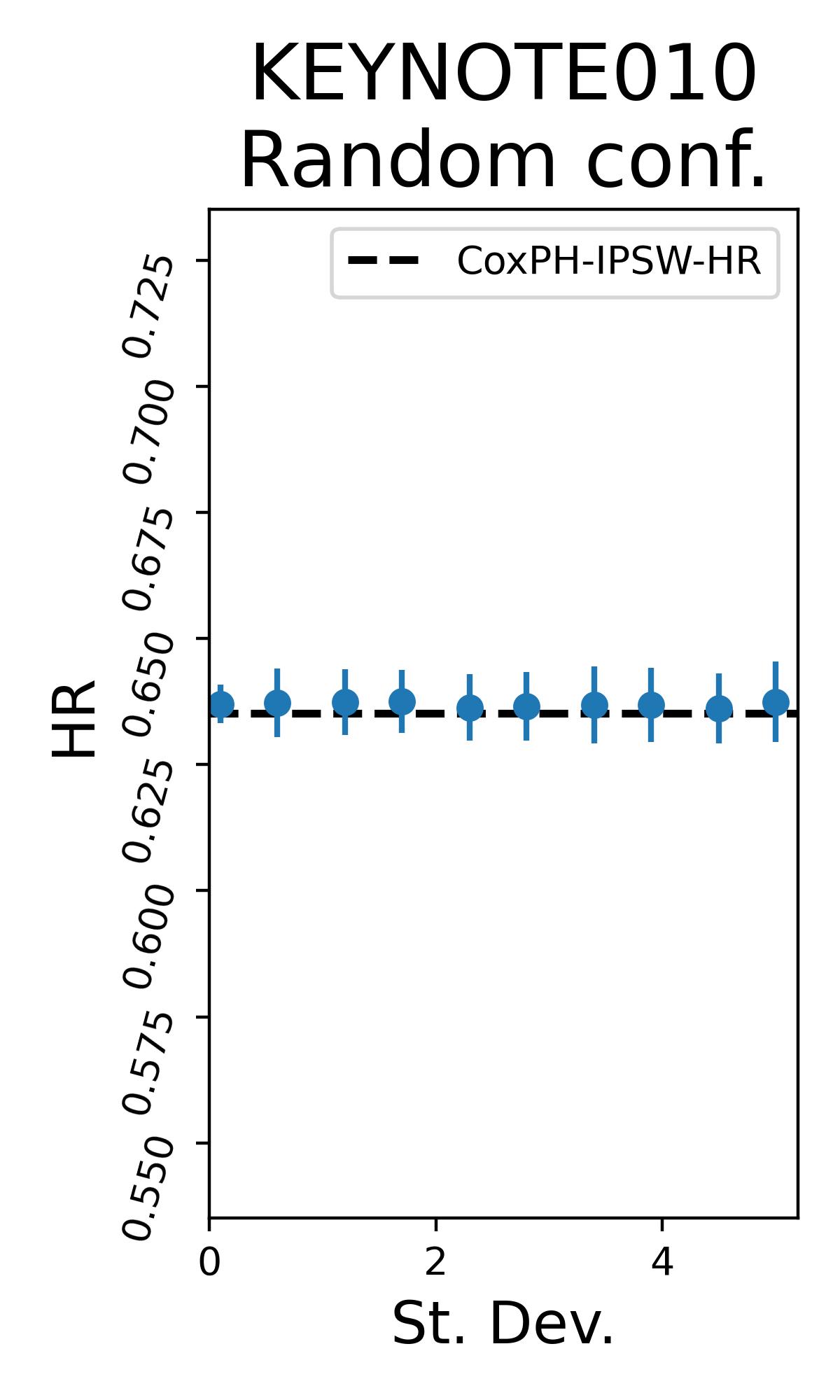}
\centering\includegraphics[width=3.2cm]{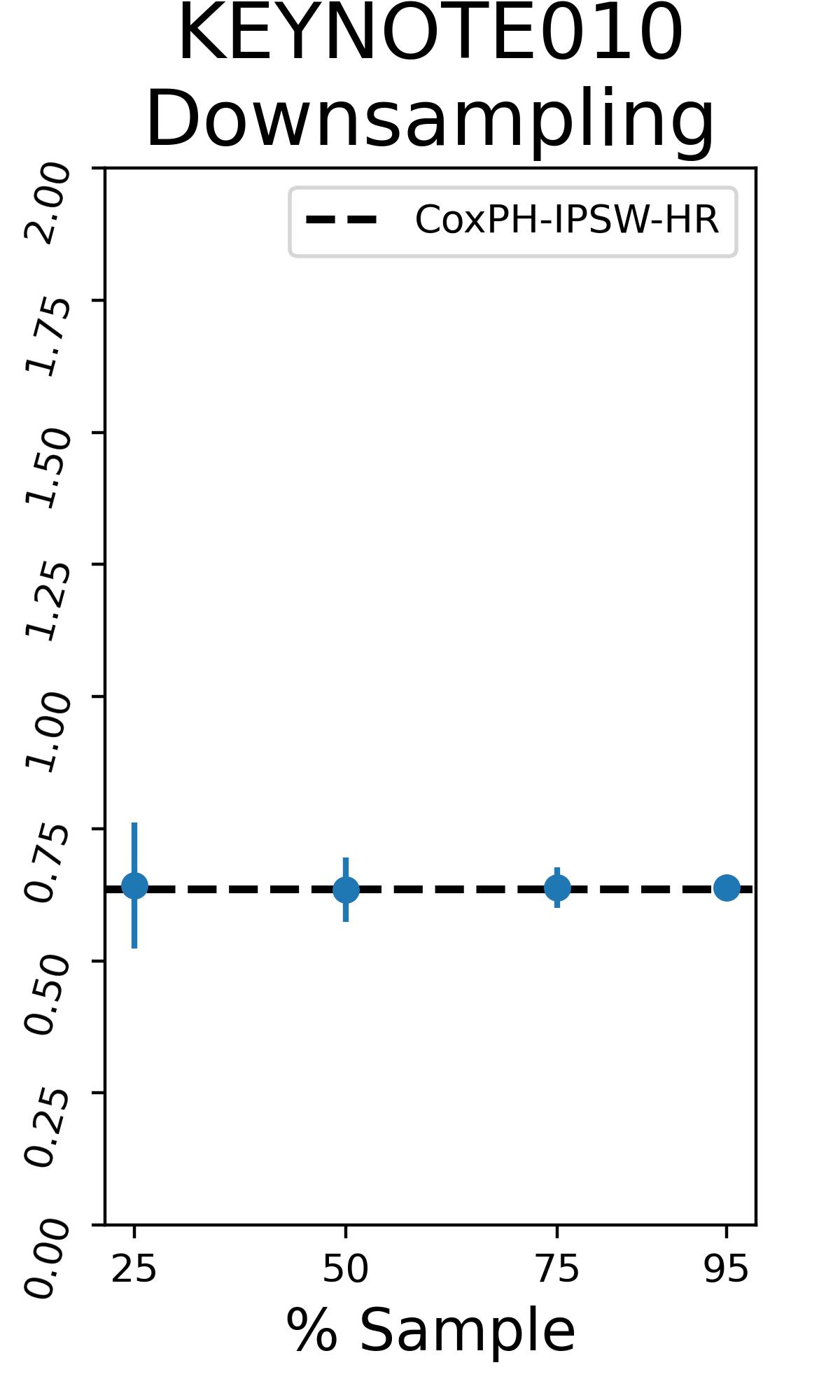}
 \\
\vspace{0.5cm}
\centering\includegraphics[width=3.2cm]{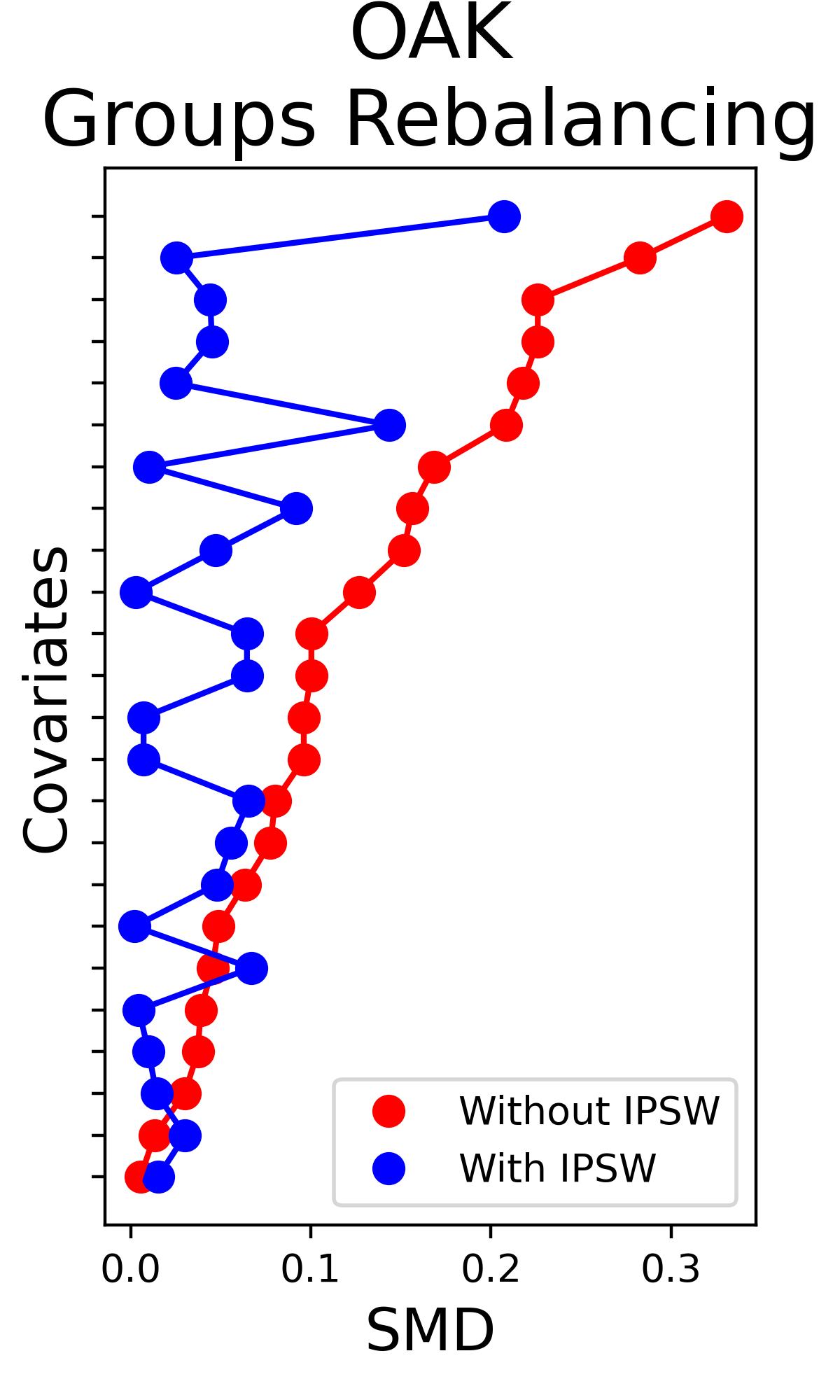}
\centering\includegraphics[width=3.2cm]{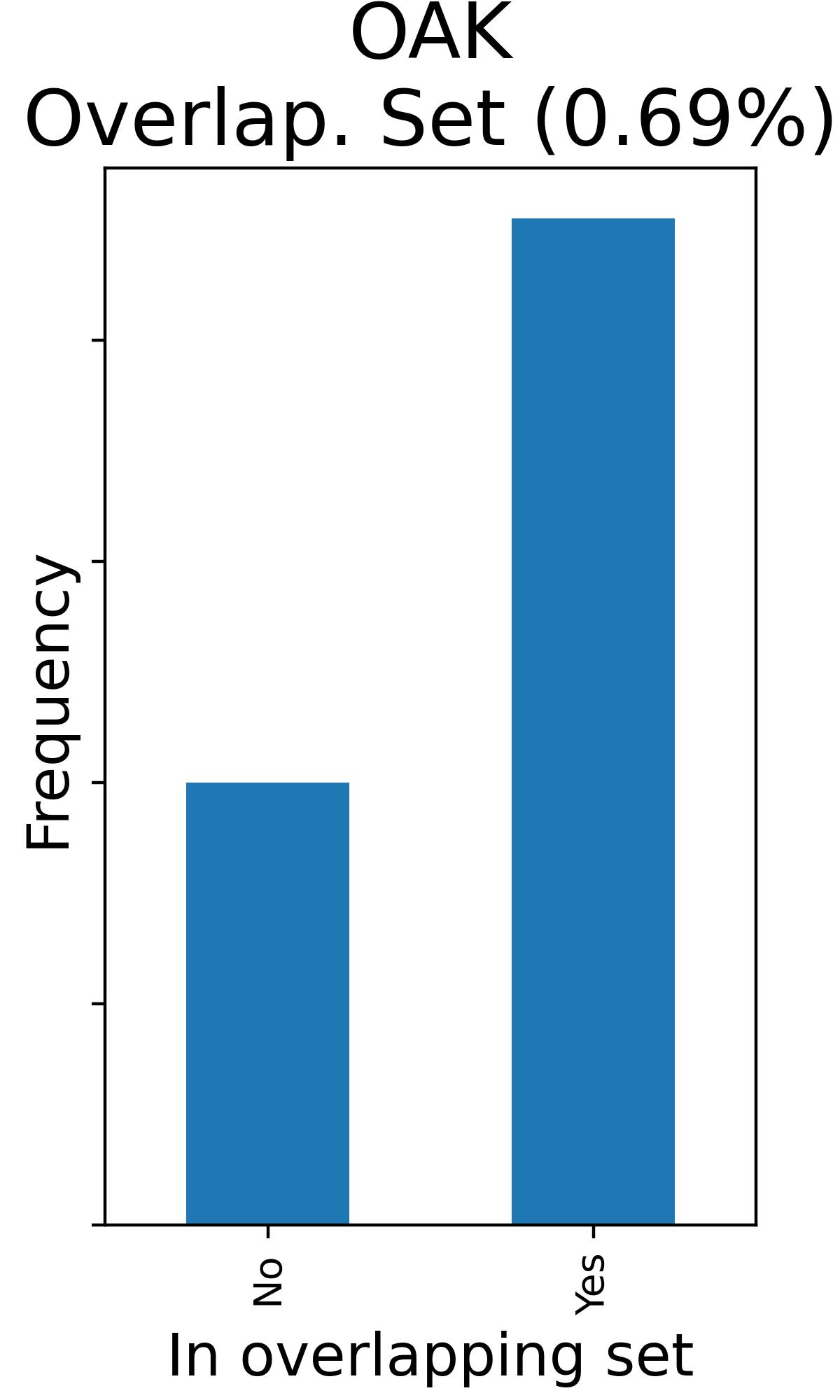}
\centering\includegraphics[width=3.3cm]{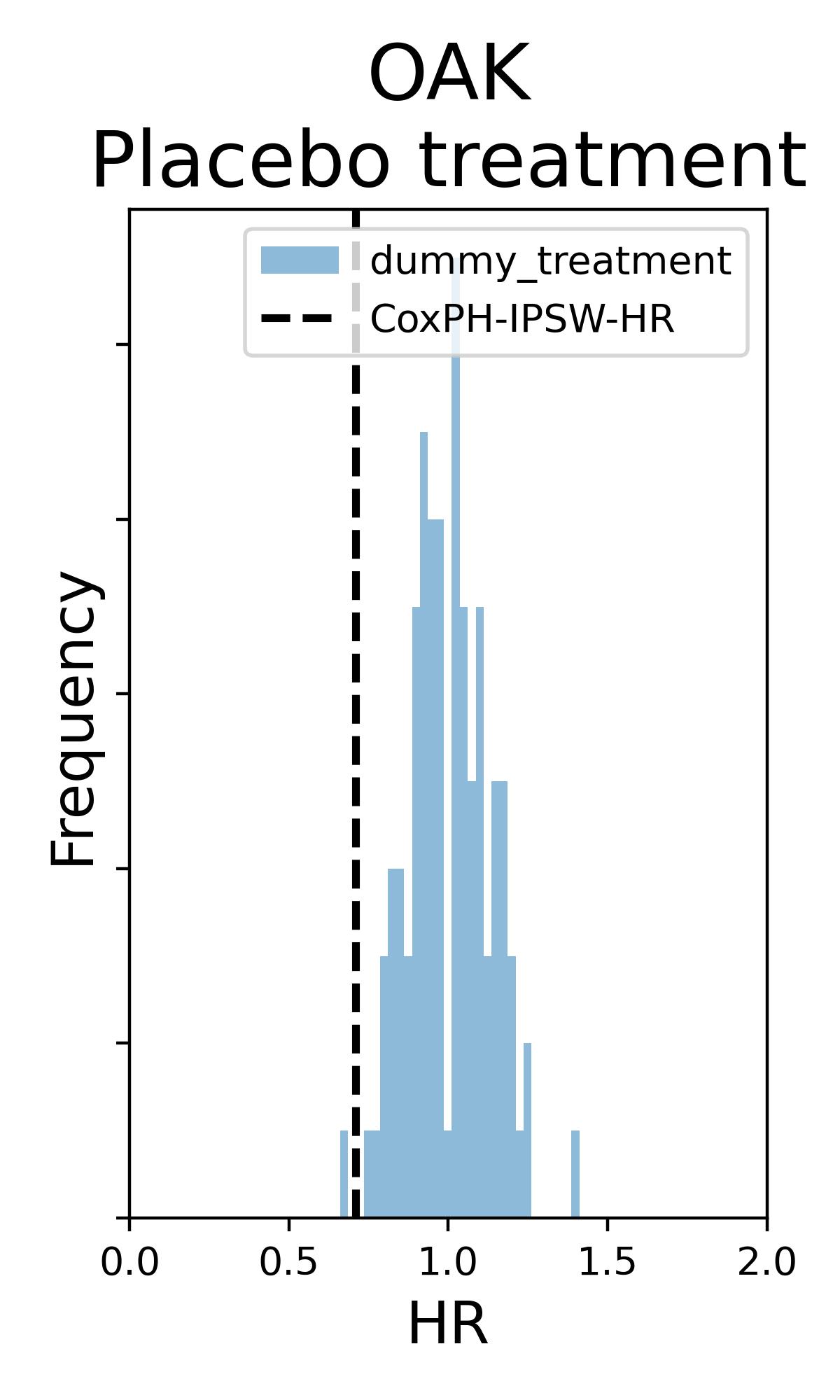}
\centering\includegraphics[width=3.3cm]{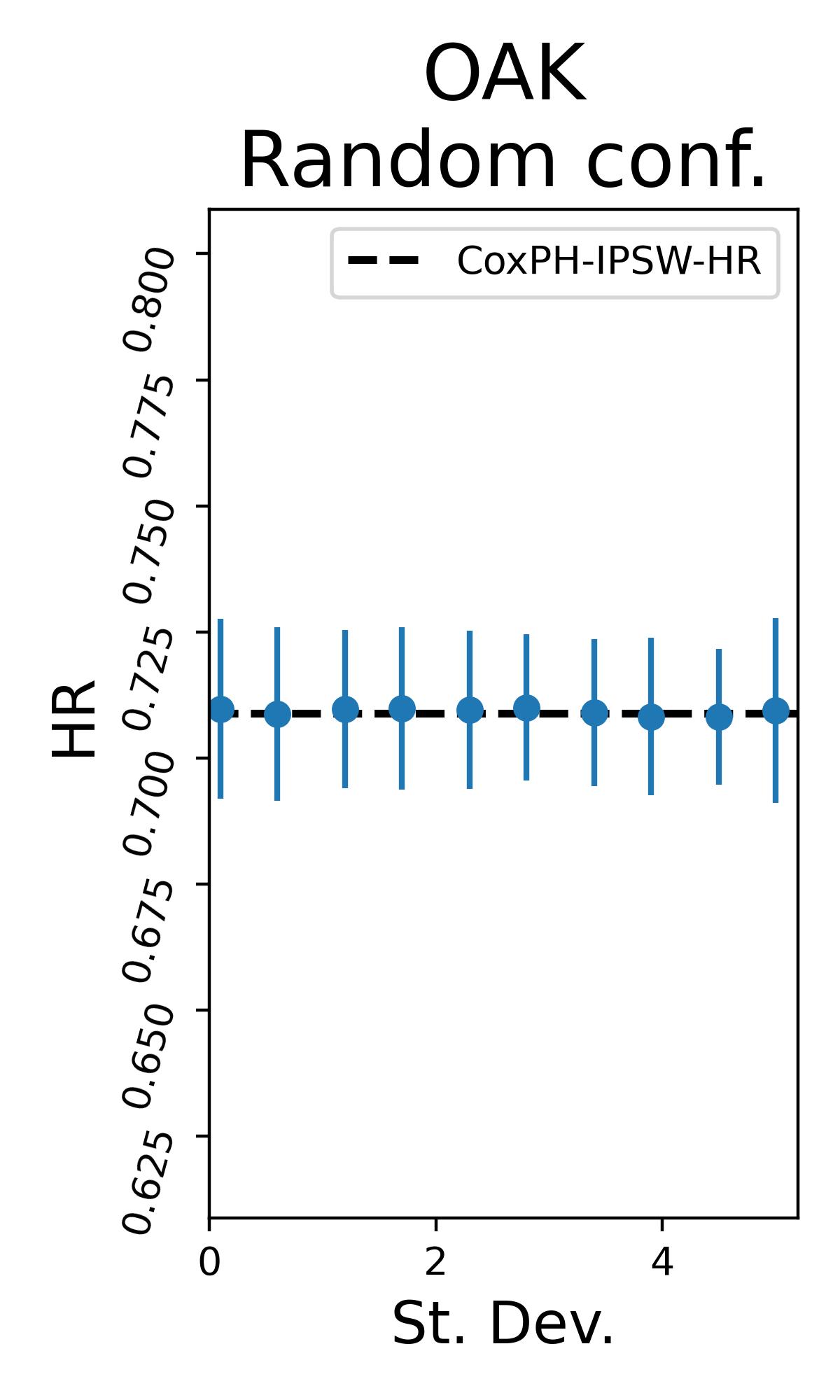}
\centering\includegraphics[width=3.2cm]{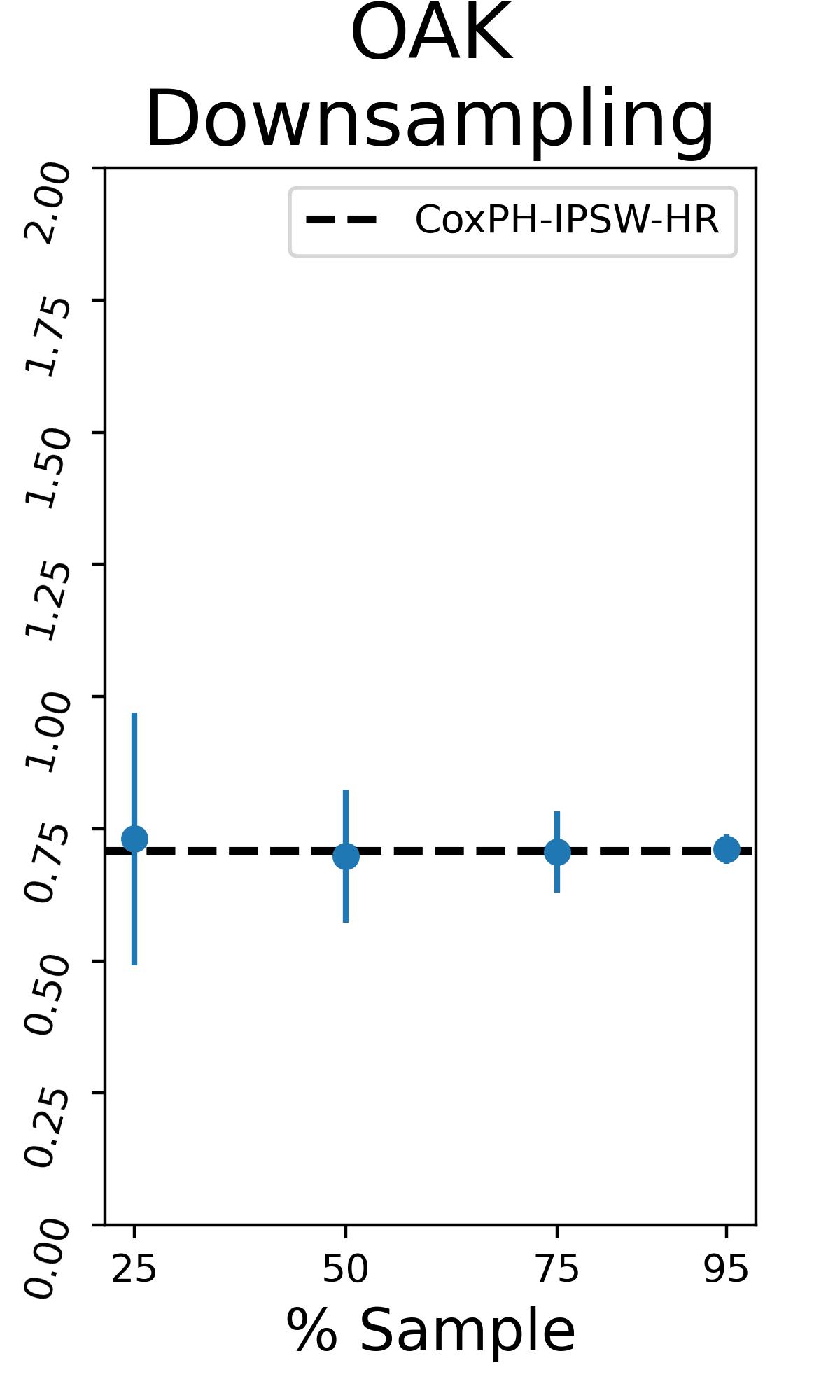} \\
\vspace{0.5cm}
\centering\includegraphics[width=3.2cm]{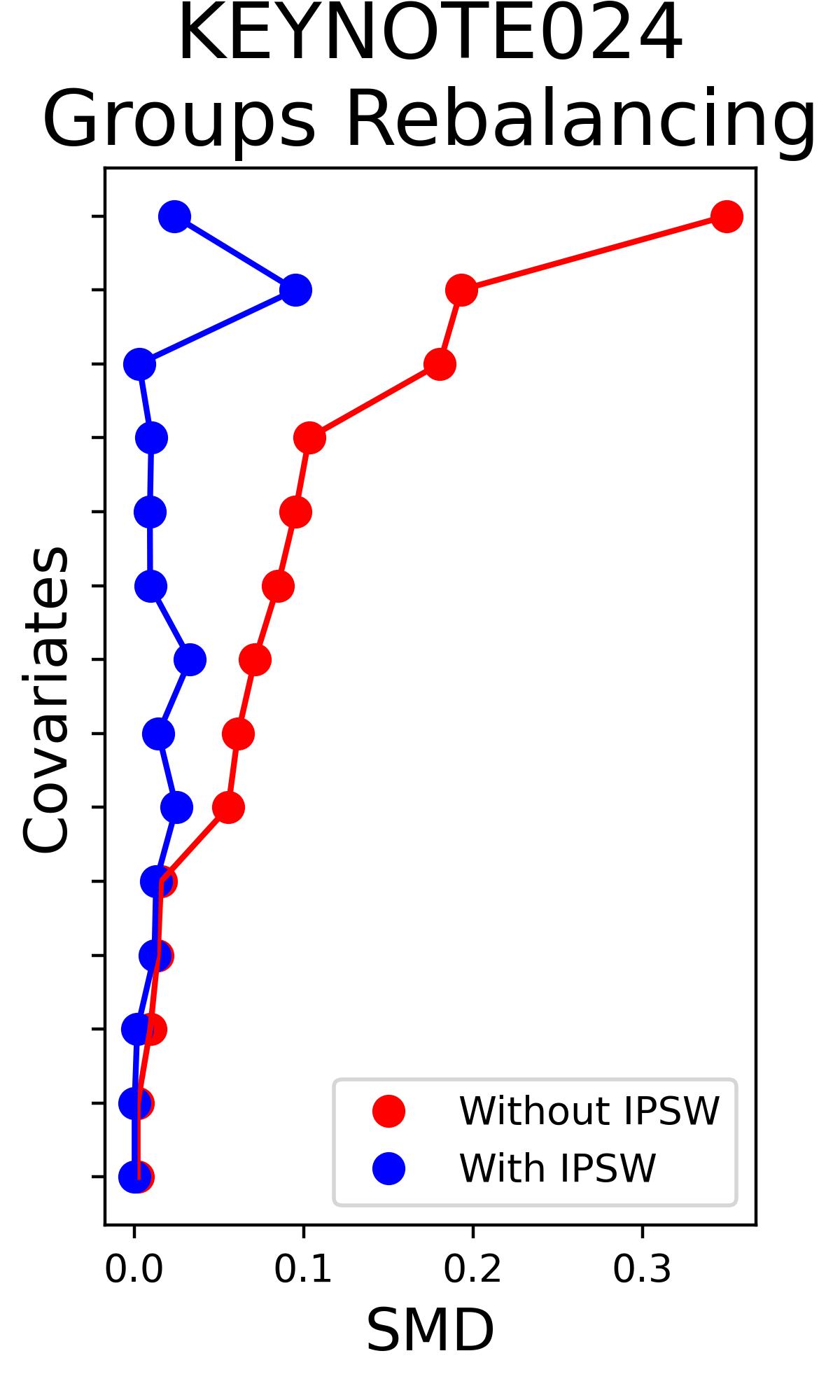}
\centering\includegraphics[width=3.2cm]{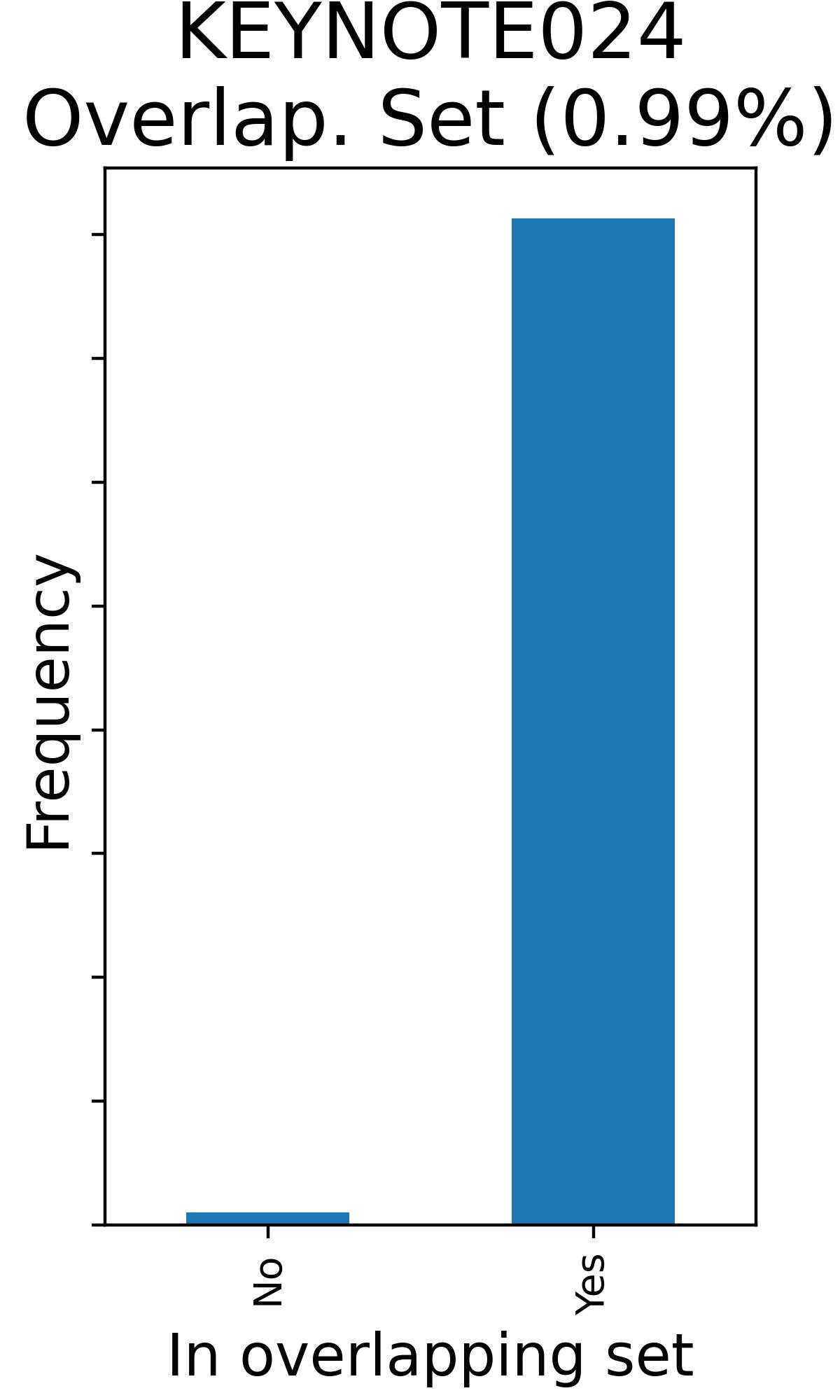}
\centering\includegraphics[width=3.3cm]{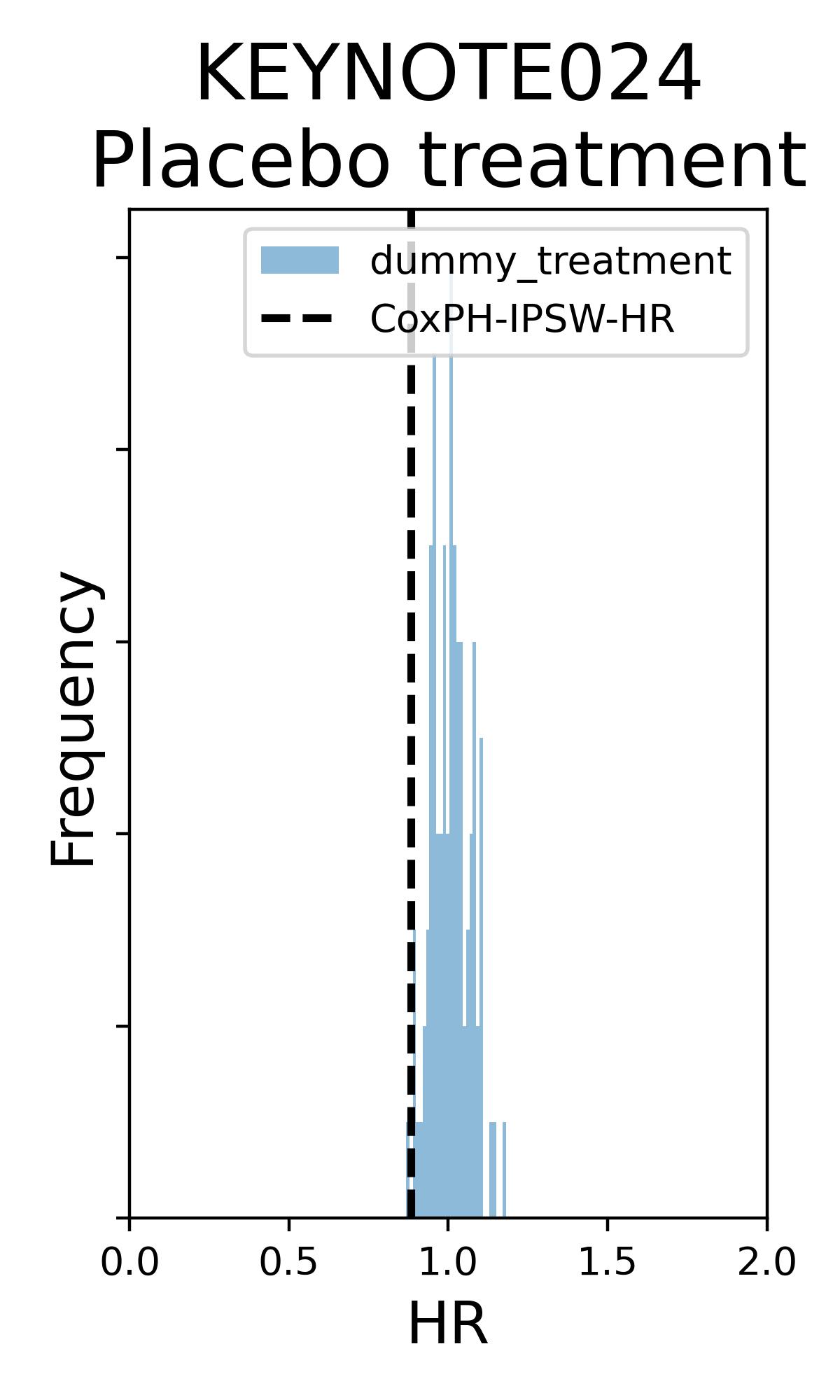}
\centering\includegraphics[width=3.3cm]{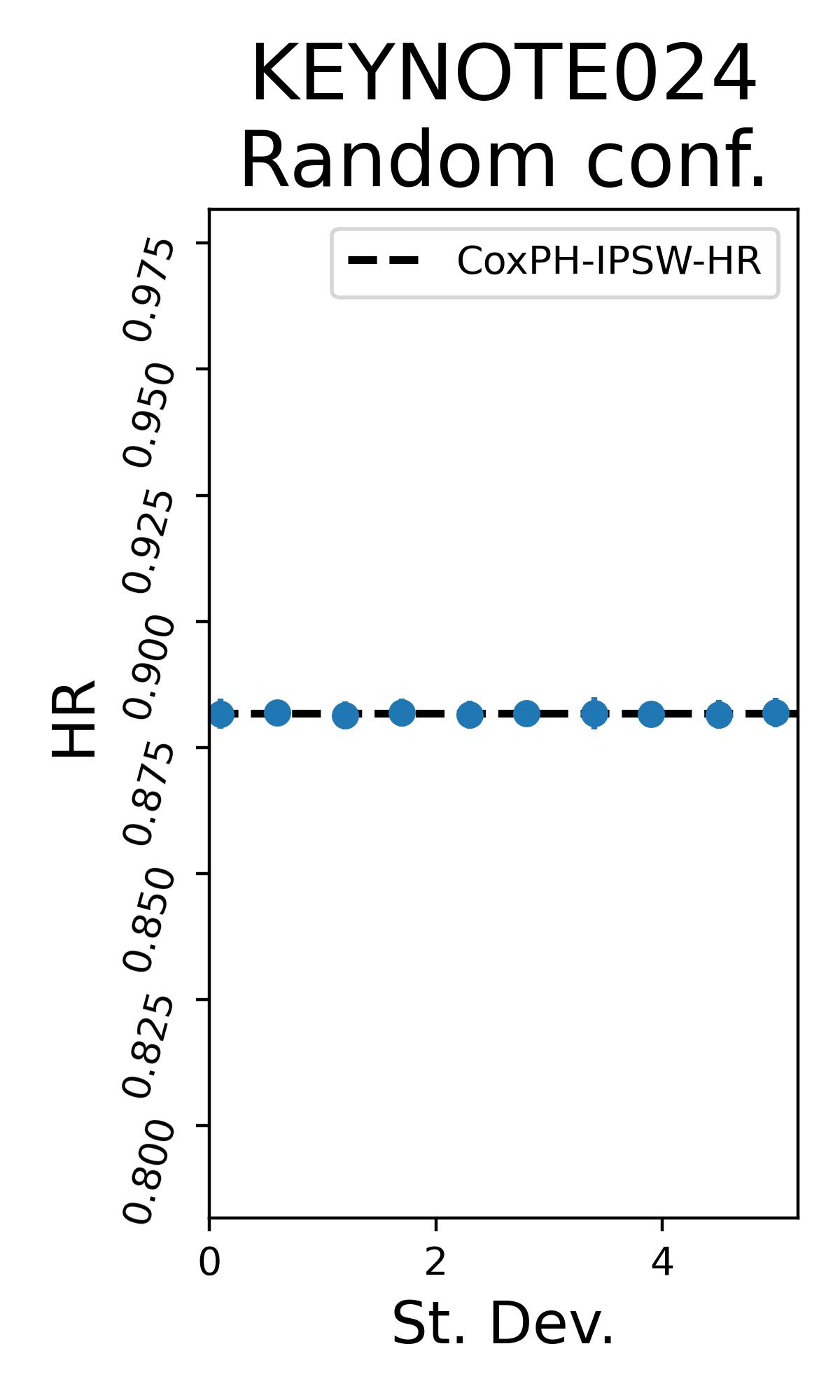}
\centering\includegraphics[width=3.2cm]{figures/diagnostics/downsampling_KEYNOTE024.jpeg}\\
\vspace{0.5cm}
\centering\includegraphics[width=3.2cm]{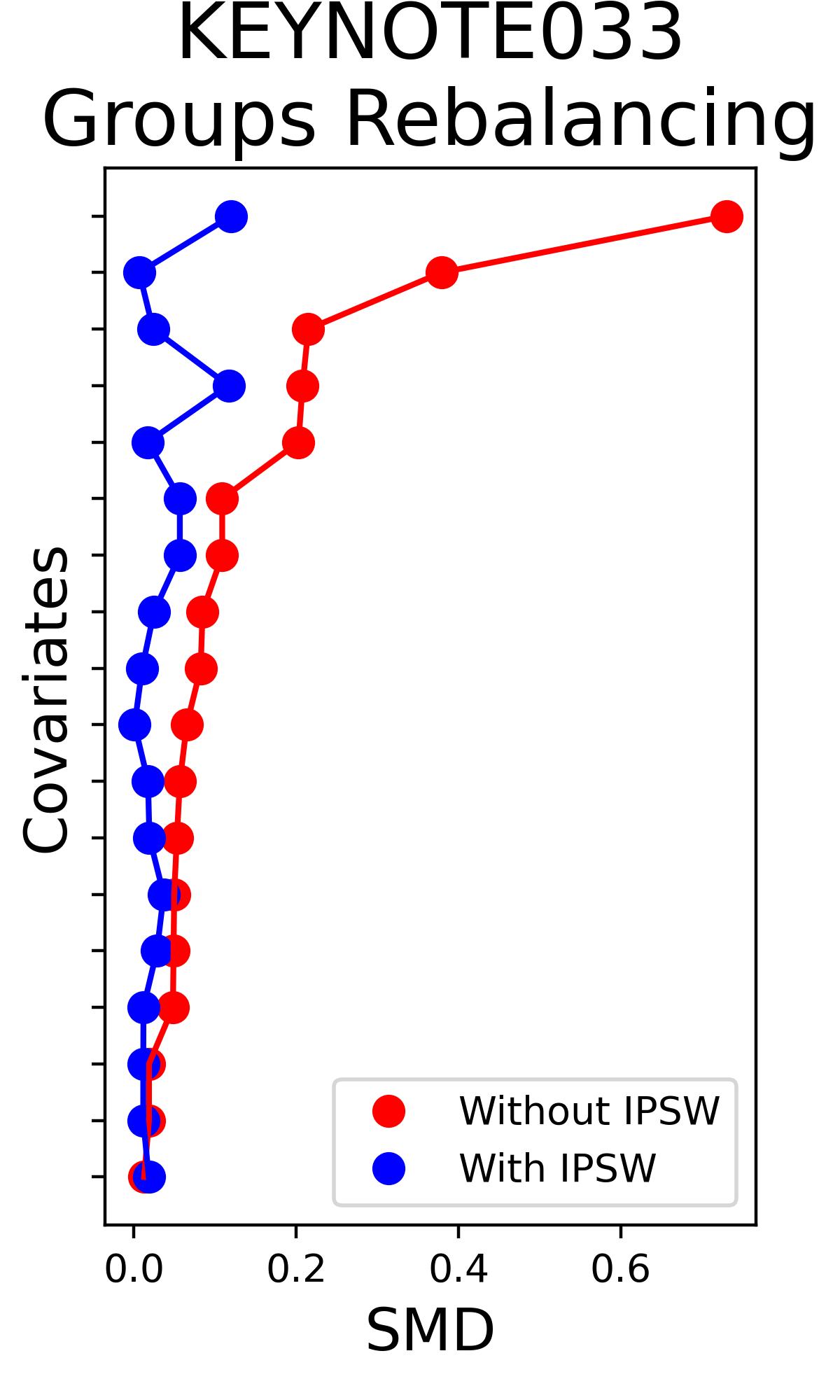}
\centering\includegraphics[width=3.2cm]{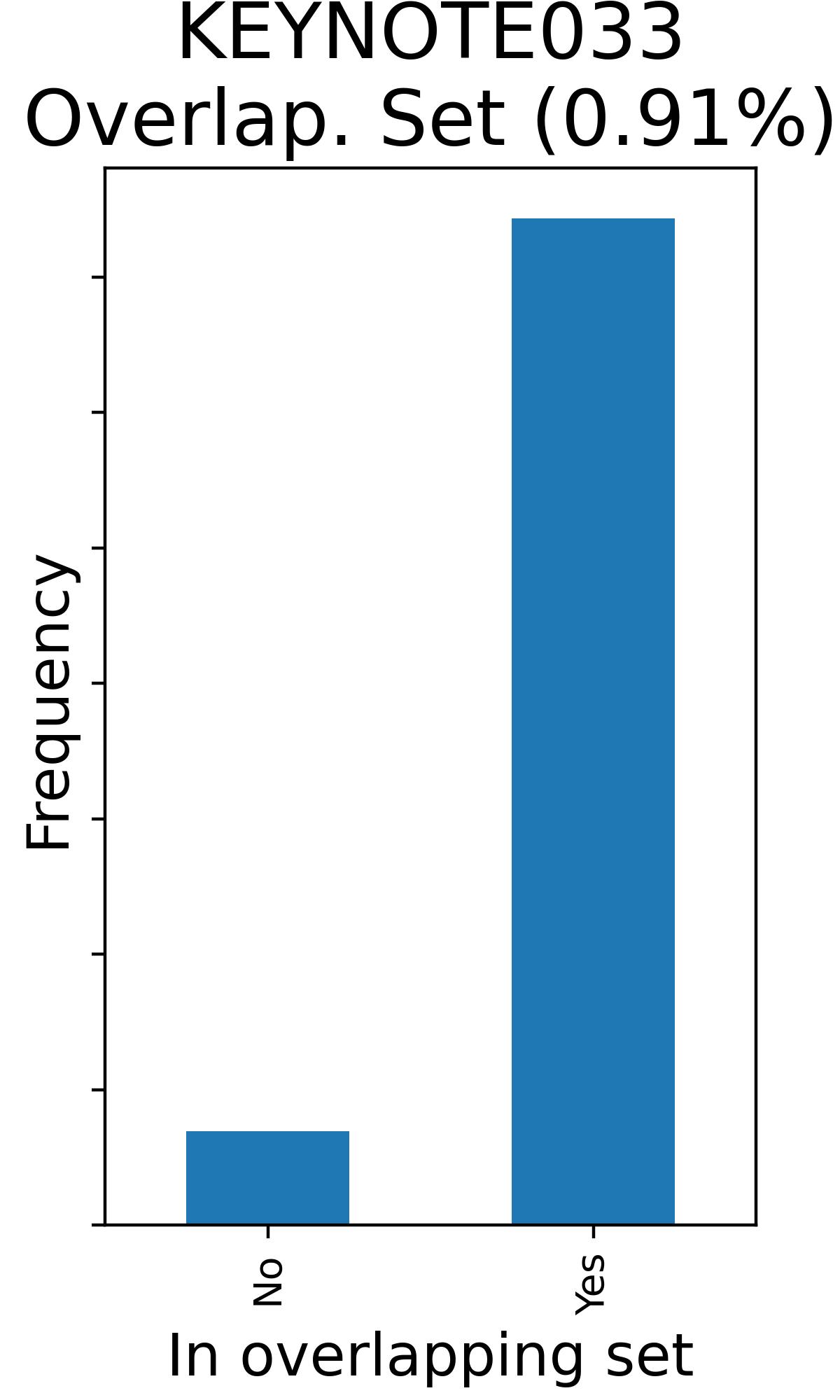}
\centering\includegraphics[width=3.3cm]{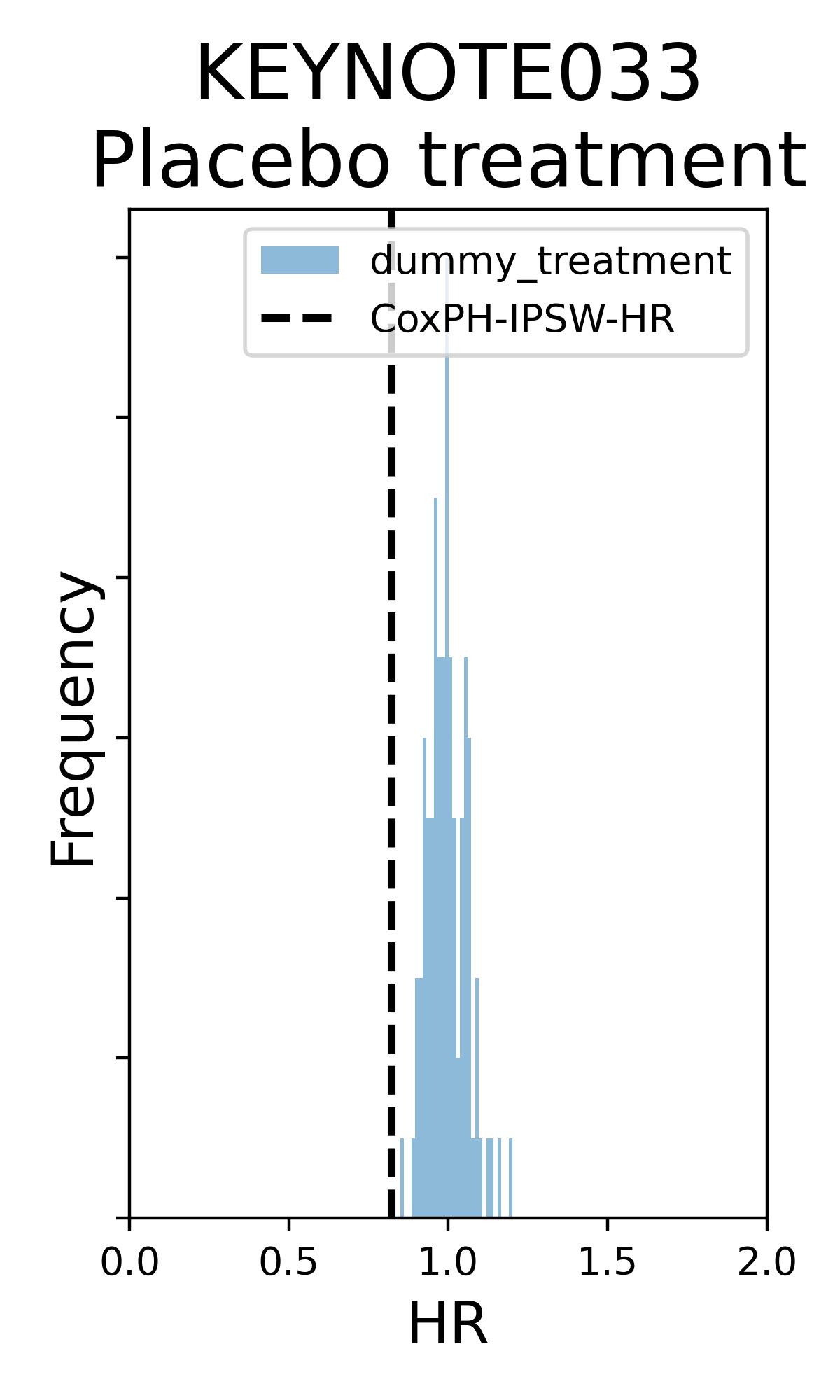}
\centering\includegraphics[width=3.3cm]{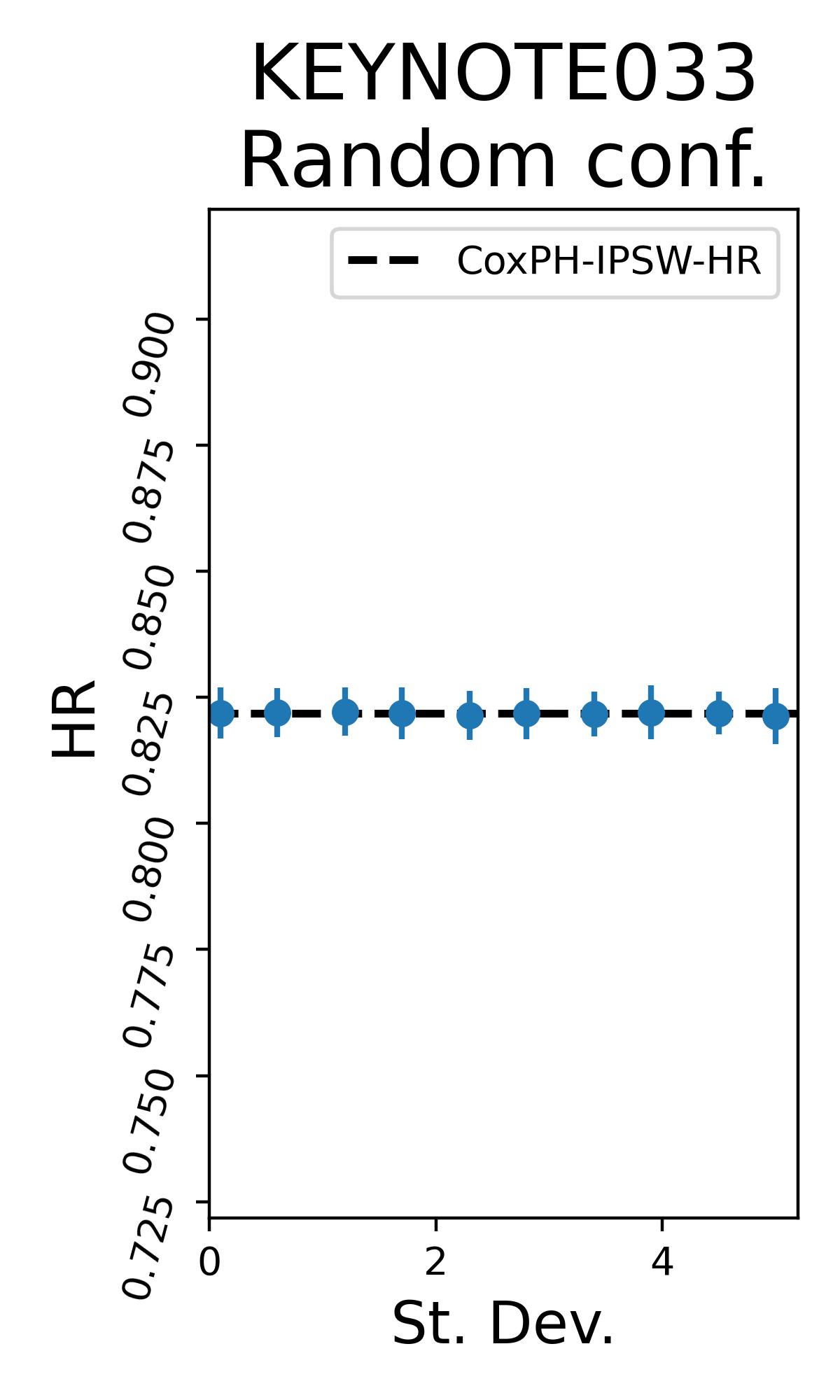}
\centering\includegraphics[width=3.2cm]{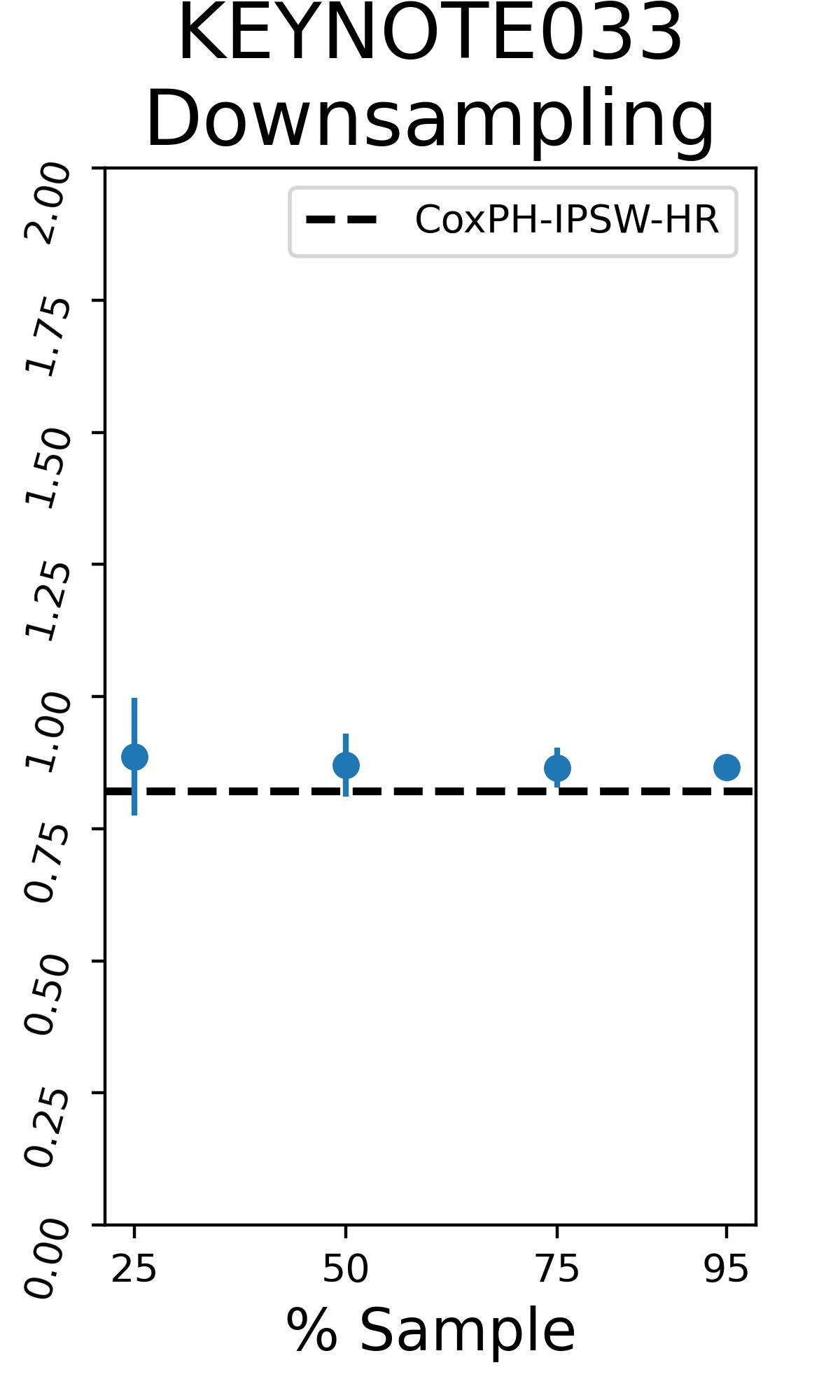}\\

\end{figure}

\begin{figure}[h!]
\caption{\emph{Results of the diagnostics test for CHECKMATE017 and EMPHASIS}}\label{fig:diagnosis_3}
\vspace{0.5cm} 

\vspace{0.5cm}
\centering\includegraphics[width=3.2cm]{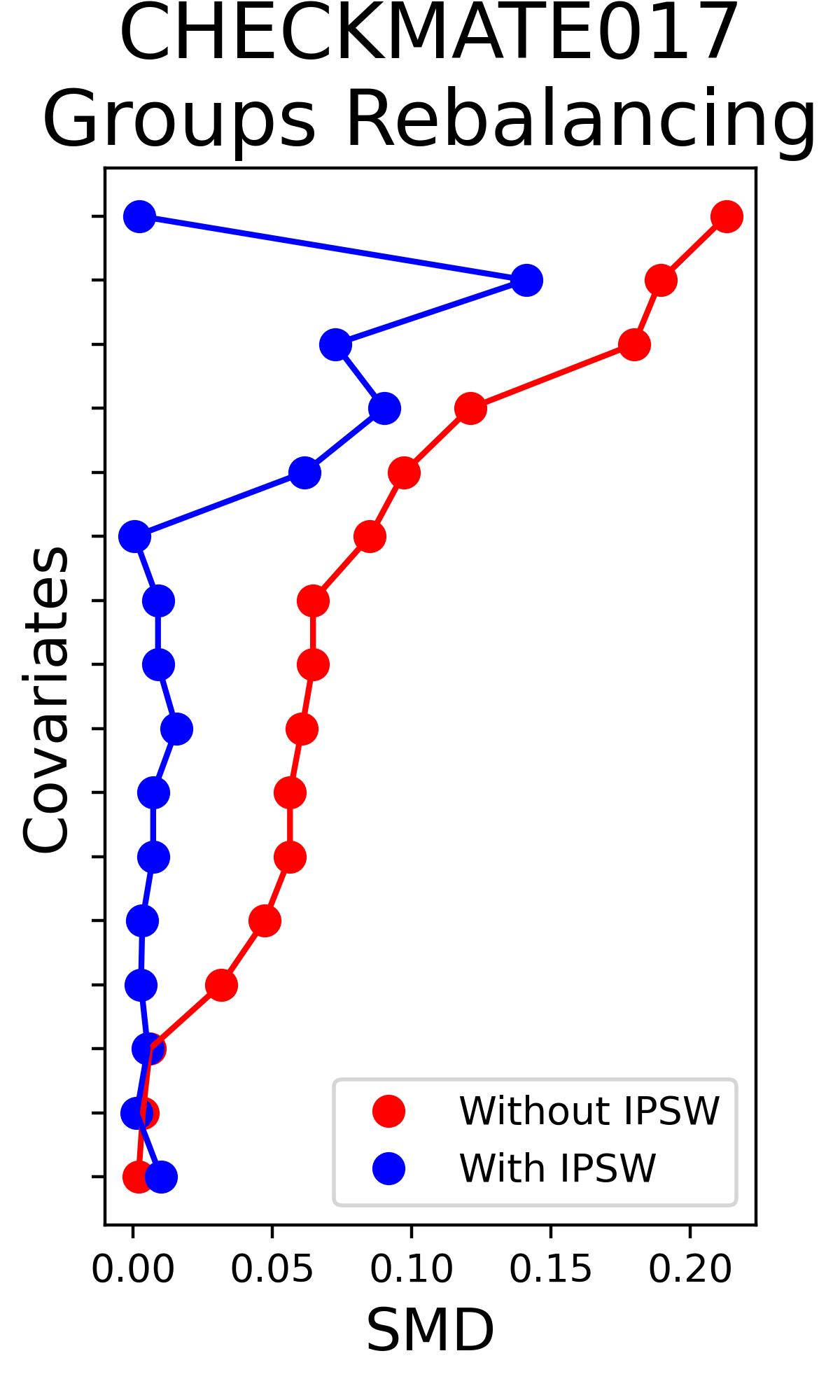}
\centering\includegraphics[width=3.2cm]{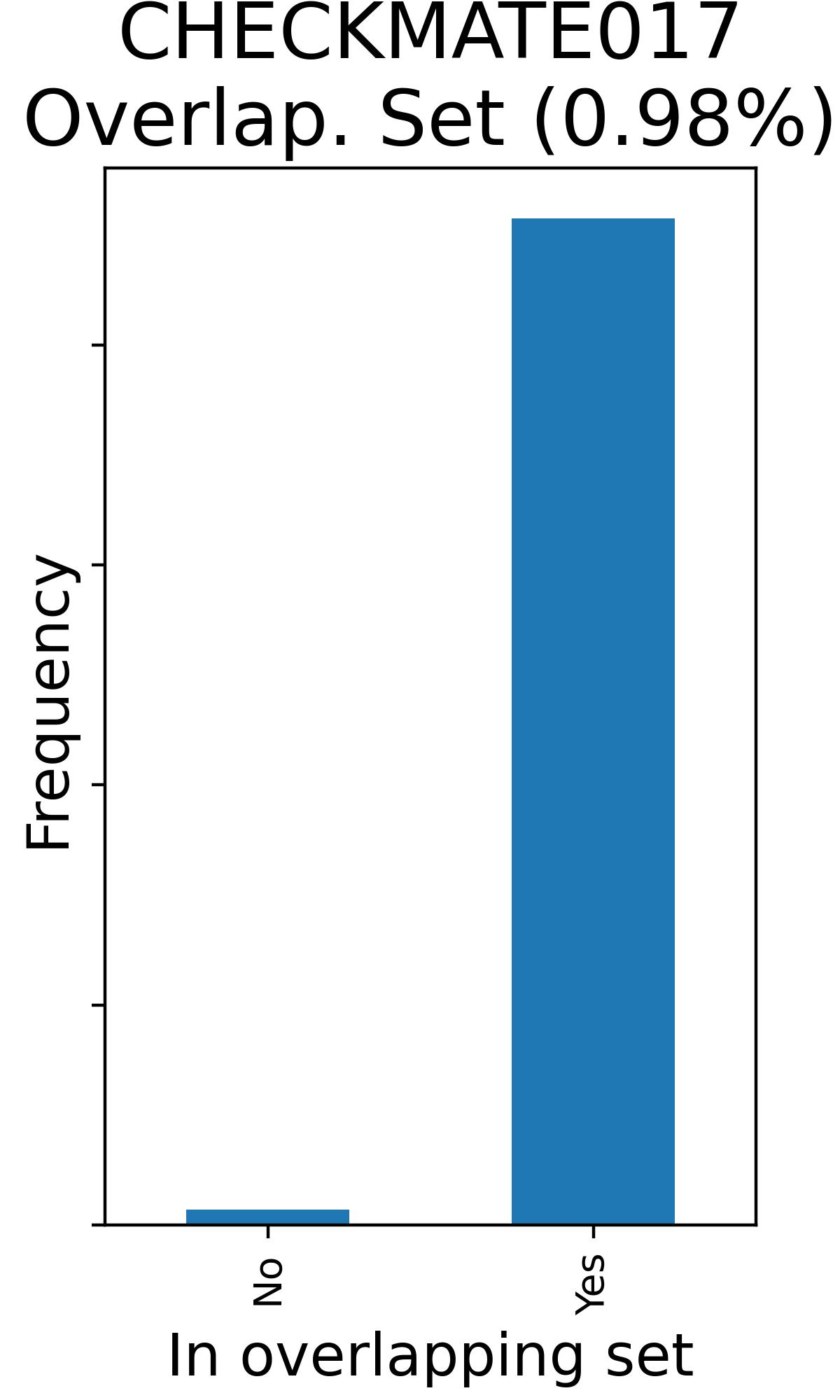}
\centering\includegraphics[width=3.3cm]{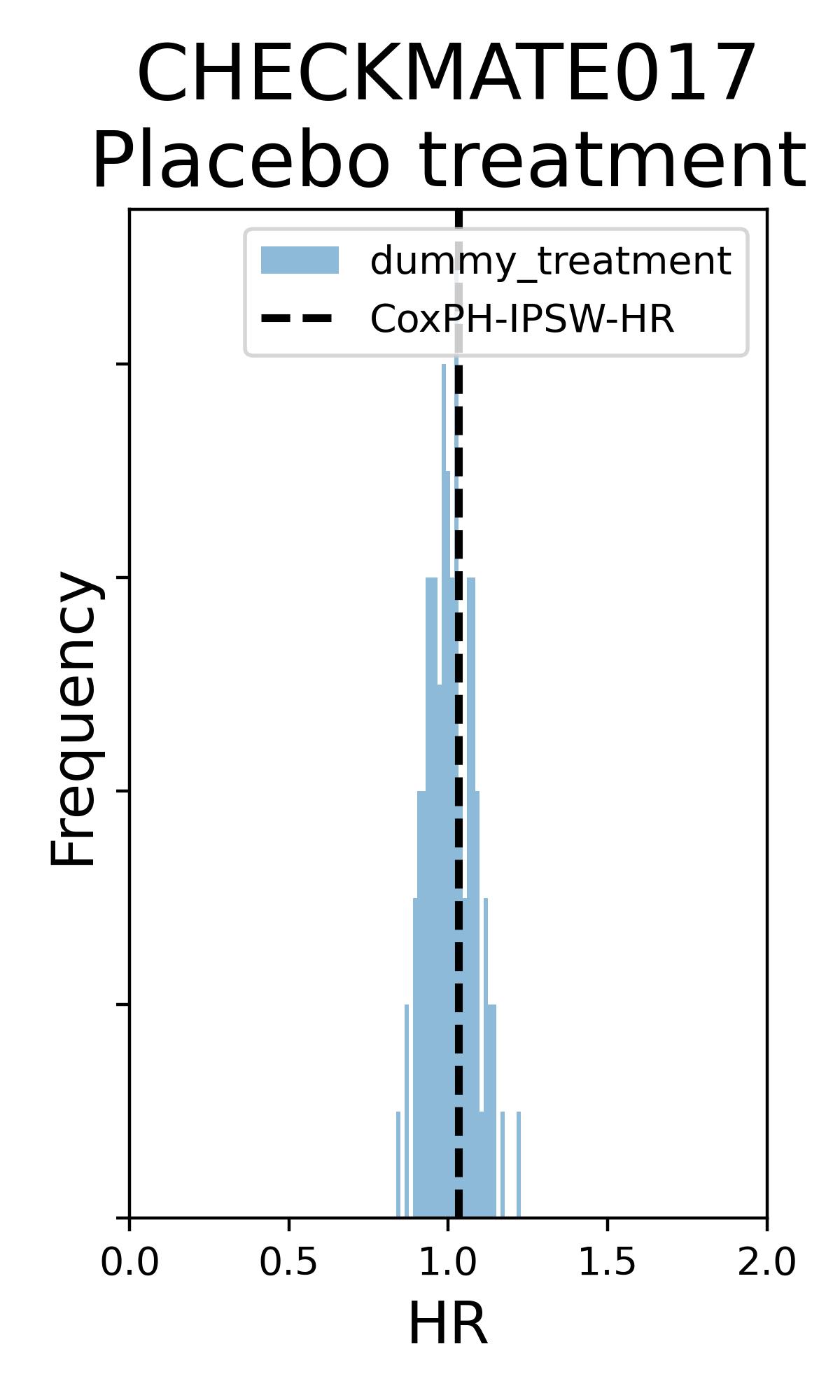}
\centering\includegraphics[width=3.3cm]{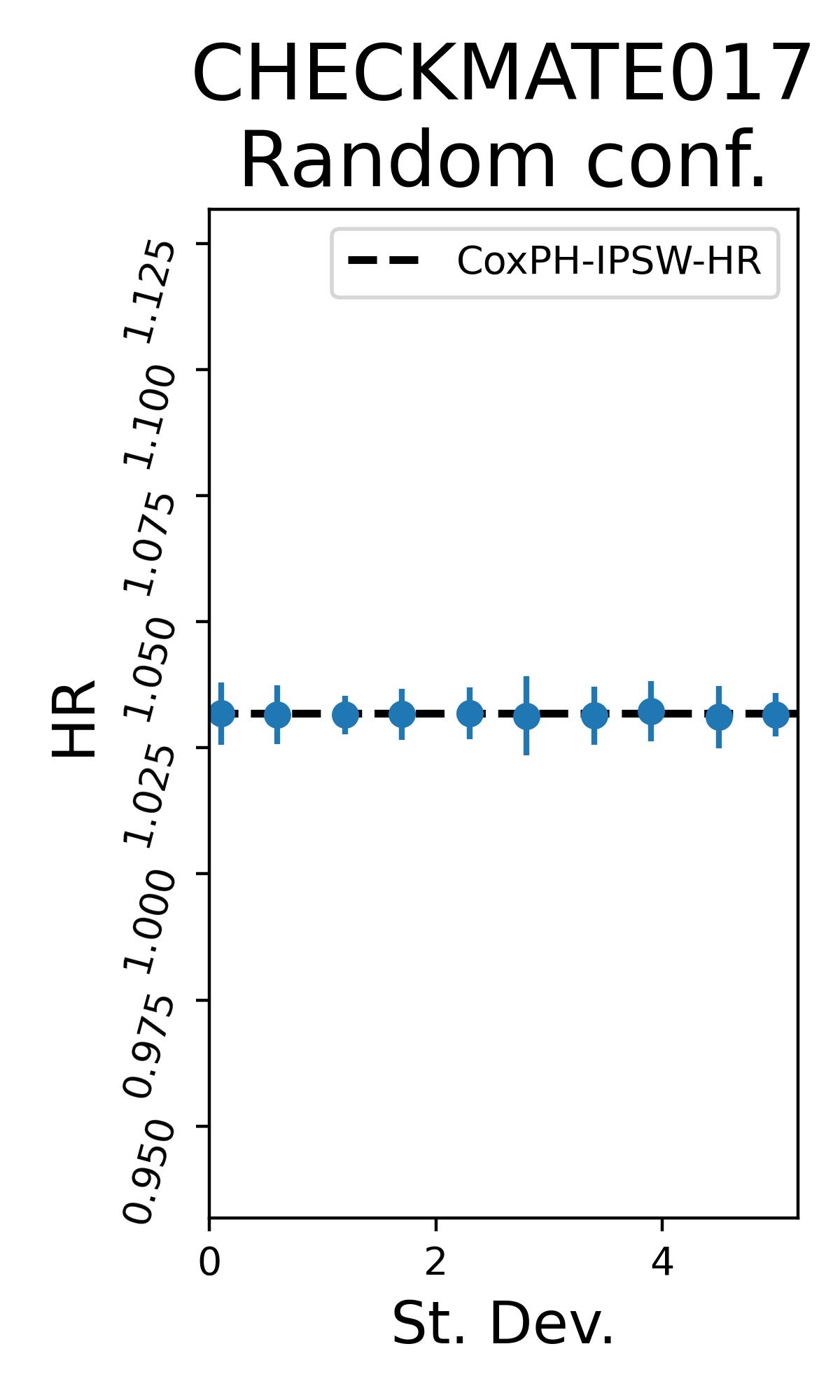}
\centering\includegraphics[width=3.2cm]{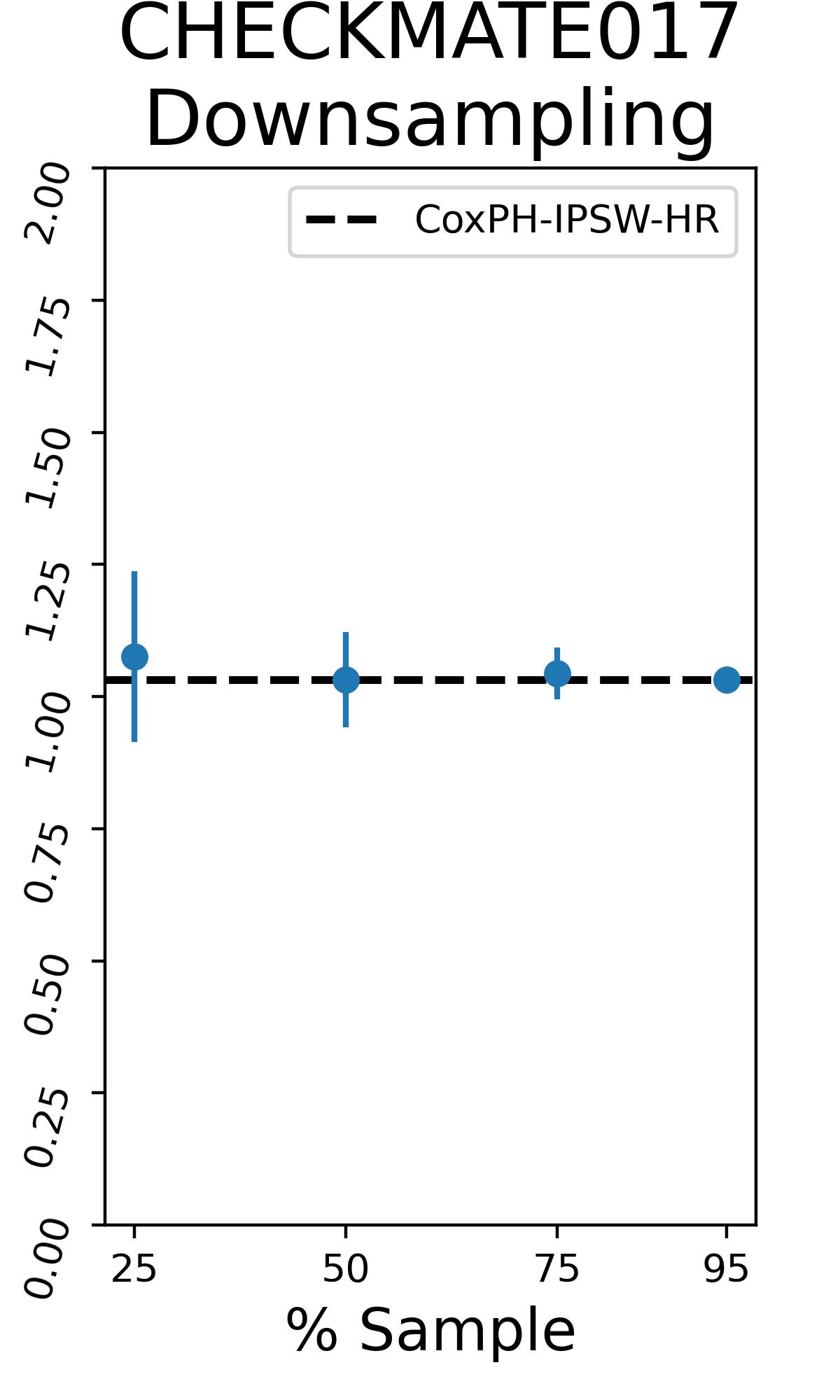}\\
\vspace{0.5cm}
\centering\includegraphics[width=3.2cm]{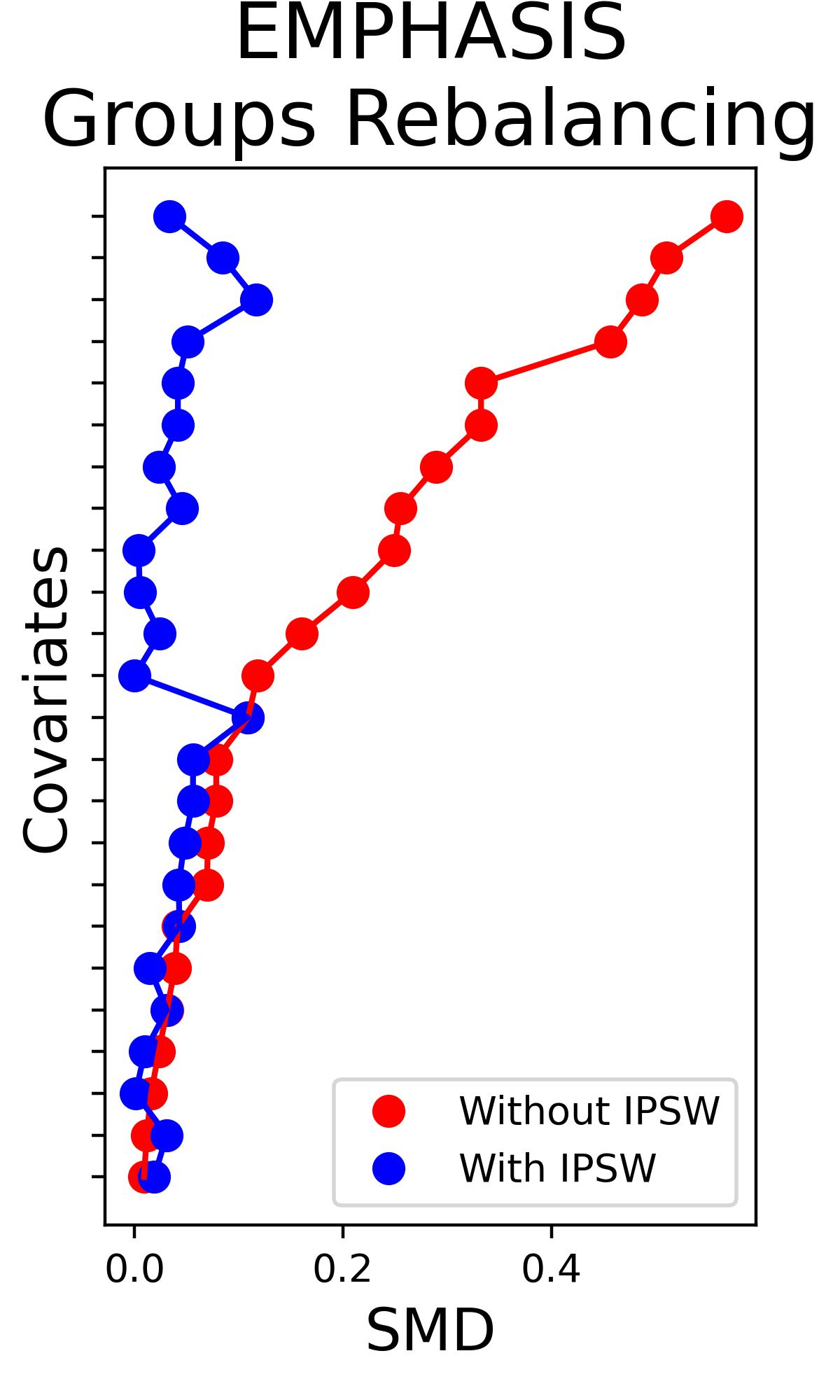}
\centering\includegraphics[width=3.2cm]{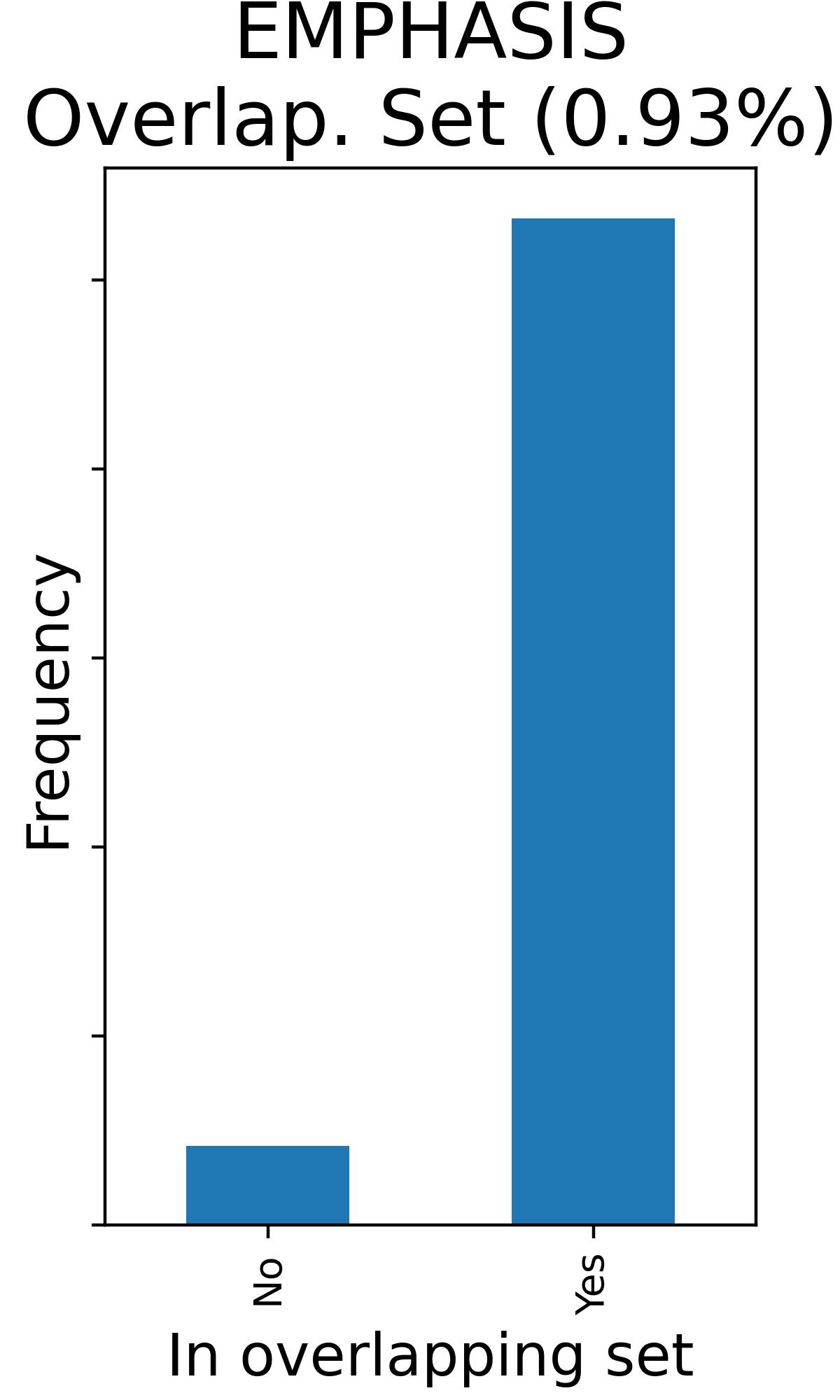}
\centering\includegraphics[width=3.3cm]{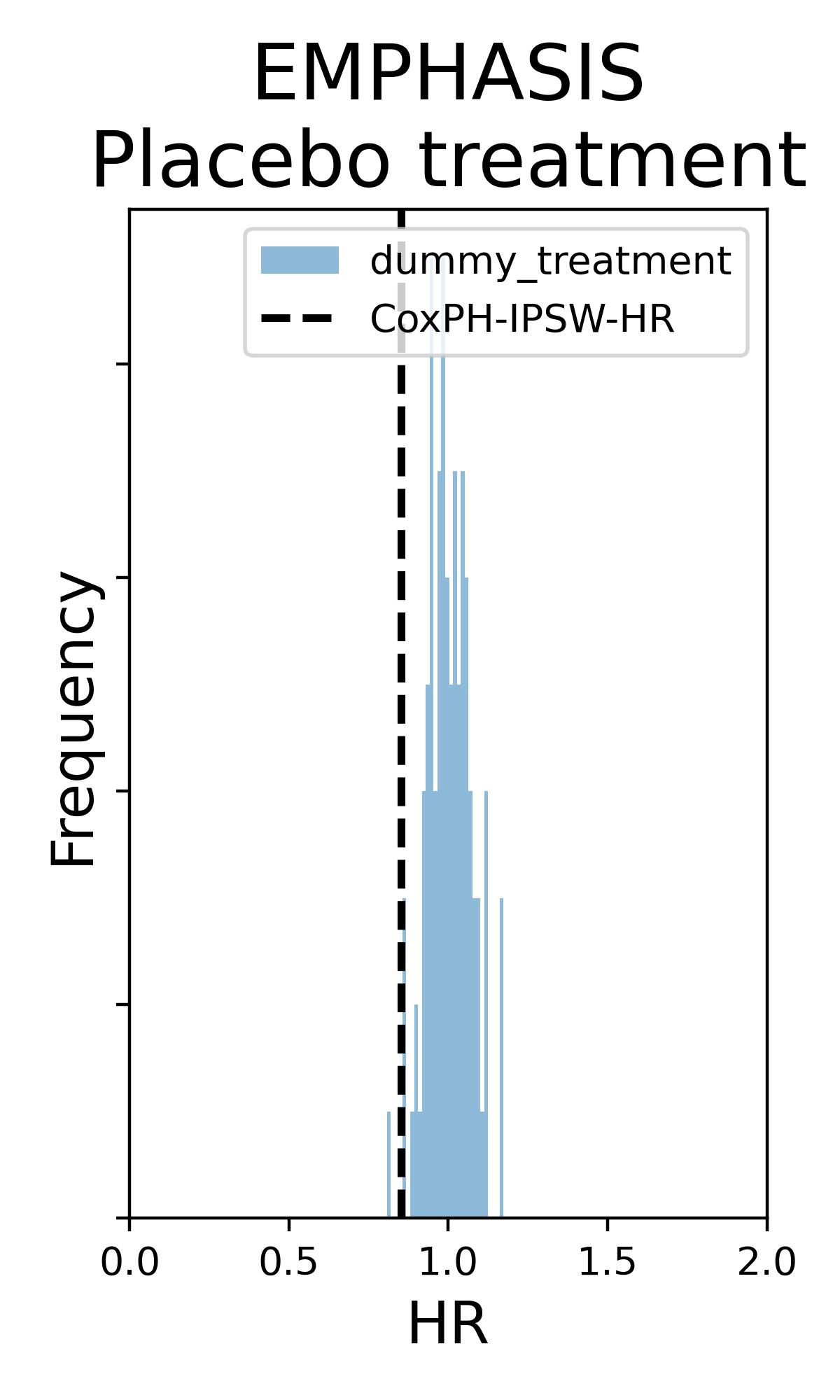}
\centering\includegraphics[width=3.3cm]{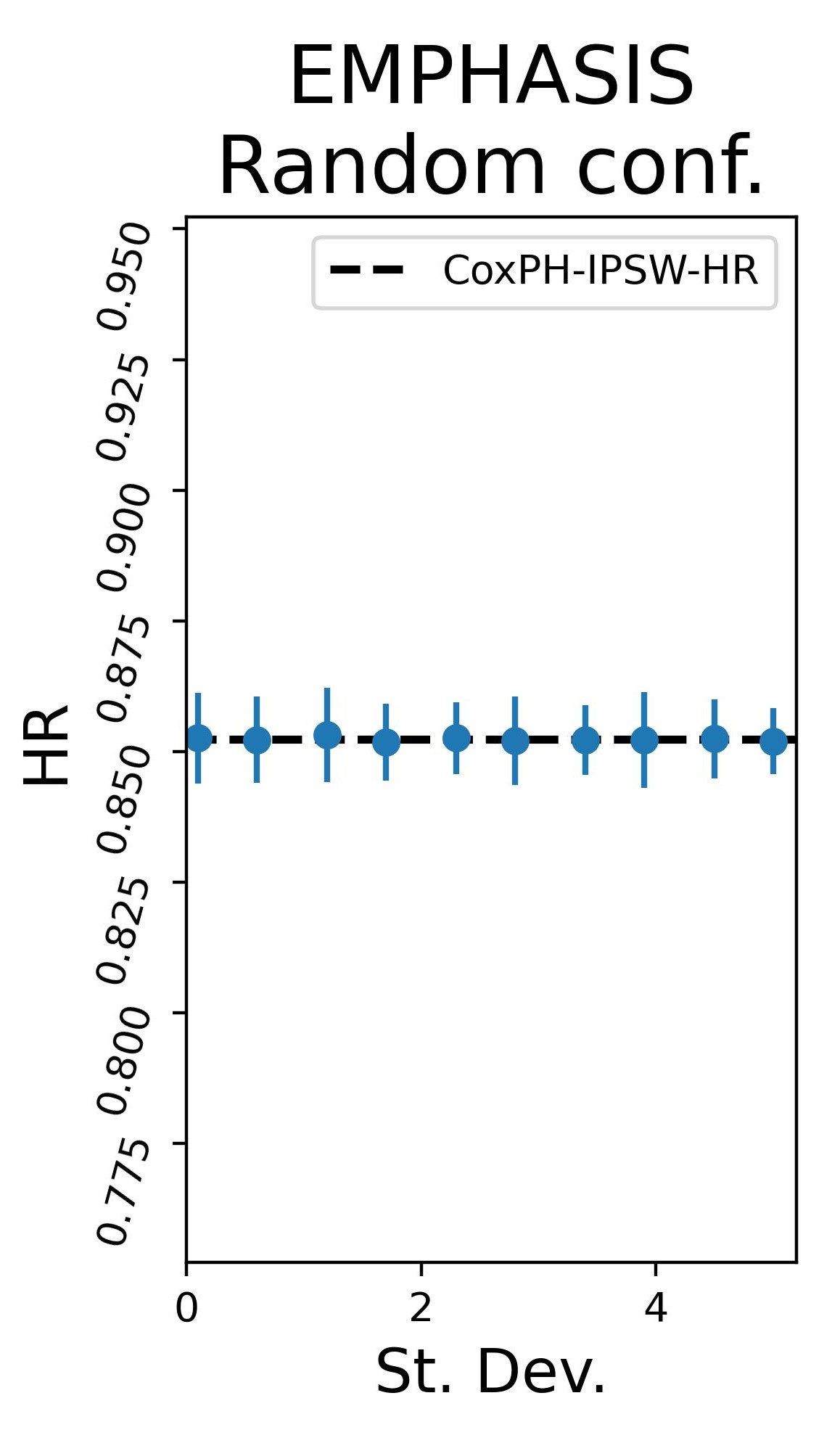}
\centering\includegraphics[width=3.2cm]{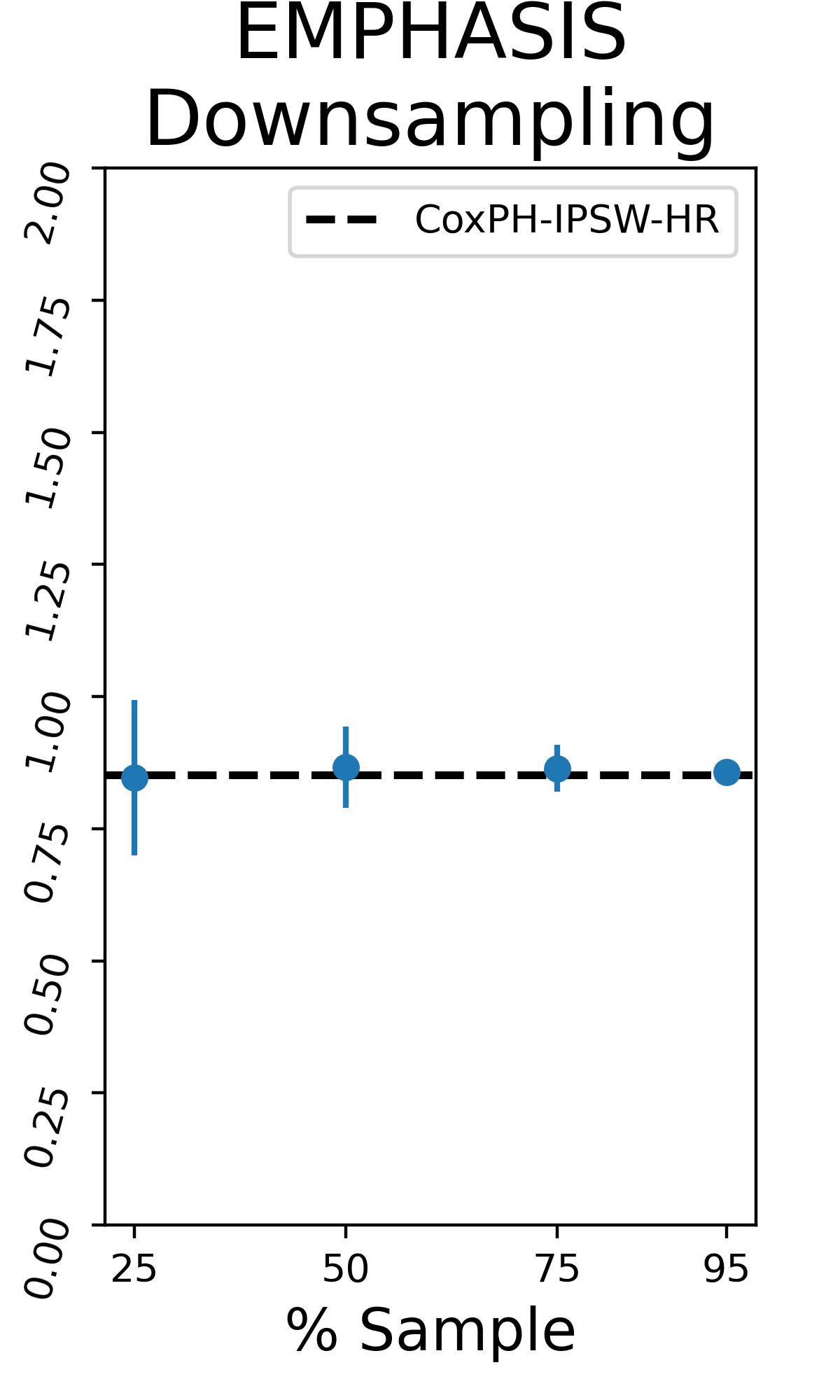}\\

\end{figure}

\begin{figure}[!ht]
\caption{\emph{Results of the diagnostics test for LUXLUNG6 and NCT02604342}. Overlapping and down-sampling test are not shown due to the limited sample size o these trials. Only re-balancing test are shown due the to the small sample size of these cohorts.}\label{fig:diagnosis_4}
\vspace{0.5cm} 

\centering\includegraphics[width=3.2cm]{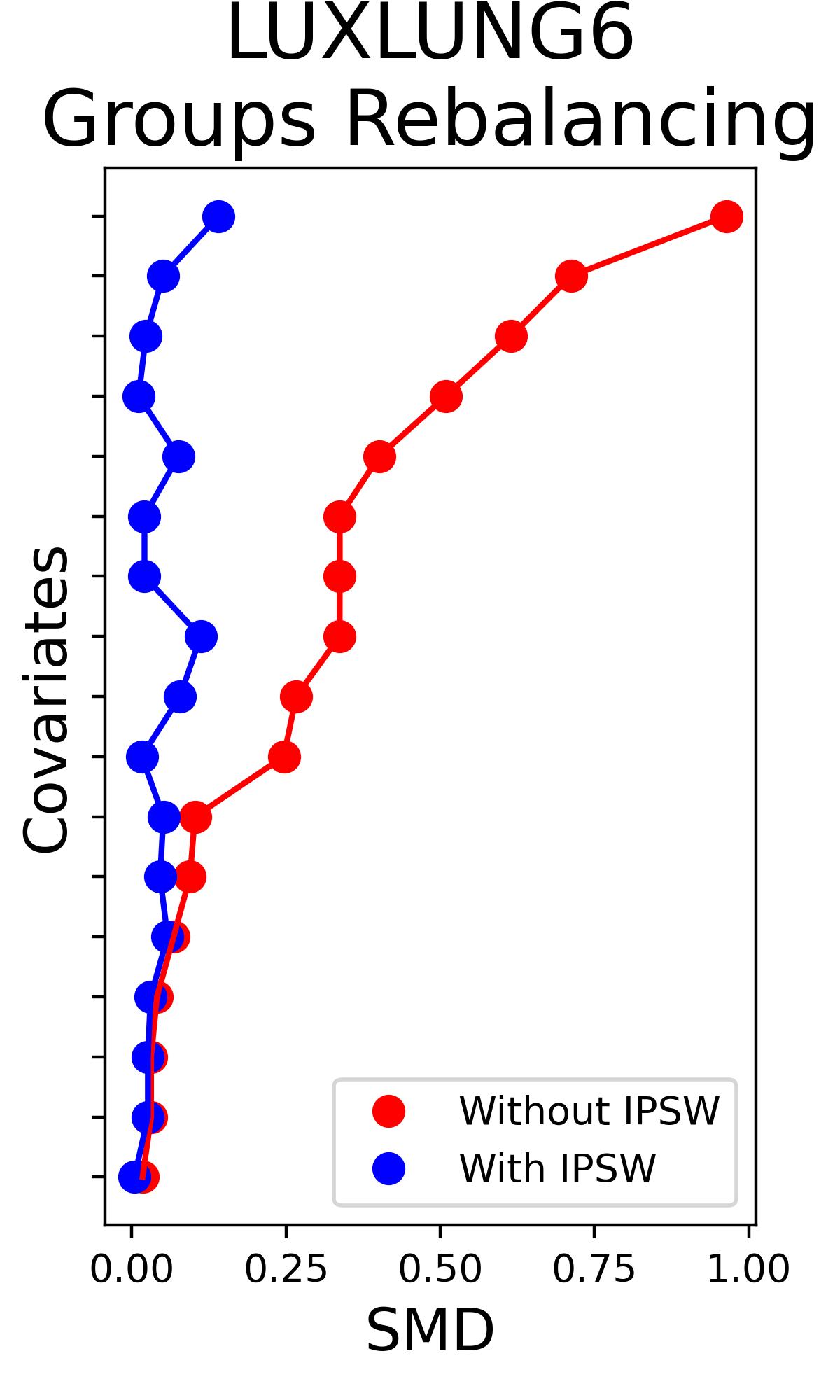}
\centering\includegraphics[width=3.2cm]{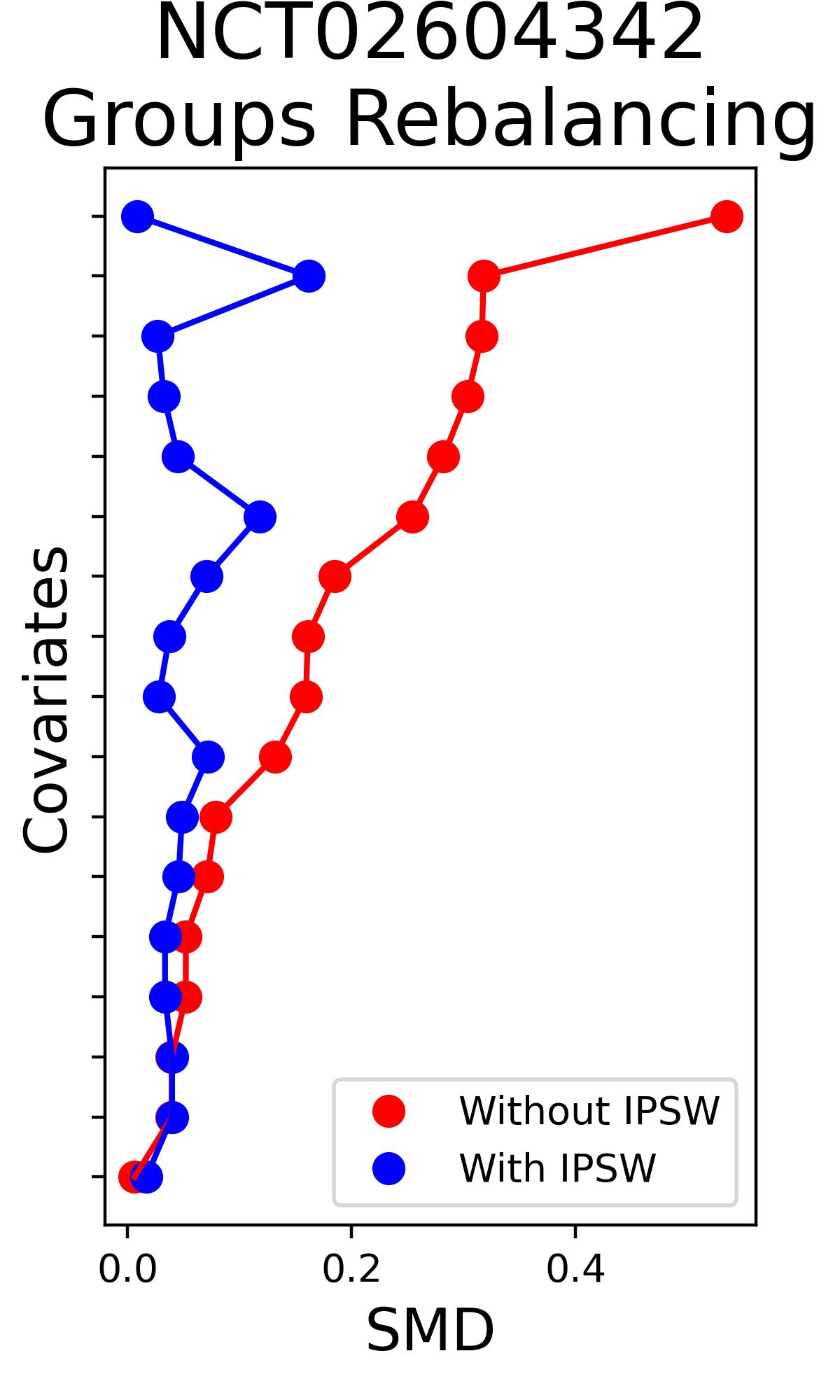}

\end{figure}

\clearpage
\section{Emulation of metastatic pancreatic cancer trial MPACT}\label{sup:pancreatic}

\begin{figure}[h!]
\caption{Results of the diagnostics test for MPACT. Only the re-balancing test is shown due the to the small sample size of these cohorts.}\label{figure:mpact}
 \vspace{0.5cm}
\centering
\includegraphics[width=5cm]{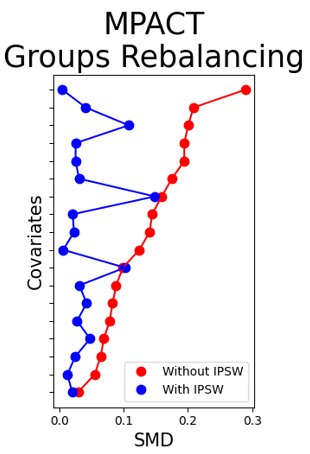}
\end{figure} 

The MPACT trial \parencite{VonHoff2013} compares the combination of nab-paclitaxel (albumin-bound paclitaxel) and gemcitabine vs. gemcitabine monotherapy for patients with metastatic pancreatic cancer as first-line treatment. The emulation of this trial incorporates the following eligibility criteria: metastatic pancreatic adenocarcinoma, age $\geq$ 18, AST and ALT $\leq 2.5 ×$ ULN, total bilirubin $\leq$ ULN, absolute neutrophil count (ANC) $\geq 1.5 × 10^9$/L, platelet count $\geq 100,000/mm^3$ ($100 × 10^9/L$), hemoglobin (Hgb) $\geq 9 g/d$, and no known brain metastases. The simulation yields a hazard ratio (HR) =0.77 (95\% CI = [0.481,0.985]) with total patient count n = 131 (control: 29, treatment: 109). This is statistically equivalent to the reported HR = 0.72 (95\% = [0.62, 0.83]). Removing the eligibility criteria and including locally advanced pancreatic cancer patients, the simulation yields HR = 0.691 (95\% CI = [0.485, 0.985]) with n=214 (control: 55, treatment: 159). 